\documentclass[letterpaper]{article} 
\usepackage{aaai2026}  
\nocopyright
\usepackage{times}  
\usepackage{helvet}  
\usepackage{courier}  
\usepackage[hyphens]{url}  
\usepackage{graphicx} 
\usepackage{natbib}  
\usepackage{caption} 
\usepackage{algorithm}
\usepackage[noend]{algorithmic}
\usepackage{bm}
\usepackage{dsfont}
\usepackage{booktabs}
\usepackage{multirow}
\usepackage{amsmath}
\usepackage{amsfonts}

\usepackage{newfloat}
\usepackage{listings}
\DeclareCaptionStyle{ruled}{labelfont=normalfont,labelsep=colon,strut=off} 
\floatstyle{ruled}
\newfloat{listing}{tb}{lst}{}
\floatname{listing}{Listing}
\usepackage{subcaption}

\title{Sparse-Autoencoder-Guided Internal Representation Unlearning\\ for Large Language Models}
\author{
Tomoya Yamashita, Akira Ito, Yuuki Yamanaka, Masanori Yamada, Takayuki Miura, Toshiki Shibahara
}
\affiliations{
NTT\\
tomoya.yamashita@ntt.com, akira.itoh@ntt.com, yuuki.yamanaka@ntt.com, masanori.yamada@ntt.com, tkyk.miura@ntt.com, toshiki.shibahara@ntt.com
}

\usepackage{bibentry}

\begin{document}

\maketitle

\begin{abstract}
As large language models (LLMs) are increasingly deployed across various applications, privacy and copyright concerns have heightened the need for more effective LLM unlearning techniques.
Many existing unlearning methods aim to suppress undesirable outputs through additional training (e.g., gradient ascent), which reduces the probability of generating such outputs.
While such suppression-based approaches can control model outputs, they may not eliminate the underlying knowledge embedded in the model's internal activations; muting a response is not the same as forgetting it.
Moreover, such suppression-based methods often suffer from model collapse.
To address these issues, we propose a novel unlearning method that directly intervenes in the model's internal activations.
In our formulation, forgetting is defined as a state in which the activation of a forgotten target is indistinguishable from that of ``unknown'' entities.
Our method introduces an unlearning objective that modifies the activation of the target entity away from those of known entities and toward those of unknown entities in a sparse autoencoder latent space.
By aligning the target’s internal activation with those of unknown entities, we shift the model’s recognition of the target entity from ``known'' to ``unknown'', achieving genuine forgetting while avoiding over-suppression and model collapse.
Empirically, we show that our method effectively aligns the internal activations of the forgotten target, a result that the suppression-based approaches do not reliably achieve.
Additionally, our method effectively reduces the model’s recall of target knowledge in question-answering tasks without significant damage to the non-target knowledge.
\end{abstract}

\begin{figure*}[tb]
\center
\includegraphics[width=0.92\textwidth]{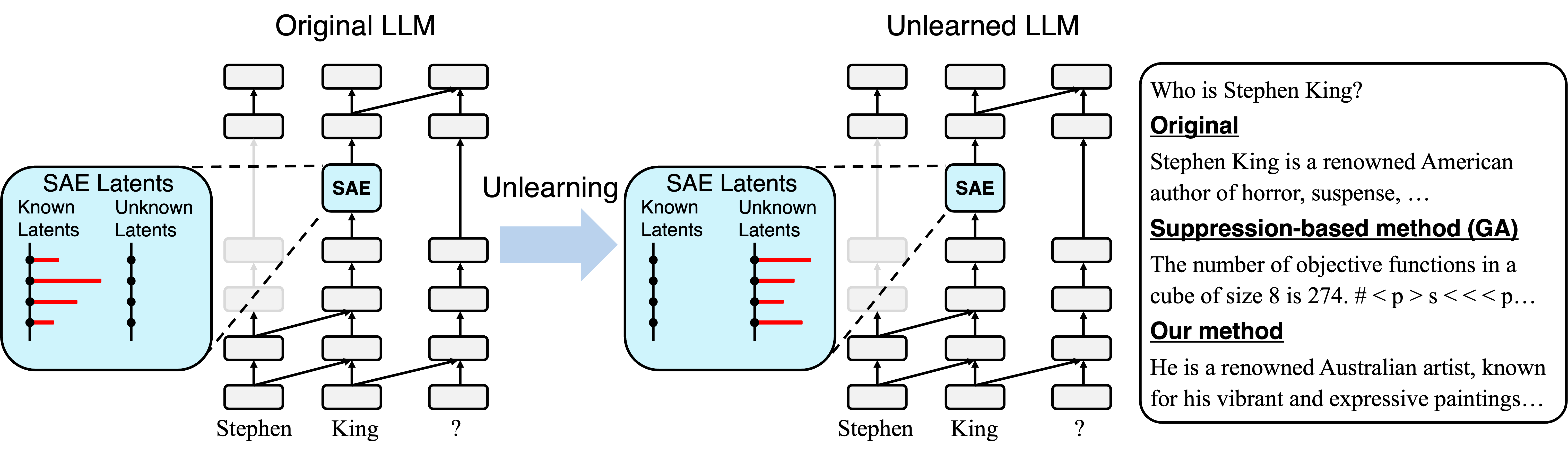}
\caption{\textbf{Overview of our proposal (forgetting ``Stephen King'').}
Known and unknown latents are pre-identified with SAEs.
Our method suppresses activations on the known latents and enhances activations on the unknown latents for the target entity through unlearning.
\textbf{Right:} examples of each LLM's generation. 
Gradient ascent (GA) often exhibits output collapse; in contrast, our method suppresses correct facts about ``Stephen King’’ while preserving coherent and fluent text generation.
}
\label{proposal_overview}
\end{figure*}

\section{Introduction}
\label{introduction}
With the recent performance improvements, large language models (LLMs) are now integrated into various applications such as text generation, data analysis, and specialized domains including medicine, law, and programming~\cite{gpt3,grattafiori2024llama3herdmodels,gemma_2024}.
However, it has been pointed out that the training data for LLMs often contains sensitive or undesirable information, leading to concerns that LLMs may memorize and output undesirable information~\cite{Carlini2020ExtractingTD,DBLP:conf/iclr/CarliniIJLTZ23}.
Furthermore, data protection regulations such as the General Data Protection Regulation (GDPR)~\citep{10.5555/3152676} require model providers to respond to deletion requests on such information promptly.
Addressing these issues is essential for building safe and trustworthy LLM systems. \par
Machine unlearning is one promising approach for addressing these issues, as it seeks to remove the influence of specific training data points from a model \citep{10.1145/3603620}.
In the LLM setting, the focus shifts from example-level unlearning to knowledge-level unlearning, removing knowledge of specific facts or entities acquired from the training dataset~\citep{DBLP:conf/nips/JinCWHYL00024,pmlr-v235-pawelczyk24a}.
Consequently, this knowledge-unlearning setting is expected to be highly interpretable, as the forgotten target aligns with users' intuition.
Among various forms of knowledge unlearning, we focus on entity unlearning, which aims to remove knowledge associated with specific individuals, organizations, locations, or other named entities. \par
While this setting offers intuitive interpretability, it also presents a considerable challenge: the lack of \textbf{an ideal forgotten state}, which would otherwise provide explicit guidance for unlearning.
In theory, this is often represented by an oracle model, i.e., a model trained from scratch without the forgetting target.
However, retraining large models from scratch is prohibitively costly and time-consuming for practical use cases, such as GDPR compliance.
As a result, existing unlearning approaches often rely on suppression-based techniques, which simply reduce the likelihood of specific outputs.
Unfortunately, such suppression-based methods tend to be unstable and cause model collapse~\citep{yao2024large,zhang2024negative}.
In addition, since their objectives focus solely on outputs, it remains unclear whether the underlying target knowledge is truly removed from the internal representations.
In this paper, internal representations refer to the activation patterns within the model that encode some semantic information.
If the target knowledge remains in the internal representations, it may re-emerge in different contexts (e.g., adversarial attacks), thereby undermining the unlearning process.
This concern is also relevant in GDPR, which emphasizes the removal of personal or sensitive data not only from model outputs but also from its parameters~\citep{10.5555/3152676}.
These considerations underscore the erasure of target knowledge from the internal representations, not just from model outputs, for both technical robustness and regulatory compliance. \par
To overcome the lack of an ideal forgotten state, we leverage the model’s internal representations on genuinely unknown entities absent from the training dataset (e.g., fictional entities) as a practical substitute.
This strategy provides a clear objective for the unlearning process, in contrast to suppression-based methods.
Furthermore, by aligning the internal representations of the forgetting target with those of unknown entities, our method enables representation‑level unlearning beyond the output level. \par
Specifically, our method first constructs two entity sets: one comprising entities known to the LLM, and another comprising genuinely unknown ones.
We then project their corresponding activations into a latent space using sparse autoencoders (SAEs).
Then, we identify recognition latents~\citep{ferrando2024know}, which systematically distinguish between known and unknown entities in the SAE latent space.
We define known latents as those frequently activated by known entities, and unknown latents as those more frequently activated by unknown entities.
During the unlearning process, we update the model parameters to suppress the known latents and enhance the unknown latents for the target entity. 
This process shifts the model's recognition of the target entity from ``known'' to ``unknown''.
An overview of our method is shown in Fig.~\ref{proposal_overview}.
\par
Through experiments, we confirm that our method effectively modifies the internal representations of the target entity to those of unknown entities in the SAE latent space.
Furthermore, our method exhibits superior unlearning performance on question-answering tasks, substantially reducing outputs about the target knowledge while preserving non-target responses, thus achieving effective forgetting without compromising overall utility.
Our contributions are as follows:
\begin{itemize}
    \item We propose a novel unlearning method that shifts the model's internal recognition of the target entity from ``known'' to ``unknown'', enabling effective forgetting in the representation space.
    \item We experimentally show that our method outperforms existing baselines on question-answering tasks without significant damage to the non-target knowledge.
    \item We demonstrate that suppression-based methods leave the target’s internal representations largely intact, while ours effectively aligns them with those of unknown entities, achieving unlearning beyond the output level.
\end{itemize}

\section{Background}
\subsection{Entity Unlearning}
\label{entity_unlearning}
In this section, we formalize the entity unlearning problem.
Let $\mathcal{V}$ denote the vocabulary set of the LLM, and $\mathcal{V}^* = \bigcup_{m \in \mathbb{N}} \mathcal{V}^m$ be the set of all finite-length token sequences. 
Let $P_{\mathcal{V}^*}$ be a data distribution over token sequences, i.e., $P_{\mathcal{V}^*} : \mathcal{V}^* \rightarrow [0, 1]$.
We define the LLM's training dataset as $D=\{\bm{x}_i\}_{i=1}^{n}$, where $\bm{x}_i\sim P_{\mathcal{V}^*}$.
We define the entity extraction function $g: \mathcal{V}^* \rightarrow 2^{\mathcal{V}^*}$, which returns a set of subsequences corresponding to entities (e.g., subject, object, named entities) in a given token sequence.
In practice, $g$ can be implemented using off-the-shelf NLP tools such as dependency parsers or named entity recognition (NER) models~\citep{DBLP:conf/emnlp/ChenM14,DBLP:conf/acl/QiZZBM20}.
We define the set of all entities in the training dataset as $E=\bigcup_{i=1}^n g(\bm{x}_i)$.
Let $\bm{e}^\mathrm{(t)}\in E$ be the forgetting target entity.
We then construct the forgetting dataset $D_\mathrm{f}=\{\bm{x}_i|\bm{e}^\mathrm{(t)}\in g(\bm{x}_i)\}$ as the subset of $D$ in which each sample explicitly contains the target entity's name.
While this definition may omit some relevant samples that are implicitly related to the target entity, it provides a precise and noise-free basis for unlearning. \par
Let $\mathcal{A}$ be a randomized training algorithm and $\bm{\theta}_{\mathrm{orig}} \sim \mathcal{A}(D)$ denote a trained model.
In the conventional unlearning setting, the goal is to ensure that the distribution of the unlearned model matches that of an oracle model trained without the forgetting dataset~\citep{DBLP:conf/nips/GinartGVZ19}:
\begin{eqnarray}
\mathcal{U}(\bm{\theta}_\mathrm{orig}, D_\mathrm{f}) = \mathcal{A}(D\setminus D_\mathrm{f}),
\label{exact_unlearning}
\end{eqnarray}
where $\mathcal{U}$ denotes the unlearning algorithm.
Although such exact unlearning offers strong guarantees, it is impractical for LLMs, as it typically requires retraining from scratch, which is infeasible at scale.
As a practical alternative, several LLM unlearning studies adopt a relaxed criterion based on output distribution similarity with an oracle model~\citep{DBLP:conf/nips/KurmanjiTHT23,DBLP:conf/iclr/GeorgievRPGIMN25,DBLP:conf/satml/HayesSTKP25}:
\begin{eqnarray}
    \forall \bm{x}\in D, \Delta(f(\bm{x}; \bm{\theta}_\mathrm{forget}), f(\bm{x};\bm{\theta}_\mathrm{oracle})) \ll 1, 
    \label{mu_requirement}
\end{eqnarray}
where $f$ denotes the LLM's output distribution, $\Delta$ measures the distance between vectors, and $\bm{\theta}_\mathrm{forget}$ and $\bm{\theta}_\mathrm{oracle}\sim\mathcal{A}(D\setminus D_\mathrm{f})$ represent the unlearned and the oracle models, respectively.
While this condition is more tractable compared to exact unlearning in Eq.~\ref{exact_unlearning}, it still assumes access to a hypothetical oracle model representing the ideal forgotten state.
Moreover, output-level unlearning may not fully modify the model's internal representations.
In this work, we instead define a proxy for the ideal forgotten state directly in the internal representation space and optimize the model toward it.

\subsection{Recognition Latents}
\label{recognition_latents}
In this section, we introduce recognition latents~\citep{ferrando2024know} as core components in our method.
Recognition latents are specific dimensions in the SAE latent space.
They represent directions in the model's activation space that are highly indicative of whether a given entity is known or unknown to the LLM.
To derive these latents, we begin by reviewing the forward computation of LLMs, which consists of $L$ transformer layers.
For simplicity, we describe the single-head case, while the approach of \citet{ferrando2024know} can naturally extend to the multi-head case.
The computation within each transformer layer is described as:
\begin{eqnarray}
    \bm{a}_0(\bm{x}; \bm{\theta}) &\leftarrow& \mathrm{input\ embedding}, \\
    \bm{a}_{l}'(\bm{x}; \bm{\theta}) &=& \bm{a}_{l-1}(\bm{x}; \bm{\theta}) + \mathrm{Attention}(\bm{a}_{l-1}(\bm{x}; \bm{\theta})), \\
    \bm{a}_l(\bm{x}; \bm{\theta}) &=& \bm{a}_l'(\bm{x}; \bm{\theta}) + \mathrm{MLP}(\bm{a}_l'(\bm{x}; \bm{\theta})),
\end{eqnarray}
where $\bm{a}_l$ denotes the activation vector at layer $l~(1\leq l \leq L)$, and $\bm{x}$ is the input token sequence.
Then, SAEs are used to project the layer-wise activation $\bm{a}_l(\bm{x}; \bm{\theta})$:
\begin{eqnarray}
    \mathrm{SAE}\left(\bm{a}_l\left(\bm{x}; \bm{\theta}\right)\right) = \bm{z}_\mathrm{post}\left(\bm{a}_l\left(\bm{x}; \bm{\theta}\right)\right)W_\mathrm{dec} + \bm{b}_\mathrm{dec}, 
\end{eqnarray}
where
\begin{eqnarray}
    \bm{z}_\mathrm{post}\left(\bm{a}_l\left(\bm{x}; \bm{\theta}\right)\right) = \sigma \left(\bm{a}_l\left(\bm{x}; \bm{\theta}\right)W_\mathrm{enc} + \bm{b}_\mathrm{enc}\right).
\end{eqnarray}
$W_\mathrm{enc}$, $W_\mathrm{dec}$ and $\bm{b}_\mathrm{enc}$, $\bm{b}_\mathrm{dec}$ denote encoder and decoder parameters, and $\sigma(\cdot)$ denotes the activation function (e.g., JumpReLU~\citep{DBLP:journals/corr/abs-2407-14435}).
The resulting latent vector $\bm{z}_\mathrm{post}$ provides a sparse, disentangled representation of the original activation. \par
To obtain recognition latents, two entity sets are constructed: $E^{\mathrm{(k)}}=\{\bm{e}_i^\mathrm{(k)}\}_{i=1}^{N^{\mathrm{(k)}}}$ for known entities (e.g., ``Harry Potter''), and $E^{\mathrm{(u)}}=\{\bm{e}_i^\mathrm{(u)}\}_{i=1}^{N^{\mathrm{(u)}}}$ for unknown entities (e.g., ``Jaime Vasquez'', a synthetically generated name).
The activations from the last token of each entity are extracted and projected into the SAE latent space. 
The activation frequency of each SAE latent dimension $j$ is calculated as follows:
\begin{eqnarray}
    r_{l, j}^\mathrm{(k)} = \frac{1}{N^\mathrm{(k)}}\sum_{i=1}^{N^\mathrm{(k)}}\mathds{1}[z_{\mathrm{post}, j}(\bm{a}_l(\bm{e}^\mathrm{(k)}_i; \bm{\theta}))>0],
    \label{known_freq}
    \\
    r_{l, j}^\mathrm{(u)} = \frac{1}{N^\mathrm{(u)}}\sum_{i=1}^{N^\mathrm{(u)}}\mathds{1}[z_{\mathrm{post}, j}(\bm{a}_l(\bm{e}^\mathrm{(u)}_i; \bm{\theta}))>0].
    \label{unknown_freq}
\end{eqnarray}
A recognition score is then computed for each latent dimension to quantify its ability to discriminate between known and unknown entities:
\begin{eqnarray}
    s_{l, j}=r_{l, j}^\mathrm{(k)} - r_{l, j}^\mathrm{(u)},
    \label{recognition_score}
\end{eqnarray}
Latent dimensions with high $s_{l, j}$ are selected as known latents, while those with low $s_{l, j}$ are as unknown latents.

\section{Proposed Method}
As discussed in Sec.~\ref{introduction}, a considerable challenge in LLM knowledge unlearning is the lack of an accessible ideal forgotten state, typically represented by an oracle model trained from scratch without the forgetting target.
To address this challenge, we leverage the original model’s internal behavior on genuinely unknown entities absent from the training dataset as a practical substitute:
\begin{eqnarray}
\forall \bm{x}\in D, \bm{a}(\bm{x}; \bm{\theta}_\mathrm{oracle})\simeq \bm{a}(\phi(\bm{x};\bm{e}^\mathrm{(t)}); \bm{\theta}_\mathrm{orig}),
\end{eqnarray}
where $\phi(\bm{x};\bm{e}^\mathrm{(t)})$ denotes a modified version of $\bm{x}$ in which the target entity $\bm{e}^\mathrm{(t)}$ is replaced by an unknown entity (e.g., $\bm{x}=$ ``Who is Harry Potter?'' and $\phi(\bm{x};\bm{e}^\mathrm{(t)})=$ ``Who is Jaime Vasquez?'').
If $\bm{x}$ does not contain the target entity, then $\phi(\bm{x};\bm{e}^\mathrm{(t)}) = \bm{x}$, i.e., no modification is applied.
This formulation allows us to derive an oracle-free and practical alternative to the original requirement in Eq.~\ref{mu_requirement}:
\begin{eqnarray}
    \forall \bm{x}\in D, \Delta(\bm{a}(\bm{x}; \bm{\theta}_\mathrm{forget}), \bm{a}(\phi(\bm{x};\bm{e}^\mathrm{(t)});\bm{\theta}_\mathrm{orig})) \ll 1.
    \label{surrogate_requirement}
\end{eqnarray}
Intuitively, this surrogate criterion requires that the unlearned model’s internal activations for the forgetting target resemble those of unknown entities, making them indistinguishable in the activation space.
Note that we extend the formulation beyond output distribution in Eq.~\ref{mu_requirement} to the internal activations, which encode the LLM's knowledge. \par
Inspired by the recognition latent perspective~\citep{ferrando2024know}, we operationalize this idea by modifying the internal representations for the target entity away from those of known entities and toward those of unknown entities in the SAE latent space.
This allows us to align the activation patterns for the target entity with those of unknown entities within a semantically meaningful representation space.
Our method consists of two steps: (i) identifying recognition latents, and (ii) performing unlearning to steer the model’s internal representations accordingly.

\subsection{Identification of Recognition Latents.}
\label{prepare_recognition_latent}
We identify the recognition latents of the target LLM.
To this end, we prepare a known entity set $E^\mathrm{(k)}$ and an unknown entity set $E^\mathrm{(u)}$.
The known entity set is sampled from a large, in-domain knowledge corpus that is likely to overlap with the LLM’s training dataset (e.g., Wikipedia), while the unknown entity set is sampled from an out-of-distribution corpus that is excluded from pre-training (e.g., the TOFU dataset~\citep{DBLP:journals/corr/abs-2401-06121}).
Using these entity sets, we compute recognition scores in Eq.~\ref{recognition_score}. 
For each layer $1 \leq l \leq L$, we define $\mathcal{S}_l^\mathrm{(k)}$ as the set of known latents whose scores exceed a predefined threshold $\tau$ (i.e., $s_{l,j} > \tau$), and $\mathcal{S}_l^\mathrm{(u)}$ as the set of unknown latents whose scores fall below $-\tau$ (i.e., $s_{l,j} < -\tau$).
These recognition latents capture the activation directions that reflect the model's recognition of whether an entity is known or unknown.
They serve as the basis for manipulating the internal representations during the unlearning phase.
\subsection{Unlearning from Internal Representations}
We leverage the recognition latents to design the unlearning process that steers the internal representations of the forgetting target.
Specifically, we minimize the following loss for the internal activation at each transformer layer:
\begin{eqnarray}
    \mathcal{L}_l (\bm{e}^\mathrm{(t)}; \bm{\theta}) = \sum_{j\in \mathcal{S}_l^\mathrm{(k)}} \max(z_{\mathrm{pre}, j}(\bm{a}_l(\bm{e}^\mathrm{(t)}; \bm{\theta}))+c, 0) \nonumber \\ + \sum_{j\in \mathcal{S}_l^\mathrm{(u)}} \max(-z_{\mathrm{pre}, j}(\bm{a}_l (\bm{e}^\mathrm{(t)}; \bm{\theta}))+c, 0),
    \label{proposal_loss}
\end{eqnarray}
where
\begin{eqnarray}
    \bm{z}_\mathrm{pre}(\bm{a}_l(\bm{e}^\mathrm{(t)}; \bm{\theta})) = \bm{a}_l(\bm{e}^\mathrm{(t)}; \bm{\theta})W_\mathrm{enc} + \bm{b}_\mathrm{enc}.
\end{eqnarray}
Here, $j$ denotes the dimension index of the recognition latents, $l$ denotes the transformer layer index, and $c$ is a hyperparameter that controls the forgetting intensity.
Minimizing this loss encourages the LLM to manipulate its internal representations in a targeted manner: the first term suppresses the positive values of $z_{\mathrm{pre}, j}$ associated with the known latents (i.e., pushing them toward or below $-c$), while the second term promotes the values of $z_{\mathrm{pre}, j}$ associated with the unknown latents (i.e., pulling them upward above $c$).
In effect, this shifts the model's internal representations of the target entity closer to those of unknown, out-of-distribution entities in the SAE latent space.
Following~\citet{ferrando2024know}, we compute the loss using the activation corresponding to the last token of the target entity name, rather than an entire token sequence.
Note also that the input $\bm{x}$ refers to the target entity name itself, not a complete natural language sentence.
We found that minimizing this loss across all layers simultaneously leads to instability.
To mitigate this, our method randomly samples layers and updates them sequentially, ensuring stable convergence. \par
Our method provides two key advantages over existing approaches.
First, it avoids reliance on an oracle model, which is typically required to define an ideal forgotten state.
Instead, our method leverages the original model's internal representations on unknown entities as a practical substitute , enabling a meaningful unlearning objective without requiring an infeasible oracle.
Second, our method offers a stable and goal-directed optimization process.
Suppression-based methods, which lack a concrete objective and typically maximize loss on target sequences, often suffer from instability and model collapse.
In contrast, our method introduces a hinge-style loss function that is bounded and converges naturally as the model's activations shift toward those of the unknown entities in the SAE latent space.
This clear optimization goal avoids instability and model collapse.
We summarize our proposal in Algorithm~\ref{proposal_alg}.

\begin{figure}[tb]
\begin{algorithm}[H]
    \caption{Our proposed algorithm}
    \label{proposal_alg}
    \begin{algorithmic}[1]
    \REQUIRE Target entity: $\bm{e}^\mathrm{(t)}$, Target LLM: $\bm{\theta}$, Known entities: $E^\mathrm{(k)}$, Unknown entities: $E^\mathrm{(u)}$, Forgetting intensity: $c$, Recognition threshold: $\tau$, Learning rate: $\lambda$, Number of layers: $L$
    \STATE $\mathcal{S}_l^\mathrm{(k)} \leftarrow \emptyset, \mathcal{S}_l^\mathrm{(u)} \leftarrow \emptyset (1\leq l\leq L)$
    \FOR{$l$ in [$1\cdots L$]}
    \FOR{$j$ in SAE dimensions}
    \STATE $r_{l, j}^\mathrm{(k)} = \frac{1}{N^\mathrm{(k)}}\sum_{i=1}^{N^\mathrm{(k)}}\mathds{1}[z_{\mathrm{post}, j}(\bm{a}_l(\bm{e}^\mathrm{(k)}_i; \bm{\theta}))>0]$
    \STATE $r_{l, j}^\mathrm{(u)} = \frac{1}{N^\mathrm{(u)}}\sum_{i=1}^{N^\mathrm{(u)}}\mathds{1}[z_{\mathrm{post}, j}(\bm{a}_l(\bm{e}^\mathrm{(u)}_i; \bm{\theta}))>0]$
    \STATE $s_{l, j}=r_{l, j}^\mathrm{(k)} - r_{l, j}^\mathrm{(u)}$
    \IF{$s_{l, j}>\tau$}
    \STATE $\mathcal{S}_l^\mathrm{(k)} \leftarrow \mathcal{S}_l^\mathrm{(k)}\cup \{j\}$
    \ENDIF
    \IF{$s_{l, j}<-\tau$}
    \STATE $\mathcal{S}_l^\mathrm{(u)} \leftarrow \mathcal{S}_l^\mathrm{(u)}\cup \{j\}$
    \ENDIF
    \ENDFOR
    \ENDFOR
    \FOR{$e$ in Epochs}
    \STATE $l \sim \mathcal{U}(\{1, L\})$
    \STATE $
    \begin{aligned}[t]
    &\mathcal{L}_l (\bm{e}^\mathrm{(t)}; \bm{\theta}) = \sum_{j\in \mathcal{S}_l^\mathrm{(k)}} \max(z_{\mathrm{pre}, j}(\bm{a}_l(\bm{e}^\mathrm{(t)}; \bm{\theta}))+c, 0) \\
    &\quad+ \sum_{j\in \mathcal{S}_l^\mathrm{(u)}} \max(-z_{\mathrm{pre}, j}(\bm{a}_l (\bm{e}^\mathrm{(t)}; \bm{\theta}))+c, 0)
    \end{aligned}$
    \STATE $\bm{\theta} \gets \bm{\theta} - \lambda \frac{\partial \mathcal{L}_l}{\partial \bm{\theta}}$
    \ENDFOR
    \RETURN Unlearned LLM $\bm{\theta}$
    \end{algorithmic}
\end{algorithm}
\end{figure}

\section{Experiments}
This section evaluates our method compared to several baselines on a real-world dataset.
Furthermore, we conduct an additional study to evaluate how each unlearning method affects the internal representations.

\subsection{Experimental Setup}
\noindent{\textbf{Dataset:}}
We leverage the Real-World Knowledge Unlearning~(RWKU) dataset~\citep{DBLP:conf/nips/JinCWHYL00024}, which comprises 200 real-world human target entities. 
To evaluate unlearning, we report three forget sub-scores in FB, QA, and AA formats, together with retain and general utility scores. 
\textbf{FB} (fill-in-the-blank) probes test knowledge memorization by masking salient facts in the target's passages and asking the model to complete the blanks.
\textbf{QA} (question–answer) probes test knowledge manipulation by paraphrasing target-related facts into natural questions.
\textbf{AA} (adversarial attack) probes test knowledge manipulation under jailbreak prompts.
We report the test accuracy for each probe, computed as the fraction of instances in which the model output exactly contains the ground truth.
Retain score offers two probes regarding non-target knowledge: FB and QA formats.
Utility score measures six capabilities with five benchmarks, i.e., general ability with MMLU~\citep{DBLP:conf/iclr/HendrycksBBZMSS21} (Gen), reasoning ability with BBH~\citep{DBLP:conf/acl/SuzgunSSGTCCLCZ23} (Rea), truthfulness with TruthfulQA~\citep{lin-etal-2022-truthfulqa} (Tru), factuality with  TriviaQA~\citep{joshi-etal-2017-triviaqa} (Fac), fluency with AlpacaEval~\citep{alpaca_eval} (Flu), and general ability with the perturbed MMLU~\citep{huutien2025improvingllmunlearningrobustness} (Per). \par
\noindent{\textbf{Model:}}
We evaluate our method on two instruction-tuned LLMs: Llama3.1-8B Instruct~\citep{grattafiori2024llama3herdmodels} and Gemma2-9B Instruct~\citep{gemma_2024}, both of which come with SAE suites, i.e., Llama-Scope~\citep{DBLP:journals/corr/abs-2410-20526} and Gemma-Scope~\citep{DBLP:journals/corr/abs-2408-05147}, respectively. \par
\noindent{\textbf{Baselines:}}
We include two suppression-based methods and one representation-level method.
As suppression-based baselines, we adopt gradient ascent (GA)~\citep{jang-etal-2023-knowledge} and negative preference optimization (NPO)~\citep{zhang2024negative}. 
As a representation-level baseline, we adopt representation misdirection for unlearning (RMU)~\citep{pmlr-v235-li24bc}, which aligns activations for target prompts with randomly sampled directions.
These baselines are applied to either the explanatory corpus in the RWKU dataset, the full name of the target entity, or just its last token. 
We apply early stopping based on the retain score for all methods, including our proposed approach. 
The process monitors performance every 10 epochs up to a maximum of 200, and selects the last checkpoint before the retain score degrades by more than 10\%.
This strategy enables us to evaluate each method’s forgetting performance under the practical constraint of maintaining the model's utility. \par
In our method, we fix the forgetting intensity $c=1.0$ in Eq.~\ref{proposal_loss}, and the recognition scores threshold $\tau=0.4$ in Sec.~\ref{prepare_recognition_latent}.
These hyperparameters are selected based on preliminary experiments and are kept constant across all experiments.
Further details on the experimental setup are provided in Appendix~\ref{experimental_setup}.

\subsection{Evaluation: Unlearning Performance}
Table~\ref{main_result} presents the unlearning performance on 100 entities in the RWKU dataset. 
Across both models, our method consistently achieves the lowest forget scores, indicating superior performance in removing target knowledge. 
Specifically, for Llama3.1-8B Instruct, our method achieves a mean forget score of 46.8\%, representing a significant improvement over all baseline methods. 
Similarly, on Gemma2-9B Instruct, our method achieves a mean forget score of 57.1\%, outperforming other approaches by a substantial margin.
Importantly, our method achieves strong selective forgetting without significant damage to the non-target knowledge. \par
By contrast, baseline methods, including suppression-based approaches such as GA and NPO, struggle to forget the target knowledge without collateral damage to non-target knowledge.
Although Table~\ref{main_result} does not show significant degradation in retain and utility scores in GA and NPO, this is primarily due to early stopping.
In other words, attempts to further reduce the forget score inevitably lead to a substantial drop in the retain and utility scores, indicating that these methods fail to achieve selective forgetting.
Similarly, RMU, which attempts to forget by steering the LLM's internal activations in random directions, is less effective than our method.
Although RMU is not suppression-based, its forgetting mechanism lacks a semantically meaningful reference, which explains its inferior performance.
In contrast, our method leverages a structured latent space, guided by recognition latents corresponding to known and unknown entities, enabling more targeted and stable unlearning.

\begin{table*}[tb]
\small
\centering
\caption{\textbf{Main results on the RWKU dataset.}
We report unlearning performance across two LLMs (Llama3.1-8B Instruct and Gemma2-9B Instruct) across 100 entities in the RWKU dataset.
Forget score ($\downarrow$) measures how well the model forgets target entities, retain score ($\uparrow$) evaluates the preservation of non-target knowledge, and utility score ($\uparrow$) reflects general language capabilities across six axes (General, Reasoning, Truthfulness, Factuality, Fluency, and Perturbed General).
The best forget score in each column is shown in bold.
Note that the differences in retain and utility scores among methods are very limited due to early stopping based on the retain score.}
\label{main_result}
    \begin{tabular}{c|c|ccccccccccccc}
    \toprule
    \multirow{2}{*}{LLM}  & \multirow{2}{*}{Method} & \multicolumn{4}{c|}{Forget score $\downarrow$} & \multicolumn{3}{c|}{Retain score $\uparrow$} & \multicolumn{6}{c}{Utility score $\uparrow$} \\
    \cmidrule{3-15}
    & & FB & QA & AA & \multicolumn{1}{c|}{All} & FB & QA & \multicolumn{1}{c|}{All} & Gen & Rea & Tru & Fac & Flu & Per \\
    \midrule
    \multirow{11}{*}{Llama3.1-8B} & Original & 80.7 & 80.3 & 82.4 & \multicolumn{1}{c|}{81.1} & 82.6 & \textbf{85.1} & \multicolumn{1}{c|}{83.9} & 65.9 & \textbf{43.1} & 36.5 & \textbf{68.9} & \textbf{724.7} & 67.0 \\
    & GA (sentence) & 68.0 & 66.5 & 72.9 & \multicolumn{1}{c|}{69.1} & 82.3 & 83.4 & \multicolumn{1}{c|}{82.9} & 65.4 & 42.4 & 36.5 & 67.4 & 721.0 & 66.6 \\
    & GA (entity) & 72.1 & 67.4 & 64.3 & \multicolumn{1}{c|}{67.9} & 83.5 & 83.7 & \multicolumn{1}{c|}{83.6} & 64.7 & 42.4 & 35.4 & 66.7 & 703.7 & 65.9  \\
    & GA (last token) & 71.2 & 66.7 & 58.6 & \multicolumn{1}{c|}{65.5} & 82.6 & 85.0 & \multicolumn{1}{c|}{83.8} & 65.9 & 42.7 & \textbf{36.8} & 67.3 & 721.5 & \textbf{67.1}  \\
    & NPO (sentence) & 67.7 & 67.2 & 72.9 & \multicolumn{1}{c|}{69.3} & 82.6 & 84.7 & \multicolumn{1}{c|}{83.7} & 65.5 & 42.6 & 36.1 & 68.0 & 722.5 & 66.9 \\
    & NPO (entity) & 75.7 & 72.2 & 72.1 & \multicolumn{1}{c|}{73.3} & 83.1 & 84.2 & \multicolumn{1}{c|}{83.7} & 64.9 & 42.5 & 35.0 & 67.1 & 715.8 & 66.4 \\
    & NPO (last token) & 77.4 & 75.1 & 73.7 & \multicolumn{1}{c|}{75.4} & 83.2 & 84.9 & \multicolumn{1}{c|}{\textbf{84.1}} & \textbf{66.1} & 43.0 & 36.6 & 68.3 & 722.4 & 67.0 \\
    & RMU (sentence) & 78.2 & 78.8 & 80.5 & \multicolumn{1}{c|}{79.2} & 82.9 & 84.9 & \multicolumn{1}{c|}{83.9} & \textbf{66.1} & 42.9 & 36.2 & 67.0 & 721.5 & 66.9 \\
    & RMU (entity) & 75.1 & 72.2 & 76.0 & \multicolumn{1}{c|}{74.4} & 83.2 & 83.2 & \multicolumn{1}{c|}{83.2} & 65.4 & 42.5 & 35.7 & 61.7 & 713.8 & 66.2 \\
    & RMU (last token) & 56.9 & 49.4 & 64.5 & \multicolumn{1}{c|}{56.9} & \textbf{83.6} & 84.0 & \multicolumn{1}{c|}{83.8} & 65.5 & 42.6 & 36.5 & 66.8 & 720.5 & 67.0 \\
    & Ours & \textbf{46.2} & \textbf{37.9} & \textbf{56.3} & \multicolumn{1}{c|}{\textbf{46.8}} & 83.1 & 84.0 & \multicolumn{1}{c|}{83.5} & 65.1 & 42.9 & 36.1 & 65.8 & 710.4 & 67.0 \\
    \midrule
    \multirow{11}{*}{Gemma2-9B} & Original & 84.6 & 77.4 & 78.2 & \multicolumn{1}{c|}{80.1} & \textbf{87.3} & \textbf{80.0} & \multicolumn{1}{c|}{\textbf{83.7}} & \textbf{72.8} & 33.9 & 38.1 & \textbf{67.5} & 644.7 & \textbf{72.1} \\
    & GA (sentence) & 79.0 & 73.0 & 74.0 & \multicolumn{1}{c|}{75.3} & 86.5 & 79.3 & \multicolumn{1}{c|}{82.9} & 72.3 & 31.1 & 38.2 & 66.9 & \textbf{671.6} & 71.4 \\
    & GA (entity) & 81.2 & 71.1 & 66.9 & \multicolumn{1}{c|}{73.1} & 86.8 & 78.4 & \multicolumn{1}{c|}{82.6} & 72.6 & 24.5 & 38.2 & 63.5 & 540.7 & 71.8 \\
    & GA (last token) & 79.6 & 70.0 & 65.3 & \multicolumn{1}{c|}{71.6} & 87.2 & 79.1 & \multicolumn{1}{c|}{83.2} & 72.7 & 26.6 & 37.9 & 65.2 & 567.6 & 71.9 \\
    & NPO (sentence) & 77.9 & 70.6 & 73.3 & \multicolumn{1}{c|}{73.9} & 86.7 & 79.6 & \multicolumn{1}{c|}{83.2} & 72.4 & 27.5 & 37.9 & 66.2 & 655.8 & 71.7 \\
    & NPO (entity) & 81.2 & 74.4 & 73.8 & \multicolumn{1}{c|}{76.5} & 87.0 & 79.8 & \multicolumn{1}{c|}{83.4} & 72.7 & 30.3 & 37.7 & 66.6 & 646.6 & 72.0 \\
    & NPO (last token) & 78.8 & 72.9 & 72.5 & \multicolumn{1}{c|}{74.7} & 85.7 & 78.9 & \multicolumn{1}{c|}{82.3} & 71.8 & 30.2 & 37.6 & 65.2 & 652.8 & 71.3 \\
    & RMU (sentence) & 76.8 & 72.6 & 73.2 & \multicolumn{1}{c|}{74.2} & 85.6 & 77.1 & \multicolumn{1}{c|}{81.4} & 65.1 & 34.8 & 37.5 & 61.9 & 629.0 & 65.5 \\
    & RMU (entity) & 77.3 & 70.0 & 74.5 & \multicolumn{1}{c|}{73.9} & 86.3 & 78.7 & \multicolumn{1}{c|}{82.5} & 63.4 & 36.3 & 37.9 & 60.7 & 622.3 & 63.6 \\
    & RMU (last token) & 78.2 & 73.4 & 73.8 & \multicolumn{1}{c|}{75.1} & 85.5 & 78.4 & \multicolumn{1}{c|}{82.0} & 64.2 & \textbf{36.6} & 37.9 & 60.1 & 626.5 & 64.9 \\
    & Ours & \textbf{60.4} & \textbf{48.2} & \textbf{62.6} & \multicolumn{1}{c|}{\textbf{57.1}} & 85.6 & 78.8 & \multicolumn{1}{c|}{82.2} & 70.9 & 30.4 & \textbf{38.9} & 64.2 & 635.5 & 69.5 \\
    \bottomrule
    \end{tabular}
\end{table*}

\subsection{Evaluation: Effects on Internal Representations}
In this section, we evaluate how each unlearning method affects the internal representations.
To assess their impact when the output of the forgotten target is fully suppressed, we evaluate the models at the checkpoint where the forget score reaches 0\%~\footnote{In some cases, forget score does not reach 0\%. 
We use the final checkpoint obtained after 200 epochs.
}.
Due to space limitations, we report results on the Llama3.1-9B Instruct model with GA as the representative baseline.
Additional results for NPO, RMU, and the Gemma2-9B Instruct are provided in Appendix~\ref{full_results}. \par
\noindent{\textbf{Effects on Attribute Rate:}}
We assess how much information about the forgotten target remains in the model's activations after unlearning.
Following~\citet{geva-etal-2023-dissecting}, we adopt the attribute rate as an automatic proxy for how many target-related tokens are still encoded within the model's activations.
For the target entity $\bm{e}^\mathrm{(t)}$, we construct a set $\mathcal{A}^\mathrm{(t)}$ of target-related tokens retrieved from passages in the RWKU dataset.
We then compute the activation corresponding to the last token of $\bm{e}^\mathrm{(t)}$ and project it into the token distribution using the embedding matrix.
Then, we obtain its top-50 projected tokens.
The attribute rate is calculated as the fraction of these top-50 tokens that fall inside $\mathcal{A}^\mathrm{(t)}$.
The details of the attribute rate are described in Appendix~\ref{attribute_rate}. \par
Figure~\ref{attr_rate} illustrates the results, revealing a clear trend in how different unlearning methods affect the attribute rate.
All GA-based methods leave the attribute rate virtually unchanged, suggesting that these methods fail to remove the target entity's footprint from the activations.
In contrast, our method substantially reduces the attribute rate, particularly across the middle and deeper layers. 

\begin{figure}[tb]
\centering
\includegraphics[width=0.42\textwidth]{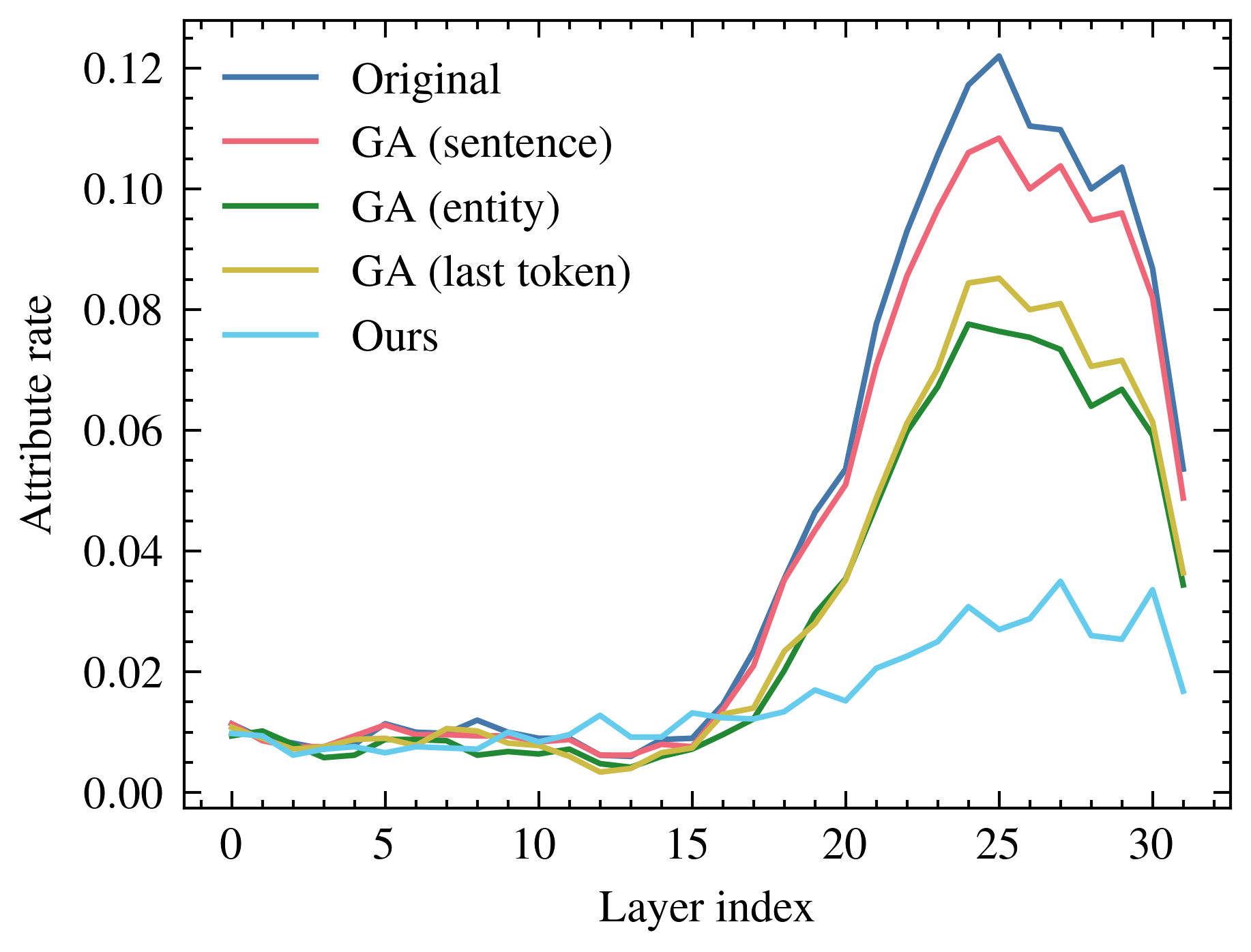}
\caption{\textbf{Attribute rate on Llama3.1-8B Instruct.} 
We show the average attribute rate over 20 target entities.
Results are reported for the original model, three GA-based baselines, and our method.
}
\label{attr_rate}
\end{figure}

\noindent{\textbf{Effects on Recognition Latents:}}
We evaluate how each unlearning method affects the activation of recognition latents.
Recognition latents are identified using the procedure described in Sec~\ref{recognition_latents}.
Specifically, for each transformer layer, we select the top-5 known and unknown latents, as determined by their recognition scores in Eq.~\ref{recognition_score}.
We report the average activation frequencies of these latents when 20 forgotten entities are input to each unlearned model. \par
\begin{figure*}[tb]
\centering
\begin{minipage}{0.42\textwidth}
\centering
\includegraphics[width=\textwidth]{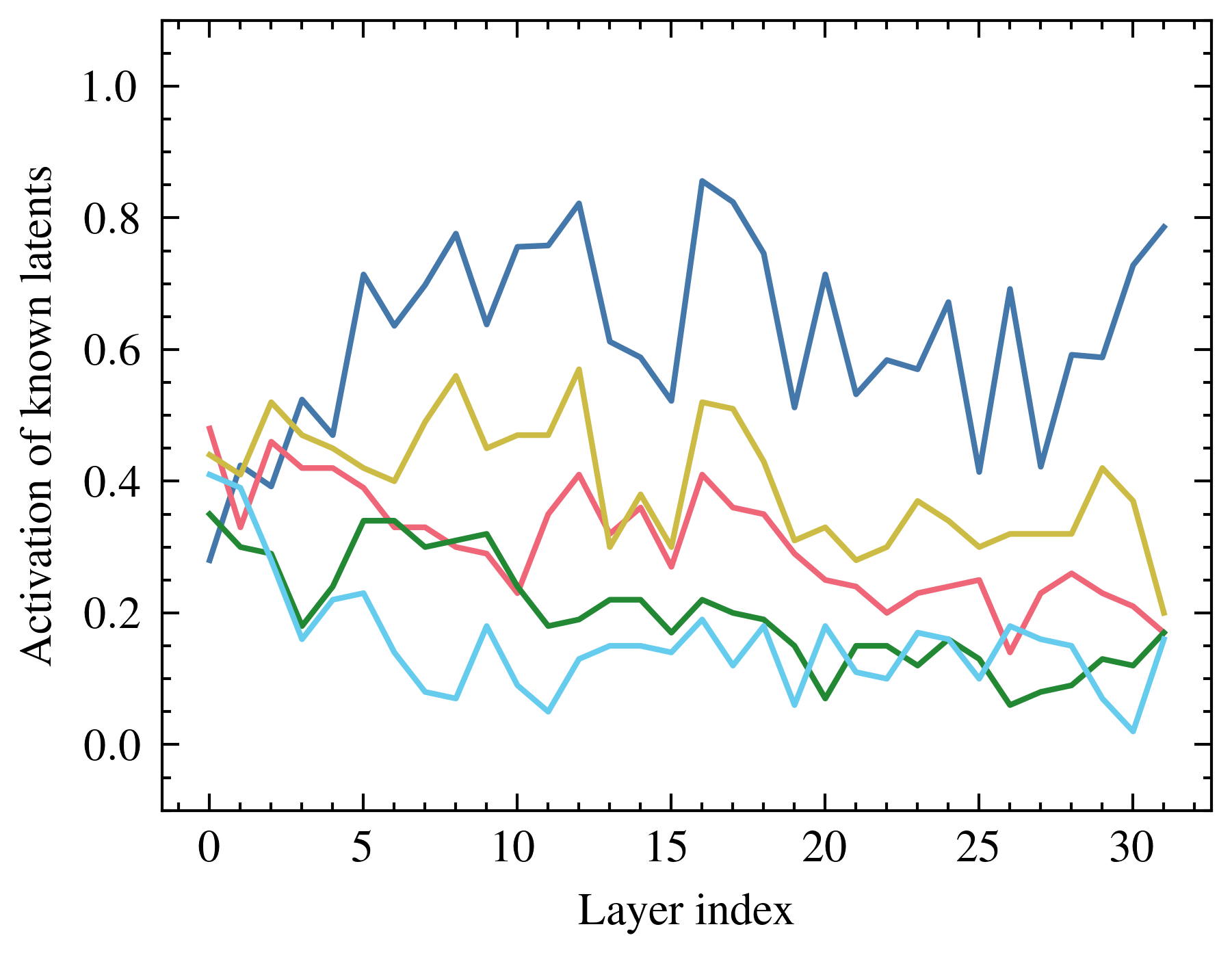}
\subcaption{Results for known latents.}
\end{minipage}
\hspace{0.06\textwidth}   
\begin{minipage}{0.42\textwidth}
\centering
\includegraphics[width=\textwidth]{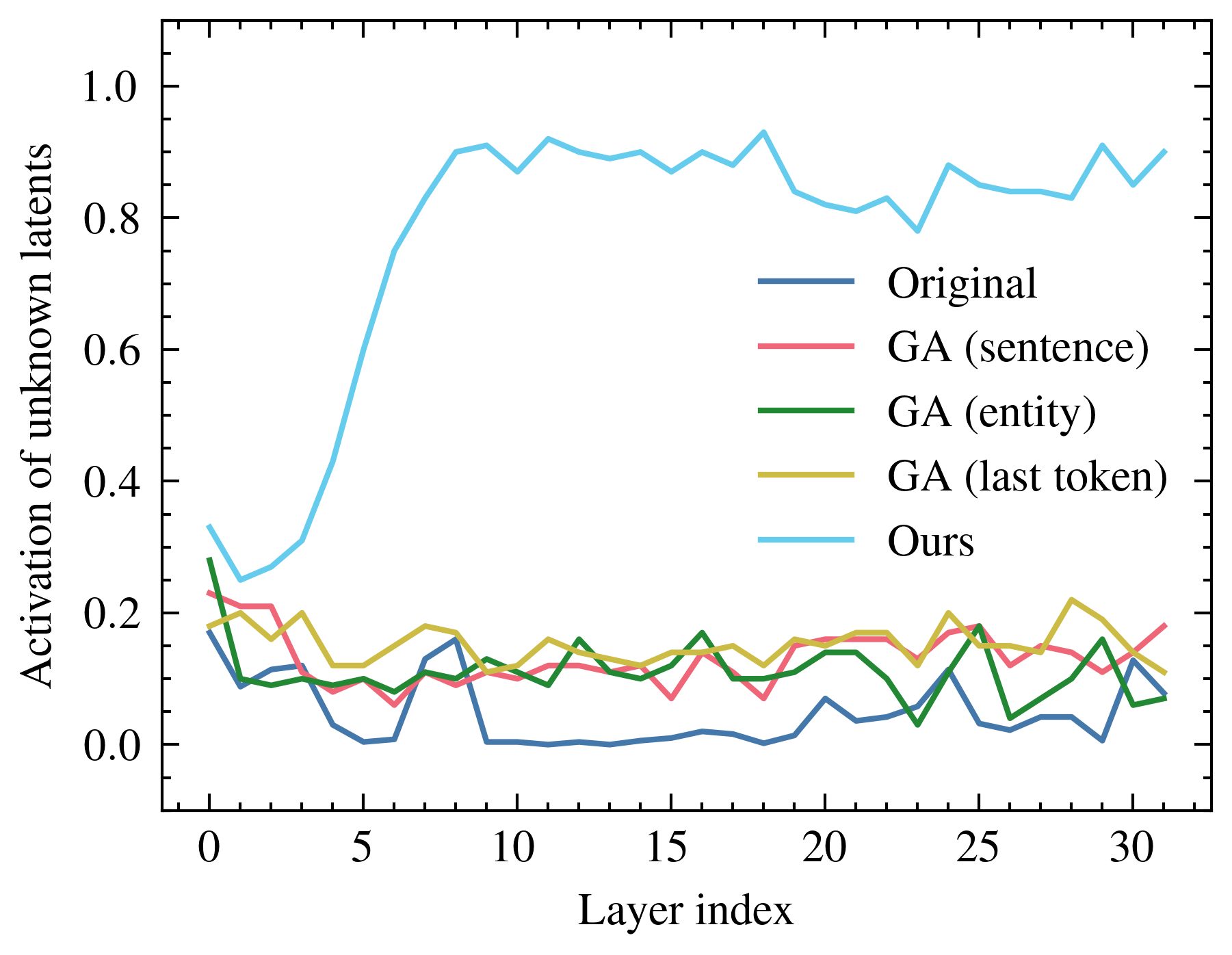}
\subcaption{Results for unknown latents.}
\end{minipage}
\caption{\textbf{Activation frequency of recognition latents.} We report the average activation frequency of recognition latents on Llama3.1-8B-Instruct in Eq.~\ref{known_freq} and~\ref{unknown_freq}.
We aggregate over 20 target entities in the RWKU dataset.
We compare the original model with several GA-based unlearning configurations and our method.}
\label{rec_latents_fire}
\end{figure*}

Figure~\ref{rec_latents_fire} shows the average activation frequencies of the recognition latents.
The results indicate that the GA-based methods achieve some degree of suppression on known latents but fail to enhance the unknown latents.
As a result, the internal representations of the target entities do not sufficiently align with those of unknown entities, implying that the model still recognizes them internally.
In contrast, our method not only suppresses the known latents but also enhances the unknown latents.
Note that the figures show that differences between methods are relatively small in the shallow layers. 
This observation aligns with the findings of~\citet{ferrando2024know}, which suggests that recognition latents are prominent in deeper layers.
Overall, these findings demonstrate that suppression-based methods are insufficient for thorough unlearning at the representational level, whereas our method effectively achieves this.

\section{Related Works}
In this section, we briefly review related works on machine unlearning and mechanistic interpretability, which respectively ground our unlearning formulation and the use of internal representations via SAEs.

\subsection{Machine Unlearning}
Machine unlearning aims to remove the effects of arbitrary training data points from a model~\citep{bourtoule2021machine,DBLP:conf/nips/GinartGVZ19,DBLP:conf/alt/Neel0S21}.
\citet{DBLP:conf/nips/GinartGVZ19} formalized exact unlearning as Eq.~\ref{exact_unlearning}, but it is generally infeasible because it typically requires retraining from scratch, which is infeasible at scale.
This motivates approximate unlearning~\citep{DBLP:conf/icml/GuoGHM20}, which requires only that the distribution over unlearned models be $(\epsilon, \delta)$-indistinguishable from retraining.
Let $\mathcal{H}$ be the hypothesis space.
Approximate unlearning holds if for all measurable subsets $\mathcal{T} \subseteq \mathcal{H}$, the following conditions are satisfied:
\begin{eqnarray}
    \mathrm{Pr}\left[\mathcal{U}(\bm{\theta}_\mathrm{orig}, D_\mathrm{f})\in \mathcal{T}\right] \leq e^\epsilon \mathrm{Pr}\left[\mathcal{A}(D\setminus D_\mathrm{f})\in \mathcal{T}\right]+\delta, \\
    \mathrm{Pr}\left[\mathcal{A}(D\setminus D_\mathrm{f})\in \mathcal{T}\right] \leq e^\epsilon\mathrm{Pr}\left[\mathcal{U}(\bm{\theta}_\mathrm{orig}, D_\mathrm{f})\in \mathcal{T}\right]+\delta, 
\end{eqnarray}
where the parameters $\epsilon$ and $\delta$ quantify the level of approximation allowed in the unlearning. \par
Recent works extend unlearning to LLMs~\citep{yao2024large,zhang2024negative,pmlr-v235-li24bc,pmlr-v235-pawelczyk24a}.
\citet{yao2024large} suppressed harmful or copyrighted contents by additional training.
\citet{zhang2024negative} introduced NPO, which improves stability and efficiency compared to naive GA unlearning.
\citet{pmlr-v235-li24bc} proposed RMU, which suppresses harmful content by steering internal activations away from hazardous content.
\citet{pmlr-v235-pawelczyk24a} presented an in-context unlearning that works at inference time and requires no parameter updates.
Beyond individual examples, some works aim to forget factual knowledge~\citep{DBLP:conf/nips/JinCWHYL00024,jang-etal-2023-knowledge}.
\citet{DBLP:conf/nips/JinCWHYL00024} constructs a real-world knowledge unlearning (RWKU) benchmark for evaluating LLM knowledge unlearning.
\citet{jang-etal-2023-knowledge} leverages GA to unlearn entity-related knowledge.
To our knowledge, prior work does not specify an explicit, representation-level proxy for the ideal forgotten state; we provide one by leveraging unknown-entity internal representations identified with SAEs.

\subsection{Mechanistic Interpretability}
Mechanistic interpretability aims to reveal the relationship between a model's internal representations and its outputs.
\citet{nanda2023factfinding} found that LLMs represent entities and facts via token-specific attention heads and aggregation mechanisms.
Other work demonstrated that factual knowledge can be traced to early MLP layers and manipulated via localized activation edits~\citep{DBLP:conf/nips/MengBAB22}.
\citet{geva-etal-2022-transformer} reverse-engineered transformer MLP layers, showing that their outputs decompose into human-interpretable updates.
\citet{geva-etal-2023-dissecting} investigated how LLMs retrieve factual attributes by intervening on attention edges.
Furthermore, \citet{DBLP:conf/iclr/WangVCSS23} reverse-engineered a full attention circuit for indirect object identification, showing the interpretability of specific linguistic behavior.
However, many neurons in LLMs are polysemantic, simultaneously encoding multiple concepts.
\citet{DBLP:conf/iclr/HubenCRES24} leveraged SAEs to disentangle such polysemantic representations.
\citet{ferrando2024know} demonstrated that SAEs can separate known and unknown entities in latent space, which we leverage in our method for representation-level unlearning.

\section{Limitations}
Our method does not achieve a forget score of zero in the main results.
Suppression-based methods can further reduce it when early stopping is disabled, but this often results in severe damage to non-target knowledge.
Under realistic constraints, preserving retain and utility scores, our method yields the lowest forget score, demonstrating its practical advantage.
Additionally, our evaluation focuses only on individuals, both as target entities and as references for recognition latents.
Extending the method to other entity types (e.g., organizations or locations) is left for future work.
Finally, our oracle-free objective relies on substituting the target entity with unknown entities.
This proxy does not rigorously replicate a true out-of-distribution state of an oracle model.
Nonetheless, this surrogate approach is widely used in recent studies~\citep{eldan2023whosharrypotterapproximate,pmlr-v235-pawelczyk24a}, and we mitigate its limitations by aligning activations in an interpretable SAE latent space.

\section{Conclusion}
This paper presents a novel LLM unlearning method that leverages recognition latents by steering activations in a sparse autoencoder latent space.
Unlike suppression-based approaches, our method directly intervenes in the model’s representations, steering them away from those of known entities and toward those of genuinely unknown ones.
Our method enables stable unlearning by leveraging a practical surrogate for the ideal forgotten state with the model’s internal representations.
On the RWKU benchmark, our approach achieves a superior balance between forgetting effectiveness and model utility compared to existing baselines.
Moreover, we confirm that our method enables the model to forget from the internal representations, a task that the suppression-based baselines can not achieve.
By re-framing unlearning as representation-level erasure rather than output suppression, our work offers a new direction for LLM unlearning research.
We hope this contributes to a deeper integration of mechanistic interpretability techniques into the design and evaluation of unlearning algorithms, leading to more trustworthy and compliant LLM systems in the future.

\bibliography{aaai2026}

\begin{thebibliography}{41}
\providecommand{\natexlab}[1]{#1}

\bibitem[{Bourtoule et~al.(2021)Bourtoule, Chandrasekaran, Choquette-Choo, Jia, Travers, Zhang, Lie, and Papernot}]{bourtoule2021machine}
Bourtoule, L.; Chandrasekaran, V.; Choquette-Choo, C.~A.; Jia, H.; Travers, A.; Zhang, B.; Lie, D.; and Papernot, N. 2021.
\newblock Machine unlearning.
\newblock In \emph{2021 IEEE Symposium on Security and Privacy (SP)}, 141--159. IEEE.

\bibitem[{Brown et~al.(2020)Brown, Mann, Ryder, Subbiah, Kaplan, Dhariwal, Neelakantan, Shyam, Sastry, Askell, Agarwal, Herbert{-}Voss et~al.}]{gpt3}
Brown, T.~B.; Mann, B.; Ryder, N.; Subbiah, M.; Kaplan, J.; Dhariwal, P.; Neelakantan, A.; Shyam, P.; Sastry, G.; Askell, A.; Agarwal, S.; Herbert{-}Voss, A.; et~al. 2020.
\newblock Language Models are Few-Shot Learners.
\newblock \emph{CoRR}, abs/2005.14165.

\bibitem[{Carlini et~al.(2023)Carlini, Ippolito, Jagielski, Lee, Tram{\`{e}}r, and Zhang}]{DBLP:conf/iclr/CarliniIJLTZ23}
Carlini, N.; Ippolito, D.; Jagielski, M.; Lee, K.; Tram{\`{e}}r, F.; and Zhang, C. 2023.
\newblock Quantifying Memorization Across Neural Language Models.
\newblock In \emph{The Eleventh International Conference on Learning Representations, {ICLR} 2023}. OpenReview.net.

\bibitem[{Carlini et~al.(2020)Carlini, Tram{\`e}r, Wallace, Jagielski, Herbert-Voss, Lee, Roberts, Brown, Song, Erlingsson, Oprea, and Raffel}]{Carlini2020ExtractingTD}
Carlini, N.; Tram{\`e}r, F.; Wallace, E.; Jagielski, M.; Herbert-Voss, A.; Lee, K.; Roberts, A.; Brown, T.~B.; Song, D.~X.; Erlingsson, {\'U}.; Oprea, A.; and Raffel, C. 2020.
\newblock Extracting Training Data from Large Language Models.
\newblock In \emph{USENIX Security Symposium}.

\bibitem[{Chen and Manning(2014)}]{DBLP:conf/emnlp/ChenM14}
Chen, D.; and Manning, C.~D. 2014.
\newblock A Fast and Accurate Dependency Parser using Neural Networks.
\newblock In Moschitti, A.; Pang, B.; and Daelemans, W., eds., \emph{Proceedings of the 2014 Conference on Empirical Methods in Natural Language Processing, {EMNLP} 2014, October 25-29, 2014, Doha, Qatar, {A} meeting of SIGDAT, a Special Interest Group of the {ACL}}, 740--750. {ACL}.

\bibitem[{Eldan and Russinovich(2023)}]{eldan2023whosharrypotterapproximate}
Eldan, R.; and Russinovich, M. 2023.
\newblock Who's Harry Potter? Approximate Unlearning in LLMs.
\newblock \emph{CoRR}, abs/2310.02238.

\bibitem[{Ferrando et~al.(2024)Ferrando, Obeso, Rajamanoharan, and Nanda}]{ferrando2024know}
Ferrando, J.; Obeso, O.; Rajamanoharan, S.; and Nanda, N. 2024.
\newblock Do I Know This Entity? Knowledge Awareness and Hallucinations in Language Models.
\newblock \emph{arXiv preprint arXiv:2411.14257}.

\bibitem[{Georgiev et~al.(2025)Georgiev, Rinberg, Park, Garg, Ilyas, Madry, and Neel}]{DBLP:conf/iclr/GeorgievRPGIMN25}
Georgiev, K.; Rinberg, R.; Park, S.~M.; Garg, S.; Ilyas, A.; Madry, A.; and Neel, S. 2025.
\newblock Machine Unlearning via Simulated Oracle Matching.
\newblock In \emph{The Thirteenth International Conference on Learning Representations, {ICLR} 2025, Singapore, April 24-28, 2025}. OpenReview.net.

\bibitem[{Geva et~al.(2023)Geva, Bastings, Filippova, and Globerson}]{geva-etal-2023-dissecting}
Geva, M.; Bastings, J.; Filippova, K.; and Globerson, A. 2023.
\newblock Dissecting Recall of Factual Associations in Auto-Regressive Language Models.
\newblock In Bouamor, H.; Pino, J.; and Bali, K., eds., \emph{Proceedings of the 2023 Conference on Empirical Methods in Natural Language Processing}, 12216--12235. Singapore: Association for Computational Linguistics.

\bibitem[{Geva et~al.(2022)Geva, Caciularu, Wang, and Goldberg}]{geva-etal-2022-transformer}
Geva, M.; Caciularu, A.; Wang, K.; and Goldberg, Y. 2022.
\newblock Transformer Feed-Forward Layers Build Predictions by Promoting Concepts in the Vocabulary Space.
\newblock In Goldberg, Y.; Kozareva, Z.; and Zhang, Y., eds., \emph{Proceedings of the 2022 Conference on Empirical Methods in Natural Language Processing}, 30--45. Abu Dhabi, United Arab Emirates: Association for Computational Linguistics.

\bibitem[{Ginart et~al.(2019)Ginart, Guan, Valiant, and Zou}]{DBLP:conf/nips/GinartGVZ19}
Ginart, A.; Guan, M.~Y.; Valiant, G.; and Zou, J. 2019.
\newblock Making {AI} Forget You: Data Deletion in Machine Learning.
\newblock In Wallach, H.~M.; Larochelle, H.; Beygelzimer, A.; d'Alch{\'{e}}{-}Buc, F.; Fox, E.~B.; and Garnett, R., eds., \emph{Advances in Neural Information Processing Systems 32: Annual Conference on Neural Information Processing Systems 2019, NeurIPS 2019, December 8-14, 2019, Vancouver, BC, Canada}, 3513--3526.

\bibitem[{Grattafiori et~al.(2024)Grattafiori, Dubey, Jauhri, Pandey, Kadian, Al-Dahle, Letman, Mathur, Schelten, Vaughan, Yang, Fan et~al.}]{grattafiori2024llama3herdmodels}
Grattafiori, A.; Dubey, A.; Jauhri, A.; Pandey, A.; Kadian, A.; Al-Dahle, A.; Letman, A.; Mathur, A.; Schelten, A.; Vaughan, A.; Yang, A.; Fan, A.; et~al. 2024.
\newblock The Llama 3 Herd of Models.
\newblock arXiv:2407.21783.

\bibitem[{Guo et~al.(2020)Guo, Goldstein, Hannun, and van~der Maaten}]{DBLP:conf/icml/GuoGHM20}
Guo, C.; Goldstein, T.; Hannun, A.~Y.; and van~der Maaten, L. 2020.
\newblock Certified Data Removal from Machine Learning Models.
\newblock In \emph{Proceedings of the 37th International Conference on Machine Learning, {ICML} 2020, 13-18 July 2020, Virtual Event}, volume 119 of \emph{Proceedings of Machine Learning Research}, 3832--3842. {PMLR}.

\bibitem[{Hayes et~al.(2025)Hayes, Shumailov, Triantafillou, Khalifa, and Papernot}]{DBLP:conf/satml/HayesSTKP25}
Hayes, J.; Shumailov, I.; Triantafillou, E.; Khalifa, A.; and Papernot, N. 2025.
\newblock Inexact Unlearning Needs More Careful Evaluations to Avoid a False Sense of Privacy.
\newblock In \emph{{IEEE} Conference on Secure and Trustworthy Machine Learning, SaTML 2025, Copenhagen, Denmark, April 9-11, 2025}, 497--519. {IEEE}.

\bibitem[{He et~al.(2024)He, Shu, Ge, Chen, Wang, Zhou, Liu, Guo, Huang, Wu, Jiang, and Qiu}]{DBLP:journals/corr/abs-2410-20526}
He, Z.; Shu, W.; Ge, X.; Chen, L.; Wang, J.; Zhou, Y.; Liu, F.; Guo, Q.; Huang, X.; Wu, Z.; Jiang, Y.; and Qiu, X. 2024.
\newblock Llama Scope: Extracting Millions of Features from Llama-3.1-8B with Sparse Autoencoders.
\newblock \emph{CoRR}, abs/2410.20526.

\bibitem[{Hendrycks et~al.(2021)Hendrycks, Burns, Basart, Zou, Mazeika, Song, and Steinhardt}]{DBLP:conf/iclr/HendrycksBBZMSS21}
Hendrycks, D.; Burns, C.; Basart, S.; Zou, A.; Mazeika, M.; Song, D.; and Steinhardt, J. 2021.
\newblock Measuring Massive Multitask Language Understanding.
\newblock In \emph{9th International Conference on Learning Representations, {ICLR} 2021, Virtual Event, Austria, May 3-7, 2021}. OpenReview.net.

\bibitem[{Huben et~al.(2024)Huben, Cunningham, Smith, Ewart, and Sharkey}]{DBLP:conf/iclr/HubenCRES24}
Huben, R.; Cunningham, H.; Smith, L.~R.; Ewart, A.; and Sharkey, L. 2024.
\newblock Sparse Autoencoders Find Highly Interpretable Features in Language Models.
\newblock In \emph{The Twelfth International Conference on Learning Representations, {ICLR} 2024, Vienna, Austria, May 7-11, 2024}. OpenReview.net.

\bibitem[{Huu-Tien et~al.(2025)Huu-Tien, Thanh-Tung, Bui, Nguyen, and Inoue}]{huutien2025improvingllmunlearningrobustness}
Huu-Tien, D.; Thanh-Tung, H.; Bui, A.; Nguyen, L.-M.; and Inoue, N. 2025.
\newblock Improving LLM Unlearning Robustness via Random Perturbations.
\newblock arXiv:2501.19202.

\bibitem[{Jang et~al.(2023)Jang, Yoon, Yang, Cha, Lee, Logeswaran, and Seo}]{jang-etal-2023-knowledge}
Jang, J.; Yoon, D.; Yang, S.; Cha, S.; Lee, M.; Logeswaran, L.; and Seo, M. 2023.
\newblock Knowledge Unlearning for Mitigating Privacy Risks in Language Models.
\newblock In Rogers, A.; Boyd-Graber, J.; and Okazaki, N., eds., \emph{Proceedings of the 61st Annual Meeting of the Association for Computational Linguistics (Volume 1: Long Papers)}, 14389--14408. Toronto, Canada: Association for Computational Linguistics.

\bibitem[{Jin et~al.(2024)Jin, Cao, Wang, He, Yuan, Li, Chen, Liu, and Zhao}]{DBLP:conf/nips/JinCWHYL00024}
Jin, Z.; Cao, P.; Wang, C.; He, Z.; Yuan, H.; Li, J.; Chen, Y.; Liu, K.; and Zhao, J. 2024.
\newblock {RWKU:} Benchmarking Real-World Knowledge Unlearning for Large Language Models.
\newblock In Globersons, A.; Mackey, L.; Belgrave, D.; Fan, A.; Paquet, U.; Tomczak, J.~M.; and Zhang, C., eds., \emph{Advances in Neural Information Processing Systems 38: Annual Conference on Neural Information Processing Systems 2024, NeurIPS 2024, Vancouver, BC, Canada, December 10 - 15, 2024}.

\bibitem[{Joshi et~al.(2017)Joshi, Choi, Weld, and Zettlemoyer}]{joshi-etal-2017-triviaqa}
Joshi, M.; Choi, E.; Weld, D.; and Zettlemoyer, L. 2017.
\newblock {T}rivia{QA}: A Large Scale Distantly Supervised Challenge Dataset for Reading Comprehension.
\newblock In Barzilay, R.; and Kan, M.-Y., eds., \emph{Proceedings of the 55th Annual Meeting of the Association for Computational Linguistics (Volume 1: Long Papers)}, 1601--1611. Vancouver, Canada: Association for Computational Linguistics.

\bibitem[{Kurmanji et~al.(2023)Kurmanji, Triantafillou, Hayes, and Triantafillou}]{DBLP:conf/nips/KurmanjiTHT23}
Kurmanji, M.; Triantafillou, P.; Hayes, J.; and Triantafillou, E. 2023.
\newblock Towards Unbounded Machine Unlearning.
\newblock In Oh, A.; Naumann, T.; Globerson, A.; Saenko, K.; Hardt, M.; and Levine, S., eds., \emph{Advances in Neural Information Processing Systems 36: Annual Conference on Neural Information Processing Systems 2023, NeurIPS 2023, New Orleans, LA, USA, December 10 - 16, 2023}.

\bibitem[{Li et~al.(2024)Li, Pan, Gopal, Yue, Berrios, Gatti, Li, Dombrowski, Goel, Mukobi, Helm-Burger, Lababidi et~al.}]{pmlr-v235-li24bc}
Li, N.; Pan, A.; Gopal, A.; Yue, S.; Berrios, D.; Gatti, A.; Li, J.~D.; Dombrowski, A.-K.; Goel, S.; Mukobi, G.; Helm-Burger, N.; Lababidi, R.; et~al. 2024.
\newblock The {WMDP} Benchmark: Measuring and Reducing Malicious Use with Unlearning.
\newblock In Salakhutdinov, R.; Kolter, Z.; Heller, K.; Weller, A.; Oliver, N.; Scarlett, J.; and Berkenkamp, F., eds., \emph{Proceedings of the 41st International Conference on Machine Learning}, volume 235 of \emph{Proceedings of Machine Learning Research}, 28525--28550. PMLR.

\bibitem[{Li et~al.(2023)Li, Zhang, Dubois, Taori, Gulrajani, Guestrin, Liang, and Hashimoto}]{alpaca_eval}
Li, X.; Zhang, T.; Dubois, Y.; Taori, R.; Gulrajani, I.; Guestrin, C.; Liang, P.; and Hashimoto, T.~B. 2023.
\newblock AlpacaEval: An Automatic Evaluator of Instruction-following Models.
\newblock \url{https://github.com/tatsu-lab/alpaca_eval}.

\bibitem[{Lieberum et~al.(2024)Lieberum, Rajamanoharan, Conmy, Smith, Sonnerat, Varma, Kram{\'{a}}r, Dragan, Shah, and Nanda}]{DBLP:journals/corr/abs-2408-05147}
Lieberum, T.; Rajamanoharan, S.; Conmy, A.; Smith, L.; Sonnerat, N.; Varma, V.; Kram{\'{a}}r, J.; Dragan, A.~D.; Shah, R.; and Nanda, N. 2024.
\newblock Gemma Scope: Open Sparse Autoencoders Everywhere All At Once on Gemma 2.
\newblock \emph{CoRR}, abs/2408.05147.

\bibitem[{Lin, Hilton, and Evans(2022)}]{lin-etal-2022-truthfulqa}
Lin, S.; Hilton, J.; and Evans, O. 2022.
\newblock {T}ruthful{QA}: Measuring How Models Mimic Human Falsehoods.
\newblock In Muresan, S.; Nakov, P.; and Villavicencio, A., eds., \emph{Proceedings of the 60th Annual Meeting of the Association for Computational Linguistics (Volume 1: Long Papers)}, 3214--3252. Dublin, Ireland: Association for Computational Linguistics.

\bibitem[{Maini et~al.(2024)Maini, Feng, Schwarzschild, Lipton, and Kolter}]{DBLP:journals/corr/abs-2401-06121}
Maini, P.; Feng, Z.; Schwarzschild, A.; Lipton, Z.~C.; and Kolter, J.~Z. 2024.
\newblock {TOFU:} {A} Task of Fictitious Unlearning for LLMs.
\newblock \emph{CoRR}, abs/2401.06121.

\bibitem[{Meng et~al.(2022)Meng, Bau, Andonian, and Belinkov}]{DBLP:conf/nips/MengBAB22}
Meng, K.; Bau, D.; Andonian, A.; and Belinkov, Y. 2022.
\newblock Locating and Editing Factual Associations in {GPT}.
\newblock In Koyejo, S.; Mohamed, S.; Agarwal, A.; Belgrave, D.; Cho, K.; and Oh, A., eds., \emph{Advances in Neural Information Processing Systems 35: Annual Conference on Neural Information Processing Systems 2022, NeurIPS 2022, New Orleans, LA, USA, November 28 - December 9, 2022}.

\bibitem[{Nanda et~al.(2023)Nanda, Rajamanoharan, Kram\'{a}r, and Shah}]{nanda2023factfinding}
Nanda, N.; Rajamanoharan, S.; Kram\'{a}r, J.; and Shah, R. 2023.
\newblock Fact Finding: Attempting to Reverse-Engineer Factual Recall on the Neuron Level.

\bibitem[{Neel, Roth, and Sharifi{-}Malvajerdi(2021)}]{DBLP:conf/alt/Neel0S21}
Neel, S.; Roth, A.; and Sharifi{-}Malvajerdi, S. 2021.
\newblock Descent-to-Delete: Gradient-Based Methods for Machine Unlearning.
\newblock In Feldman, V.; Ligett, K.; and Sabato, S., eds., \emph{Algorithmic Learning Theory, 16-19 March 2021, Virtual Conference, Worldwide}, volume 132 of \emph{Proceedings of Machine Learning Research}, 931--962. {PMLR}.

\bibitem[{Pawelczyk, Neel, and Lakkaraju(2024)}]{pmlr-v235-pawelczyk24a}
Pawelczyk, M.; Neel, S.; and Lakkaraju, H. 2024.
\newblock In-Context Unlearning: Language Models as Few-Shot Unlearners.
\newblock In Salakhutdinov, R.; Kolter, Z.; Heller, K.; Weller, A.; Oliver, N.; Scarlett, J.; and Berkenkamp, F., eds., \emph{Proceedings of the 41st International Conference on Machine Learning}, volume 235 of \emph{Proceedings of Machine Learning Research}, 40034--40050. PMLR.

\bibitem[{Qi et~al.(2020)Qi, Zhang, Zhang, Bolton, and Manning}]{DBLP:conf/acl/QiZZBM20}
Qi, P.; Zhang, Y.; Zhang, Y.; Bolton, J.; and Manning, C.~D. 2020.
\newblock Stanza: {A} Python Natural Language Processing Toolkit for Many Human Languages.
\newblock In Celikyilmaz, A.; and Wen, T., eds., \emph{Proceedings of the 58th Annual Meeting of the Association for Computational Linguistics: System Demonstrations, {ACL} 2020, Online, July 5-10, 2020}, 101--108. Association for Computational Linguistics.

\bibitem[{Rajamanoharan et~al.(2024)Rajamanoharan, Lieberum, Sonnerat, Conmy, Varma, Kram{\'{a}}r, and Nanda}]{DBLP:journals/corr/abs-2407-14435}
Rajamanoharan, S.; Lieberum, T.; Sonnerat, N.; Conmy, A.; Varma, V.; Kram{\'{a}}r, J.; and Nanda, N. 2024.
\newblock Jumping Ahead: Improving Reconstruction Fidelity with JumpReLU Sparse Autoencoders.
\newblock \emph{CoRR}, abs/2407.14435.

\bibitem[{Rivi{\`{e}}re et~al.(2024)Rivi{\`{e}}re, Pathak, Sessa, Hardin, Bhupatiraju, Hussenot, Mesnard, Shahriari, Ram{\'{e}}, Ferret, Liu, Tafti et~al.}]{gemma_2024}
Rivi{\`{e}}re, M.; Pathak, S.; Sessa, P.~G.; Hardin, C.; Bhupatiraju, S.; Hussenot, L.; Mesnard, T.; Shahriari, B.; Ram{\'{e}}, A.; Ferret, J.; Liu, P.; Tafti, P.; et~al. 2024.
\newblock Gemma 2: Improving Open Language Models at a Practical Size.
\newblock \emph{CoRR}, abs/2408.00118.

\bibitem[{Suzgun et~al.(2023)Suzgun, Scales, Sch{\"{a}}rli, Gehrmann, Tay, Chung, Chowdhery, Le, Chi, Zhou, and Wei}]{DBLP:conf/acl/SuzgunSSGTCCLCZ23}
Suzgun, M.; Scales, N.; Sch{\"{a}}rli, N.; Gehrmann, S.; Tay, Y.; Chung, H.~W.; Chowdhery, A.; Le, Q.~V.; Chi, E.~H.; Zhou, D.; and Wei, J. 2023.
\newblock Challenging BIG-Bench Tasks and Whether Chain-of-Thought Can Solve Them.
\newblock In Rogers, A.; Boyd{-}Graber, J.~L.; and Okazaki, N., eds., \emph{Findings of the Association for Computational Linguistics: {ACL} 2023, Toronto, Canada, July 9-14, 2023}, 13003--13051. Association for Computational Linguistics.

\bibitem[{Voigt and Bussche(2017)}]{10.5555/3152676}
Voigt, P.; and Bussche, A. v.~d. 2017.
\newblock \emph{The EU General Data Protection Regulation (GDPR): A Practical Guide}.
\newblock Springer Publishing Company, Incorporated, 1st edition.
\newblock ISBN 3319579584.

\bibitem[{Wang et~al.(2023)Wang, Variengien, Conmy, Shlegeris, and Steinhardt}]{DBLP:conf/iclr/WangVCSS23}
Wang, K.~R.; Variengien, A.; Conmy, A.; Shlegeris, B.; and Steinhardt, J. 2023.
\newblock Interpretability in the Wild: a Circuit for Indirect Object Identification in {GPT-2} Small.
\newblock In \emph{The Eleventh International Conference on Learning Representations, {ICLR} 2023, Kigali, Rwanda, May 1-5, 2023}. OpenReview.net.

\bibitem[{Xu et~al.(2023)Xu, Zhu, Zhang, Zhou, and Yu}]{10.1145/3603620}
Xu, H.; Zhu, T.; Zhang, L.; Zhou, W.; and Yu, P.~S. 2023.
\newblock Machine Unlearning: A Survey.
\newblock \emph{ACM Comput. Surv.}, 56(1).

\bibitem[{Yao, Xu, and Liu(2024)}]{yao2024large}
Yao, Y.; Xu, X.; and Liu, Y. 2024.
\newblock Large Language Model Unlearning.
\newblock In \emph{The Thirty-eighth Annual Conference on Neural Information Processing Systems}.

\bibitem[{Zhang et~al.(2024)Zhang, Lin, Bai, and Mei}]{zhang2024negative}
Zhang, R.; Lin, L.; Bai, Y.; and Mei, S. 2024.
\newblock Negative Preference Optimization: From Catastrophic Collapse to Effective Unlearning.
\newblock In \emph{First Conference on Language Modeling}.

\bibitem[{Zhang et~al.(2018)Zhang, Galley, Gao, Gan, Li, Brockett, and Dolan}]{DBLP:conf/nips/ZhangGGGLBD18}
Zhang, Y.; Galley, M.; Gao, J.; Gan, Z.; Li, X.; Brockett, C.; and Dolan, B. 2018.
\newblock Generating Informative and Diverse Conversational Responses via Adversarial Information Maximization.
\newblock In Bengio, S.; Wallach, H.~M.; Larochelle, H.; Grauman, K.; Cesa{-}Bianchi, N.; and Garnett, R., eds., \emph{Advances in Neural Information Processing Systems 31: Annual Conference on Neural Information Processing Systems 2018, NeurIPS 2018, December 3-8, 2018, Montr{\'{e}}al, Canada}, 1815--1825.

\end{thebibliography}

\appendix
\section{Full Results}
\label{full_results}
This section provides additional evaluation results that complement the main paper. We include:
(i) per-epoch evaluation, (ii) effects on attribute rate, and (iii) effects on recognition latents.
\subsection{Per-Epoch Evaluation}
\begin{figure*}[tb]
\centering
\begin{minipage}{0.42\textwidth}
\centering
\includegraphics[width=\textwidth]{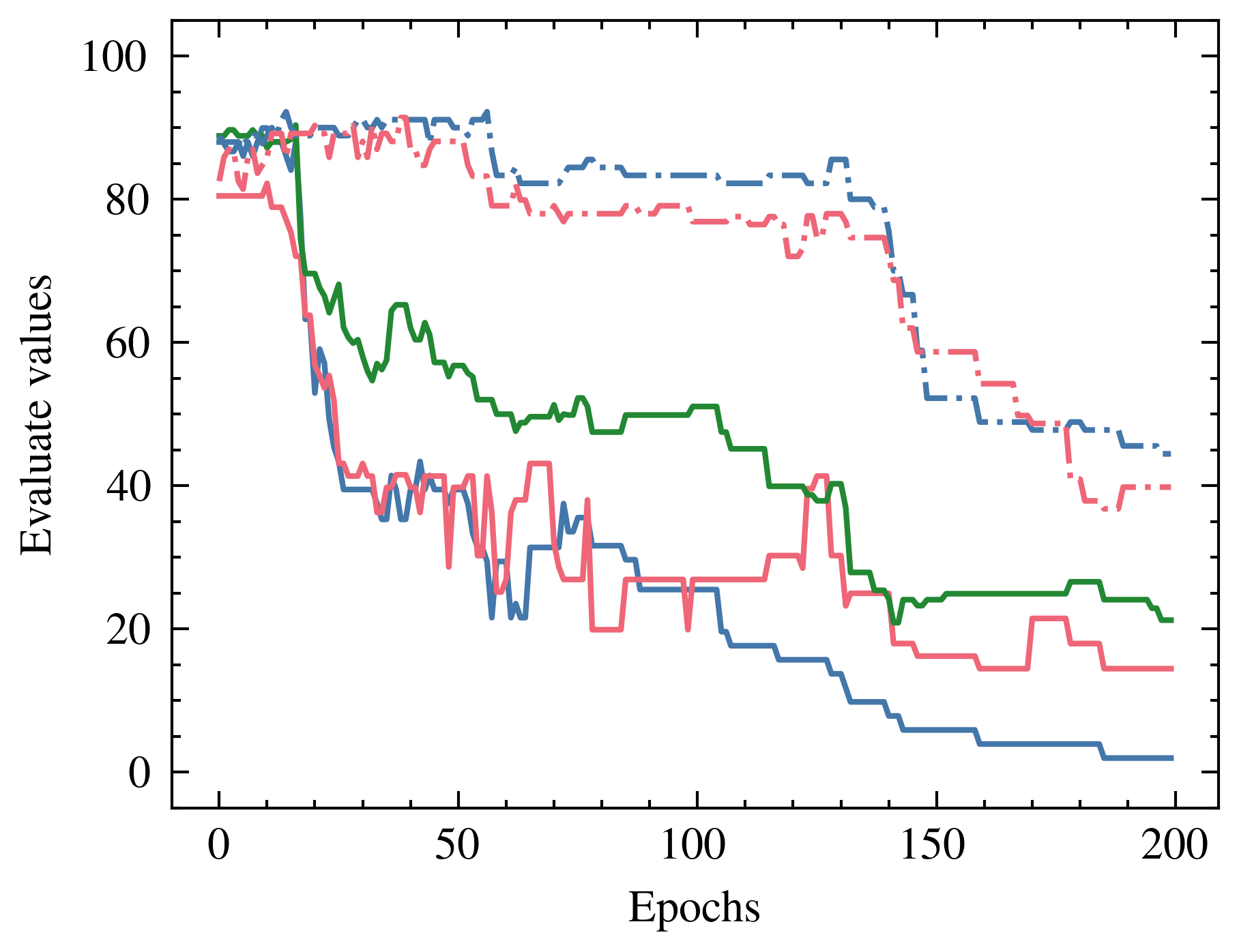}
\subcaption{Results on Llama3.1-8B Instruct.}
\label{proposal_perepoch}
\end{minipage}
\hspace{0.06\textwidth}   
\begin{minipage}{0.42\textwidth}
\centering
\includegraphics[width=\textwidth]{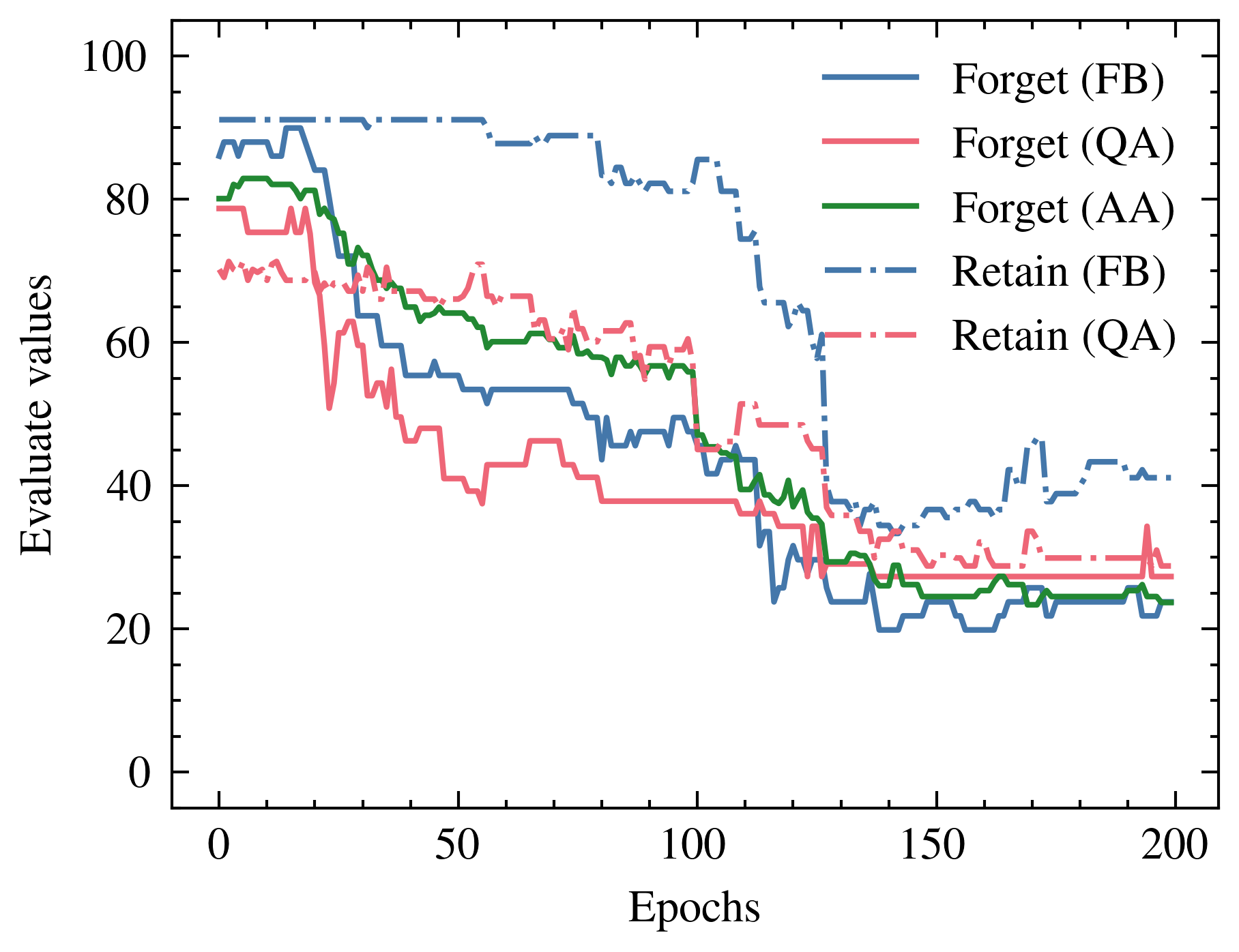}
\subcaption{Results on Gemma2-9B Instruct.}
\label{proposal_perepoch_gemma}
\end{minipage}
\caption{\textbf{Per-epoch evaluation for proposal.}
Blue denotes fill-in-the-blank (FB), pink denotes question answering (QA), and green denotes adversarial attack (AA). 
Solid lines indicate forget scores, while dashed lines indicate retain scores. 
Forget scores are reported for FB, QA, and AA; retain scores are reported for FB and QA. 
Unlearning is run for 200 epochs, and all metrics are evaluated and plotted at every epoch.}
\end{figure*}

\begin{figure*}
    \centering
    \begin{minipage}[t]{0.33\textwidth}
    \centering
    \includegraphics[width=\linewidth]{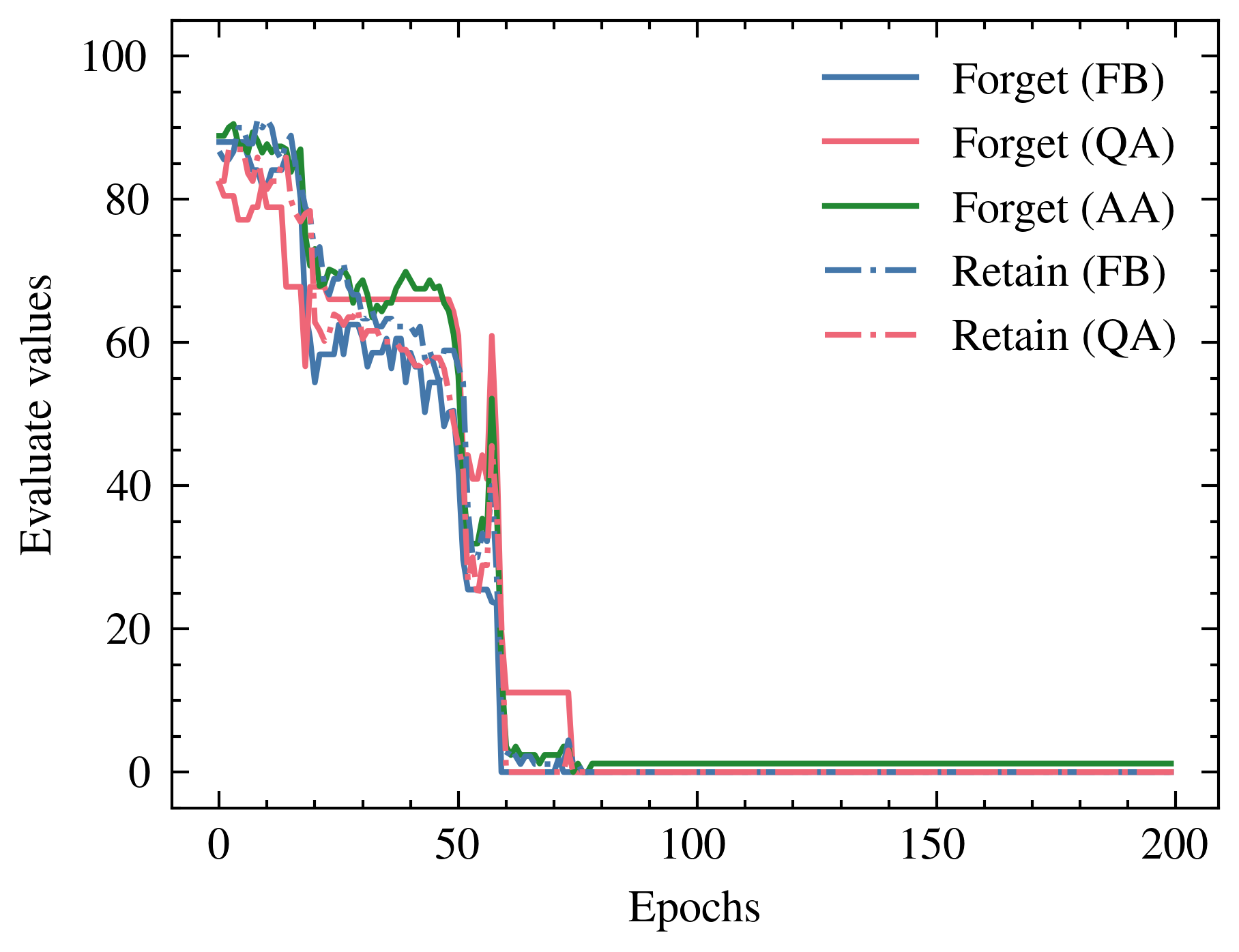}
    \subcaption{GA (sentence)}
    \end{minipage}
    \begin{minipage}[t]{0.33\textwidth}
    \centering
    \includegraphics[width=\linewidth]{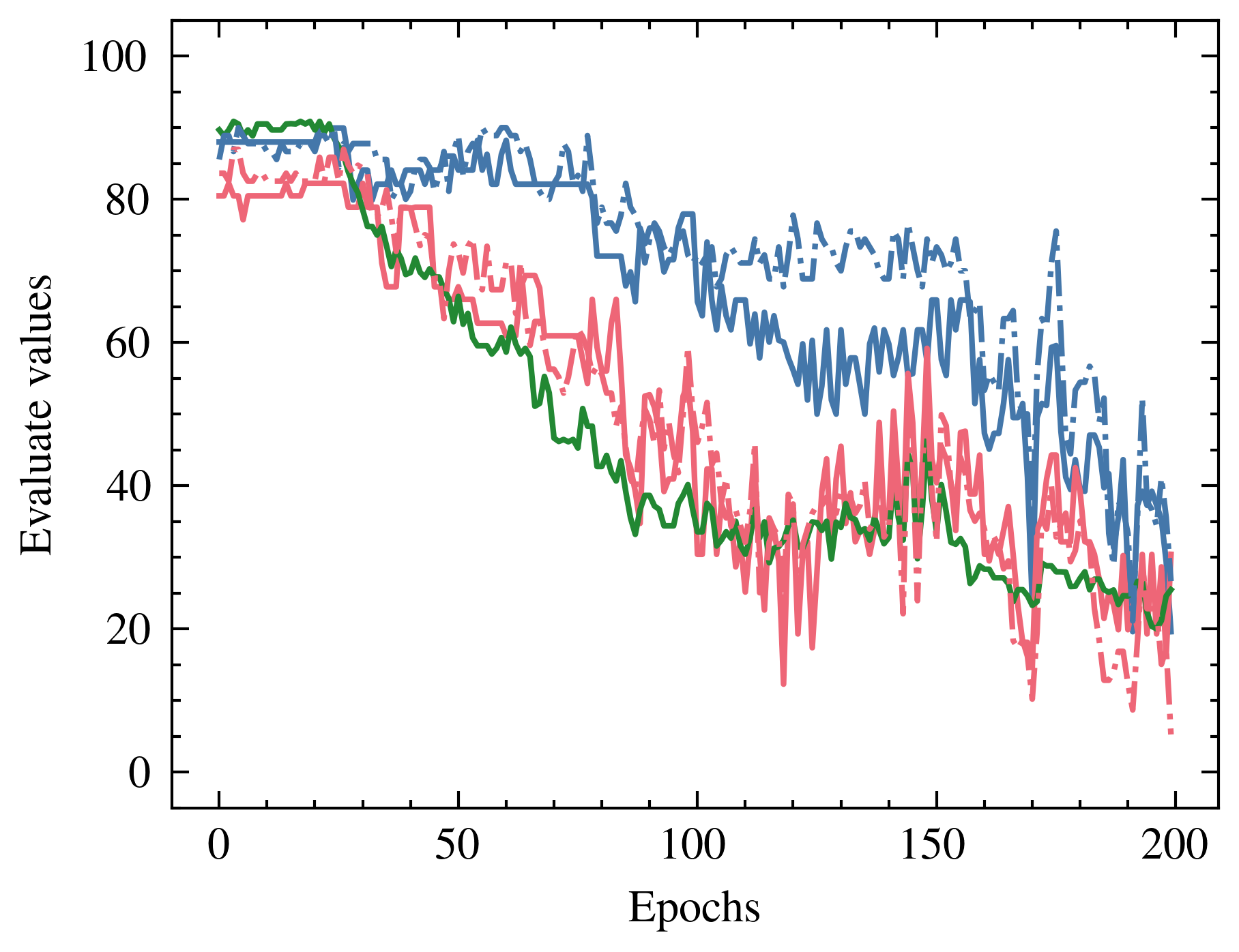}
    \subcaption{GA (entity)}
    \end{minipage}
    \begin{minipage}[t]{0.33\textwidth}
    \centering
    \includegraphics[width=\linewidth]{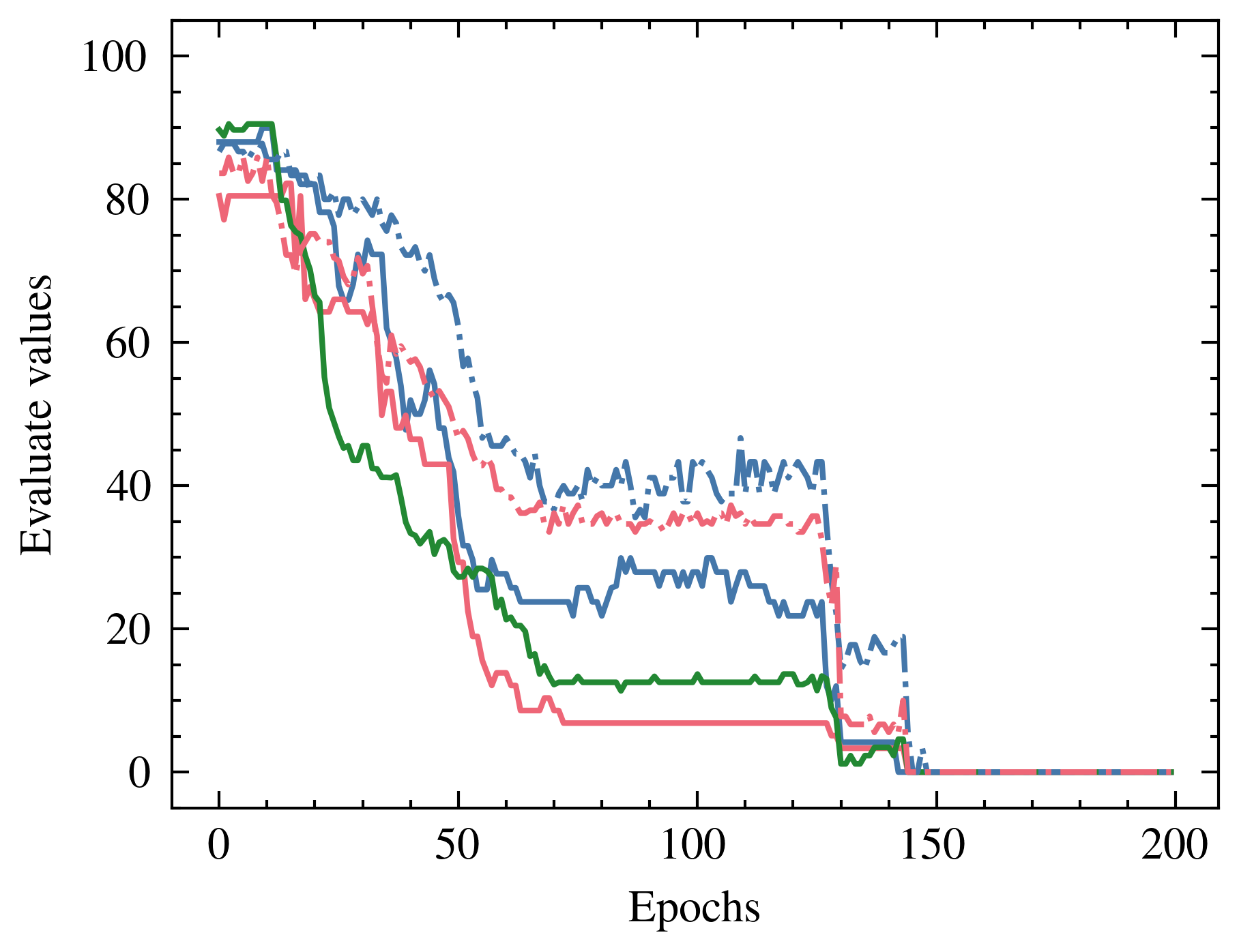}
    \subcaption{GA (last token)}
    \end{minipage}

    \begin{minipage}[t]{0.33\textwidth}
    \centering
    \includegraphics[width=\linewidth]{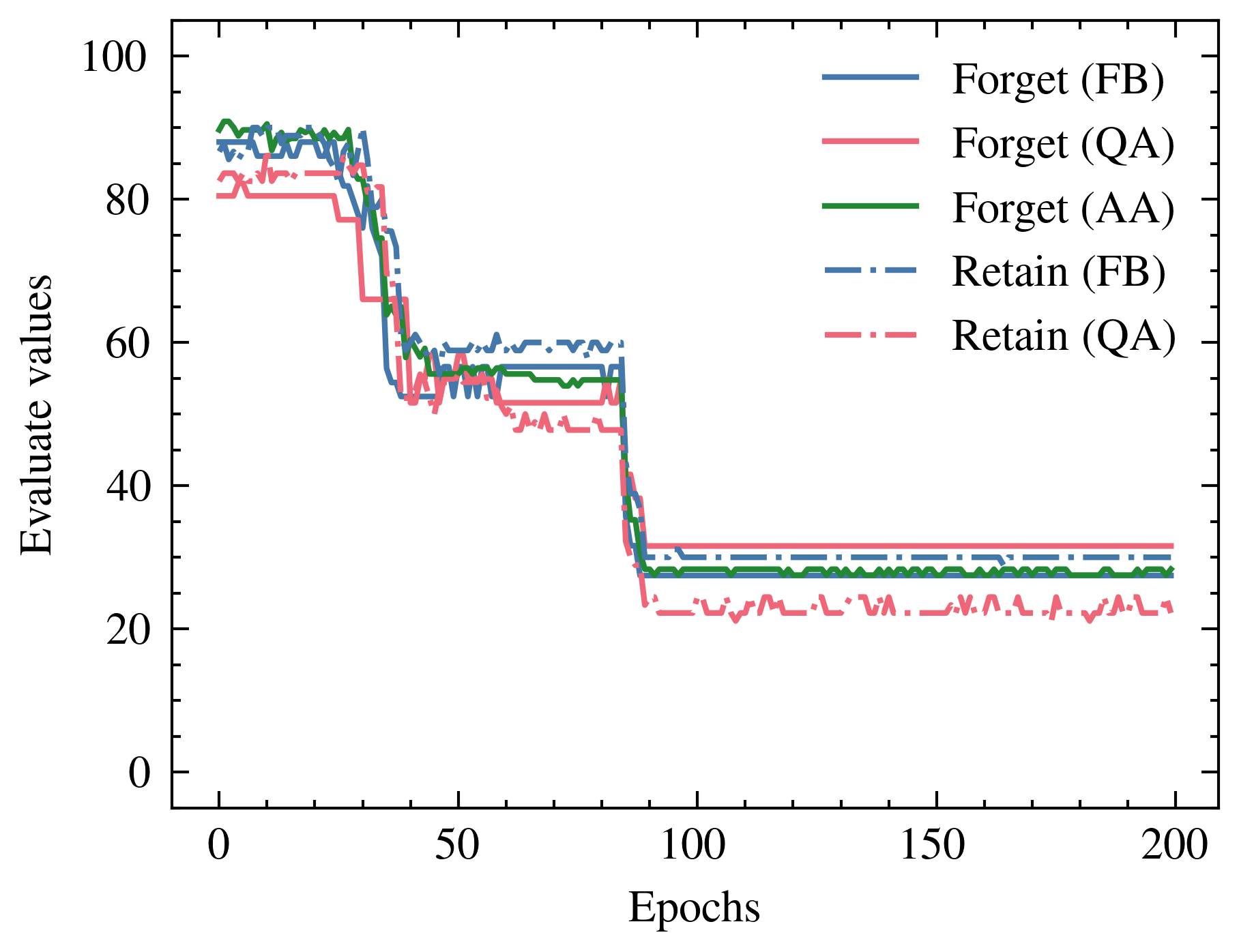}
    \subcaption{NPO (sentence)}
    \end{minipage}
    \begin{minipage}[t]{0.33\textwidth}
    \centering
    \includegraphics[width=\linewidth]{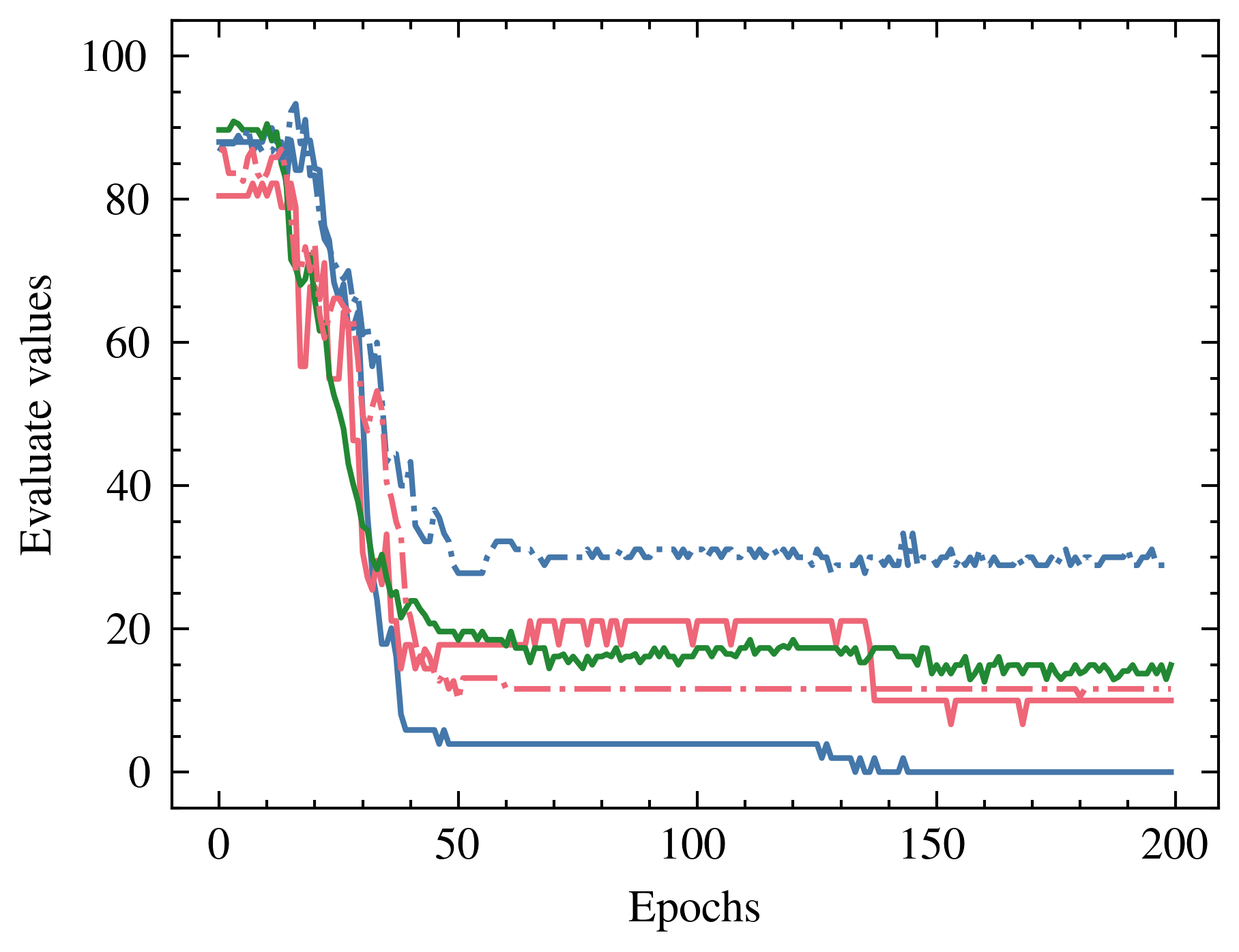}
    \subcaption{NPO (entity)}
    \end{minipage}
    \begin{minipage}[t]{0.33\textwidth}
    \centering
    \includegraphics[width=\linewidth]{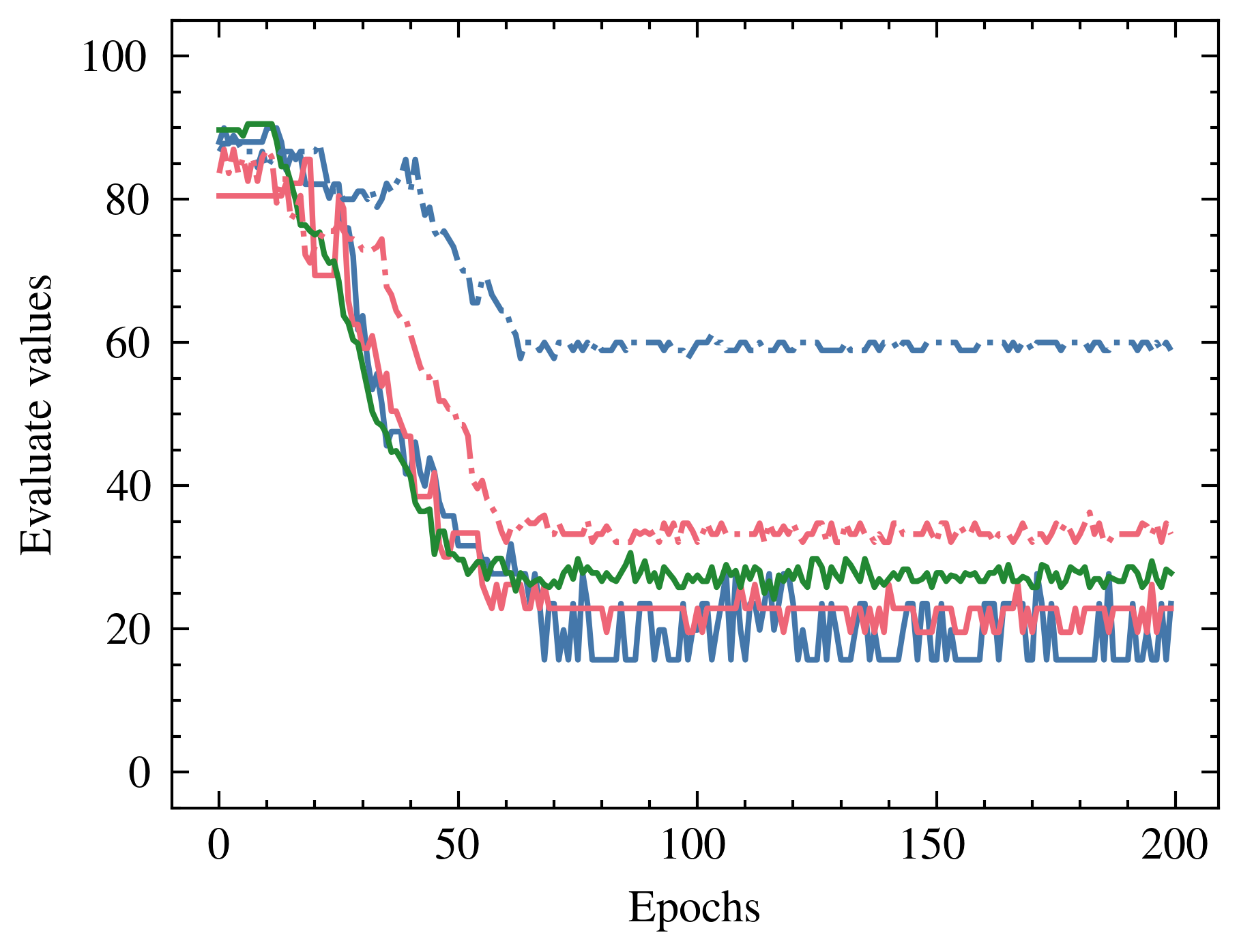}
    \subcaption{NPO (last token)}
    \end{minipage}

    \begin{minipage}[t]{0.33\textwidth}
    \centering
    \includegraphics[width=\linewidth]{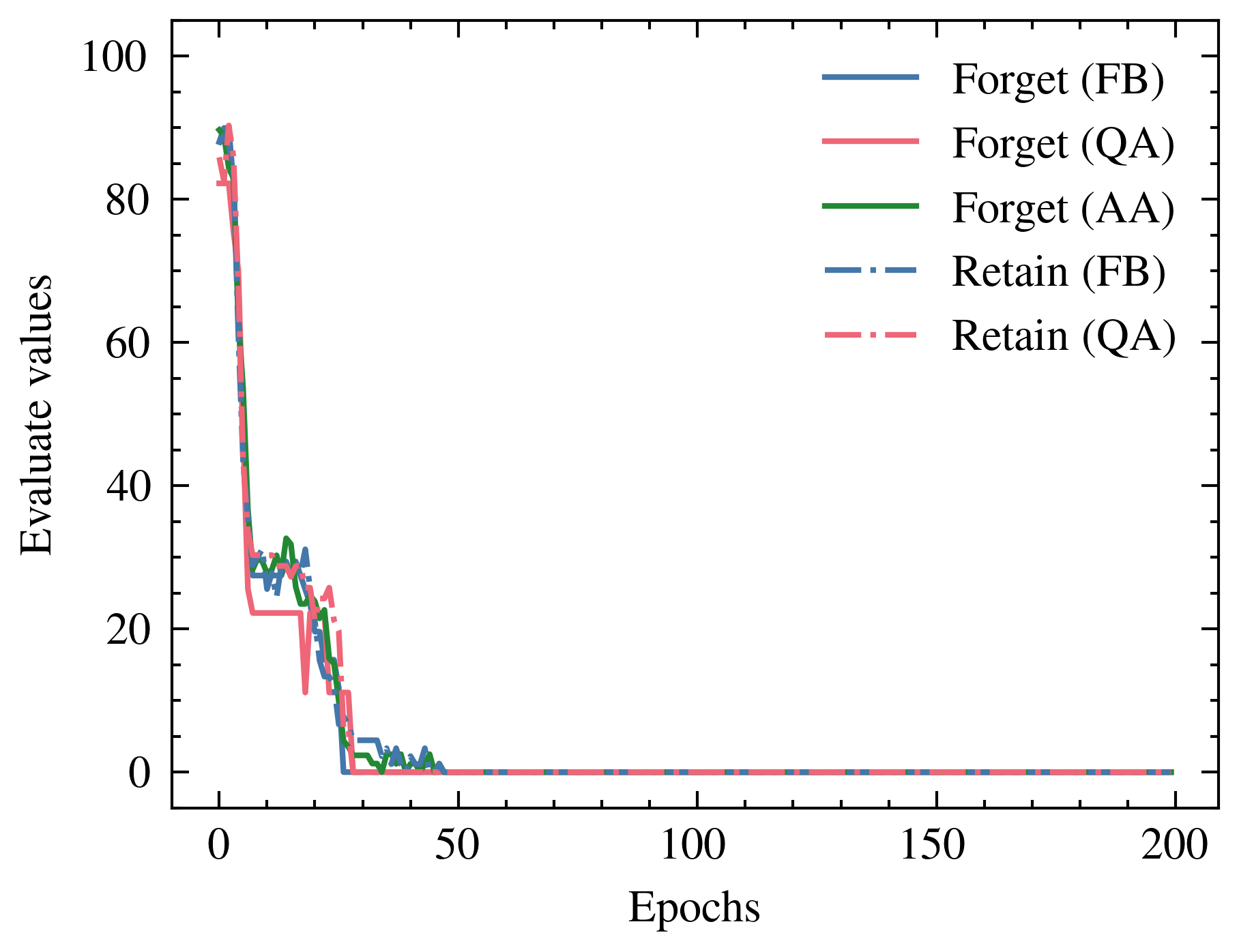}
    \subcaption{RMU (sentence)}
    \end{minipage}
    \begin{minipage}[t]{0.33\textwidth}
    \centering
    \includegraphics[width=\linewidth]{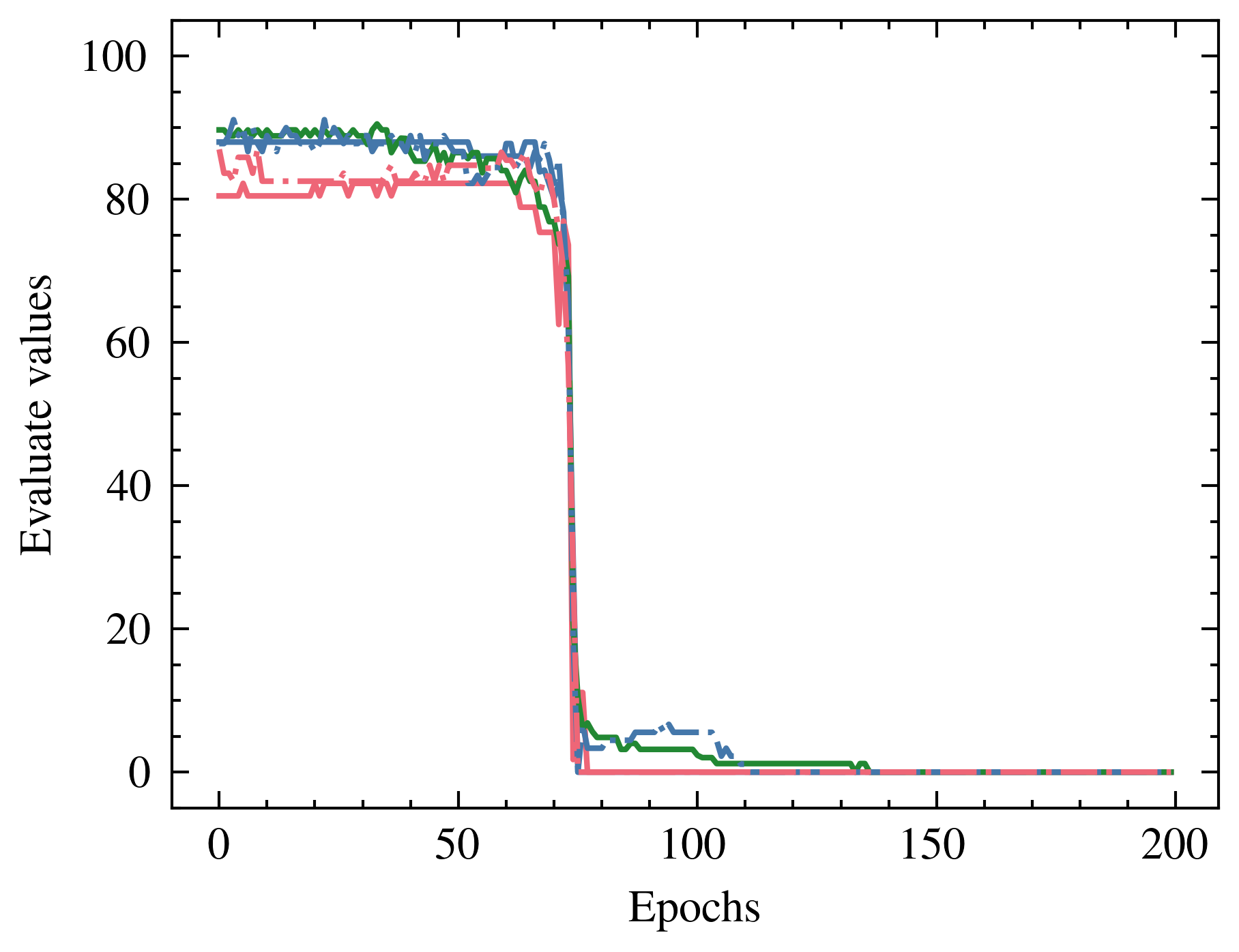}
    \subcaption{RMU (entity)}
    \end{minipage}
    \begin{minipage}[t]{0.33\textwidth}
    \centering
    \includegraphics[width=\linewidth]{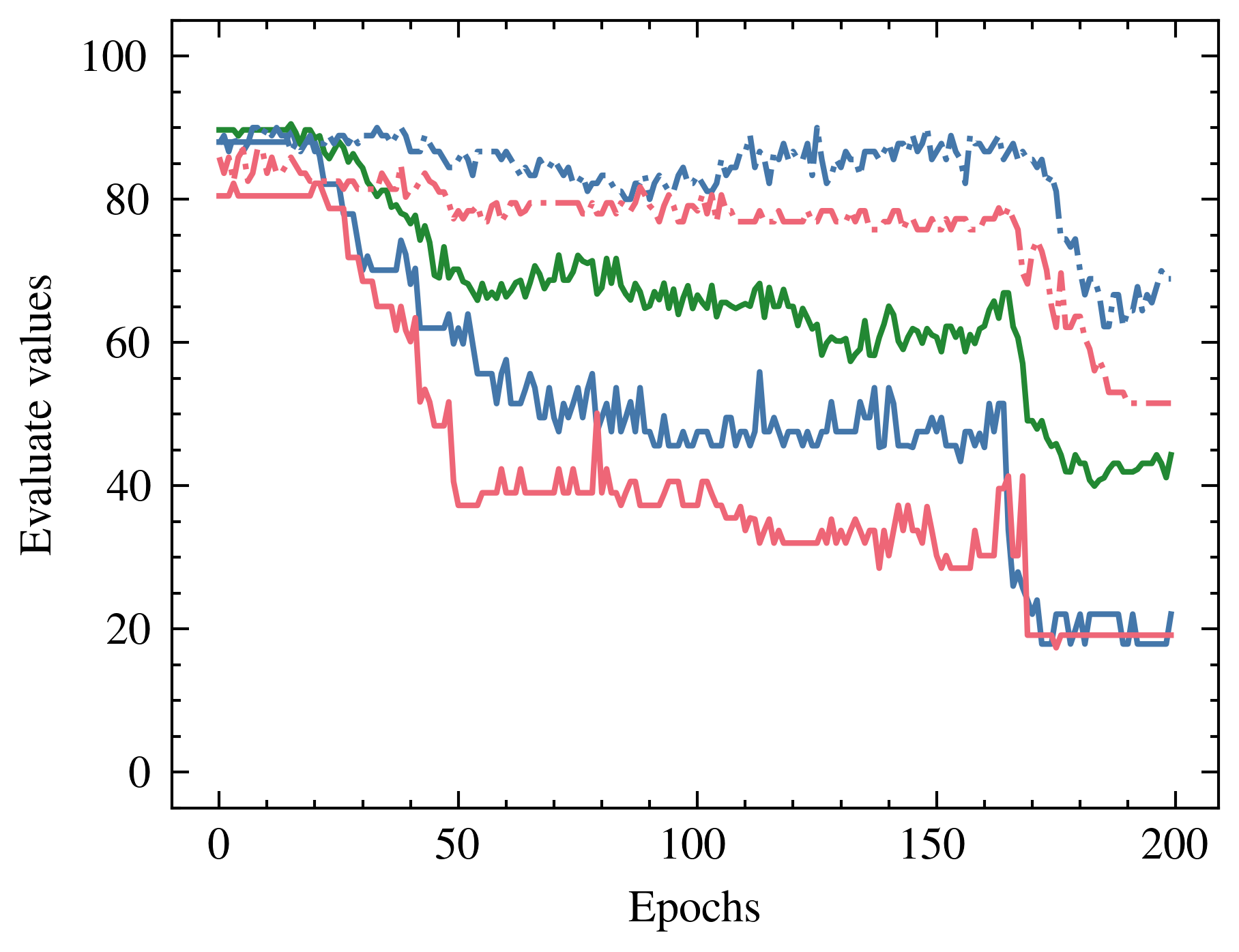}
    \subcaption{RMU (last token)}
    \end{minipage}
    \caption{\textbf{Per-epoch evaluation for baselines on Llama3.1-8B Instruct.}}
    \label{perepoch}
\end{figure*}

\begin{figure*}
    \centering
    \begin{minipage}[t]{0.33\textwidth}
    \centering
    \includegraphics[width=\linewidth]{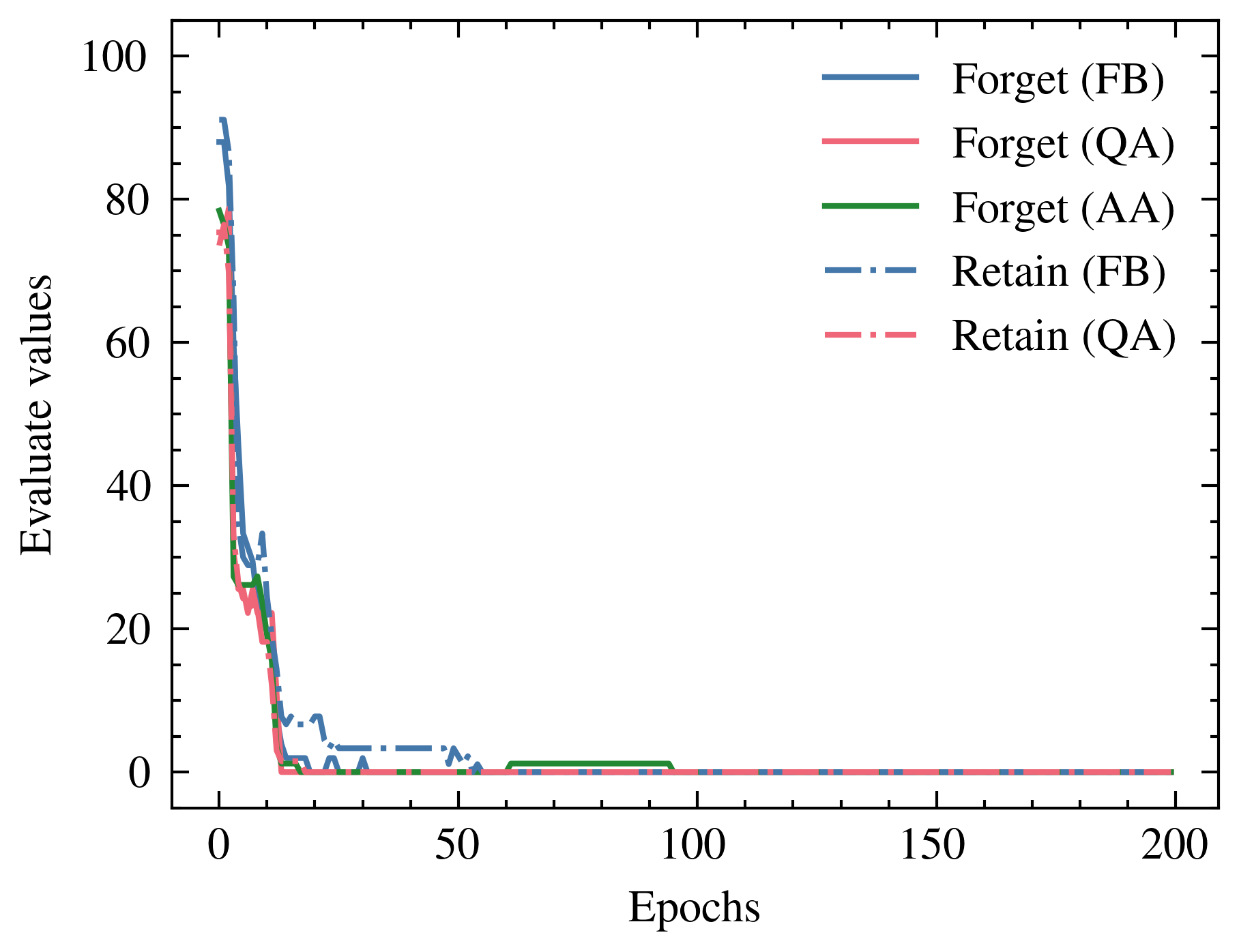}
    \subcaption{GA (sentence)}
    \end{minipage}
    \begin{minipage}[t]{0.33\textwidth}
    \centering
    \includegraphics[width=\linewidth]{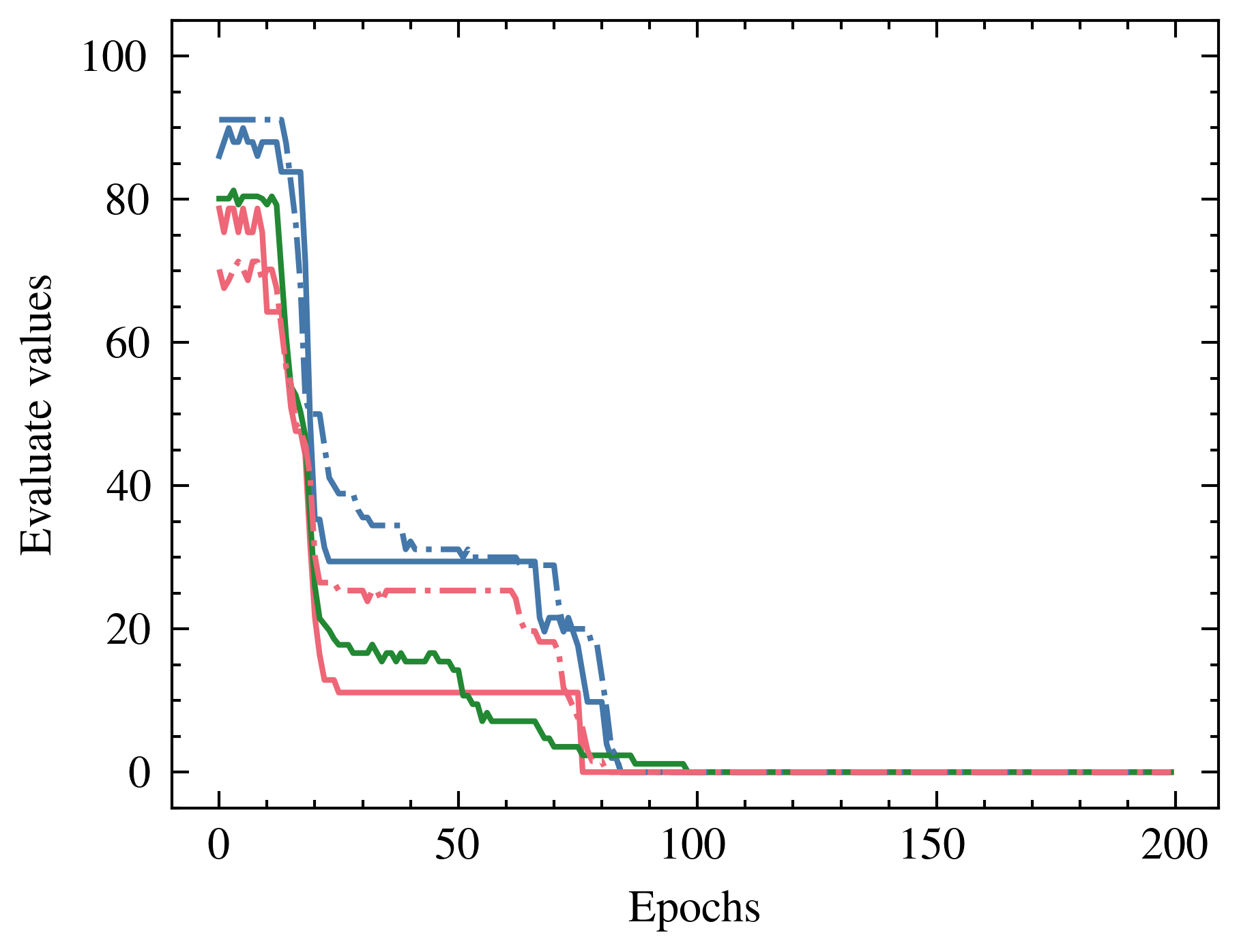}
    \subcaption{GA (entity)}
    \end{minipage}
    \begin{minipage}[t]{0.33\textwidth}
    \centering
    \includegraphics[width=\linewidth]{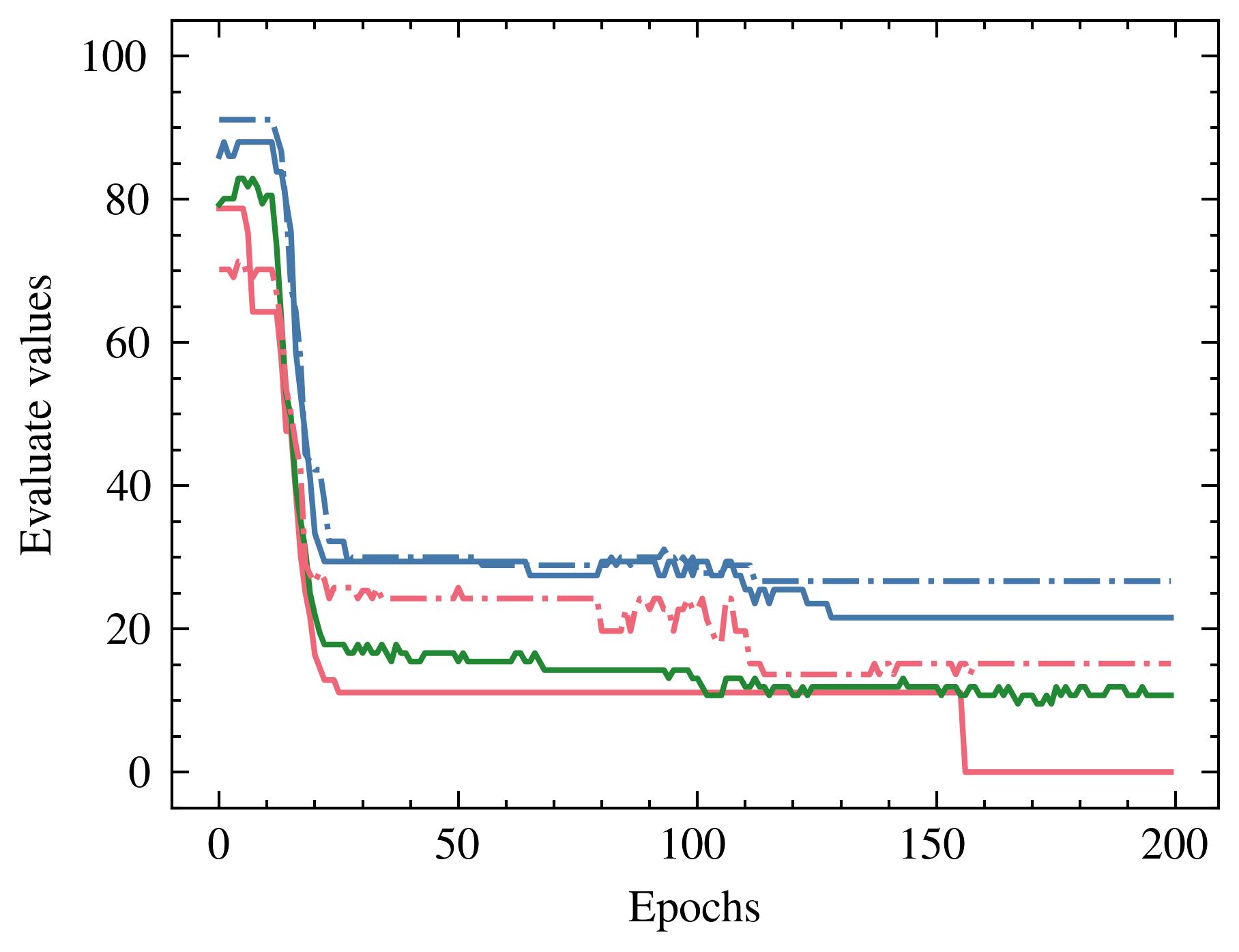}
    \subcaption{GA (last token)}
    \end{minipage}

    \begin{minipage}[t]{0.33\textwidth}
    \centering
    \includegraphics[width=\linewidth]{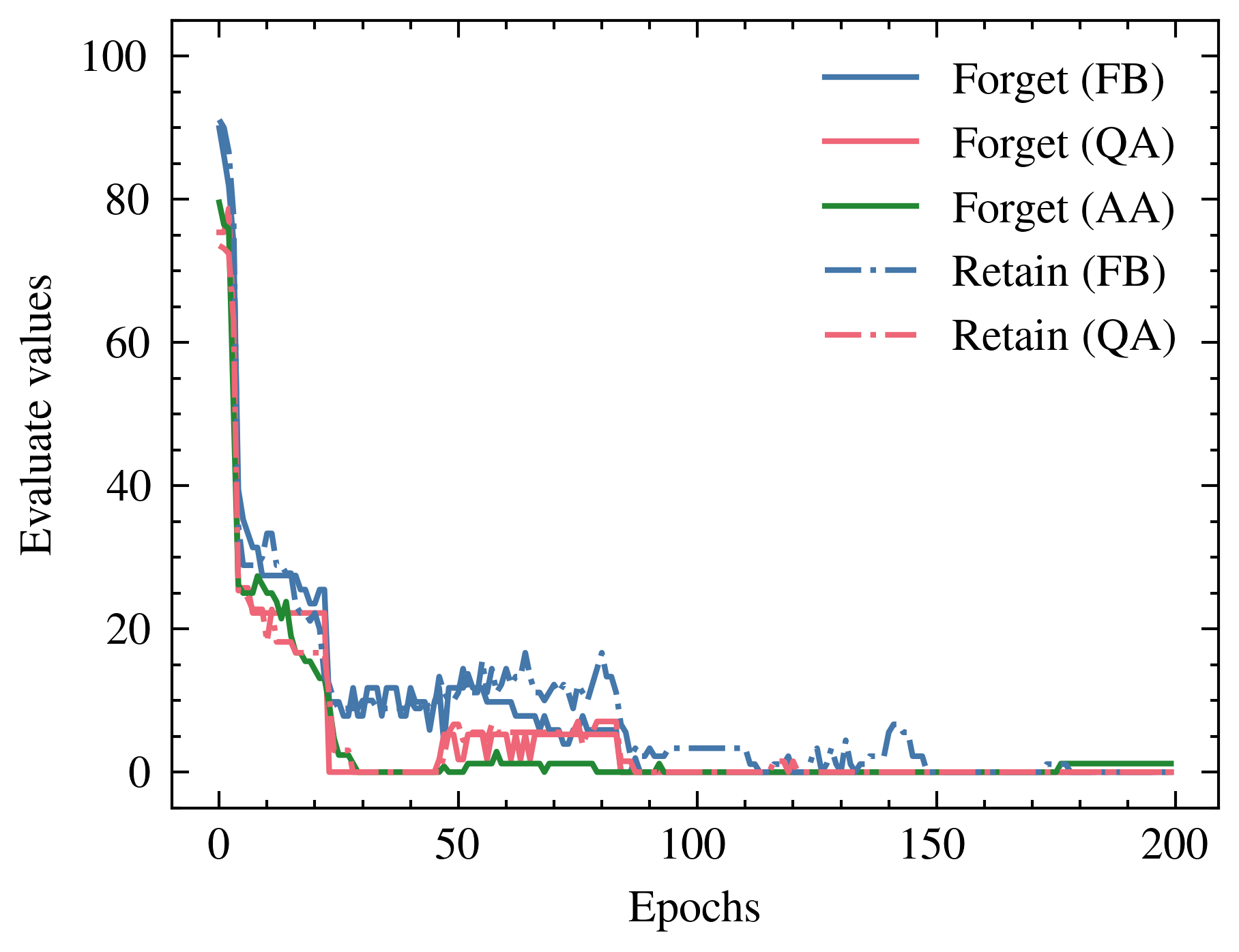}
    \subcaption{NPO (sentence)}
    \end{minipage}
    \begin{minipage}[t]{0.33\textwidth}
    \centering
    \includegraphics[width=\linewidth]{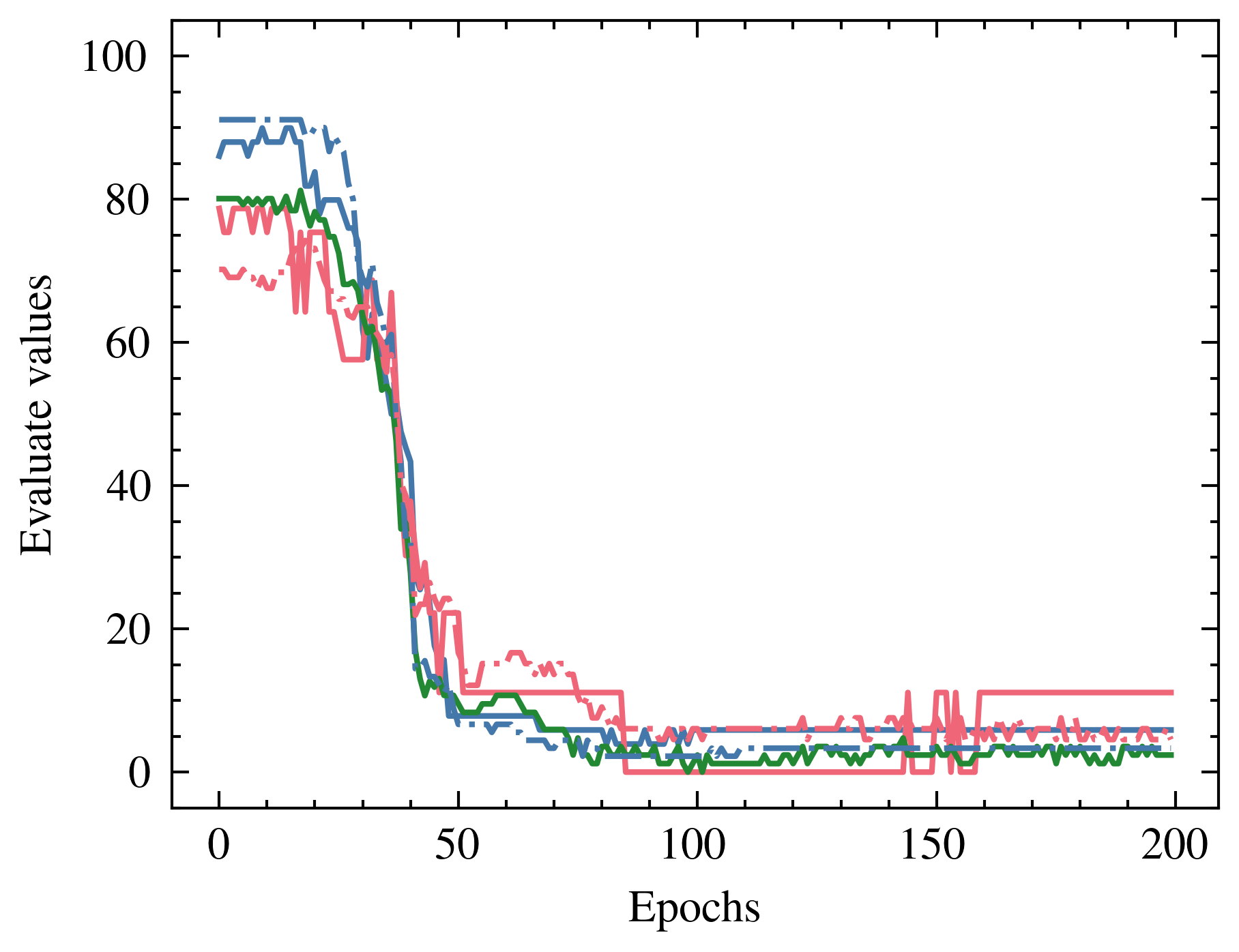}
    \subcaption{NPO (entity)}
    \end{minipage}
    \begin{minipage}[t]{0.33\textwidth}
    \centering
    \includegraphics[width=\linewidth]{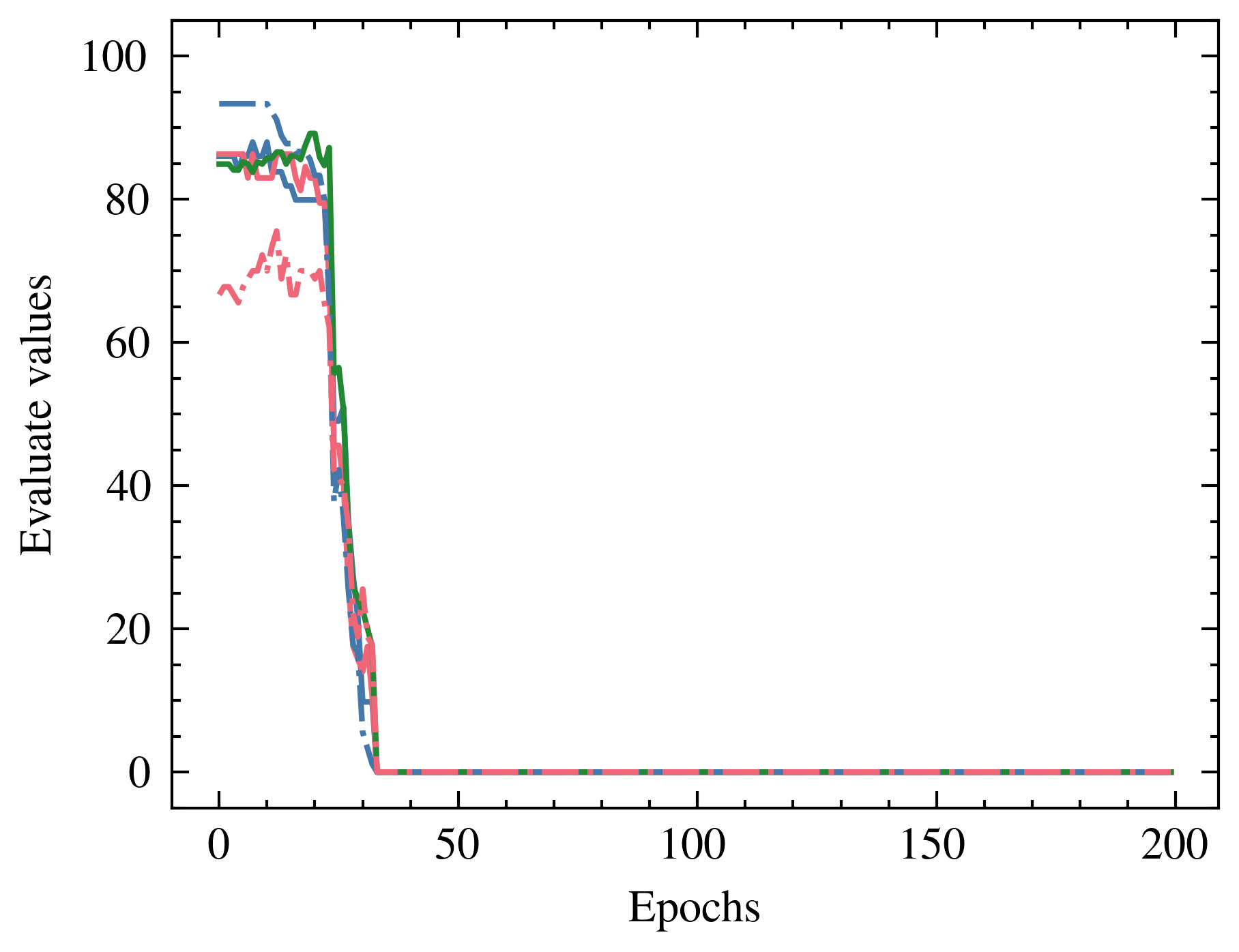}
    \subcaption{NPO (last token)}
    \end{minipage}

    \begin{minipage}[t]{0.33\textwidth}
    \centering
    \includegraphics[width=\linewidth]{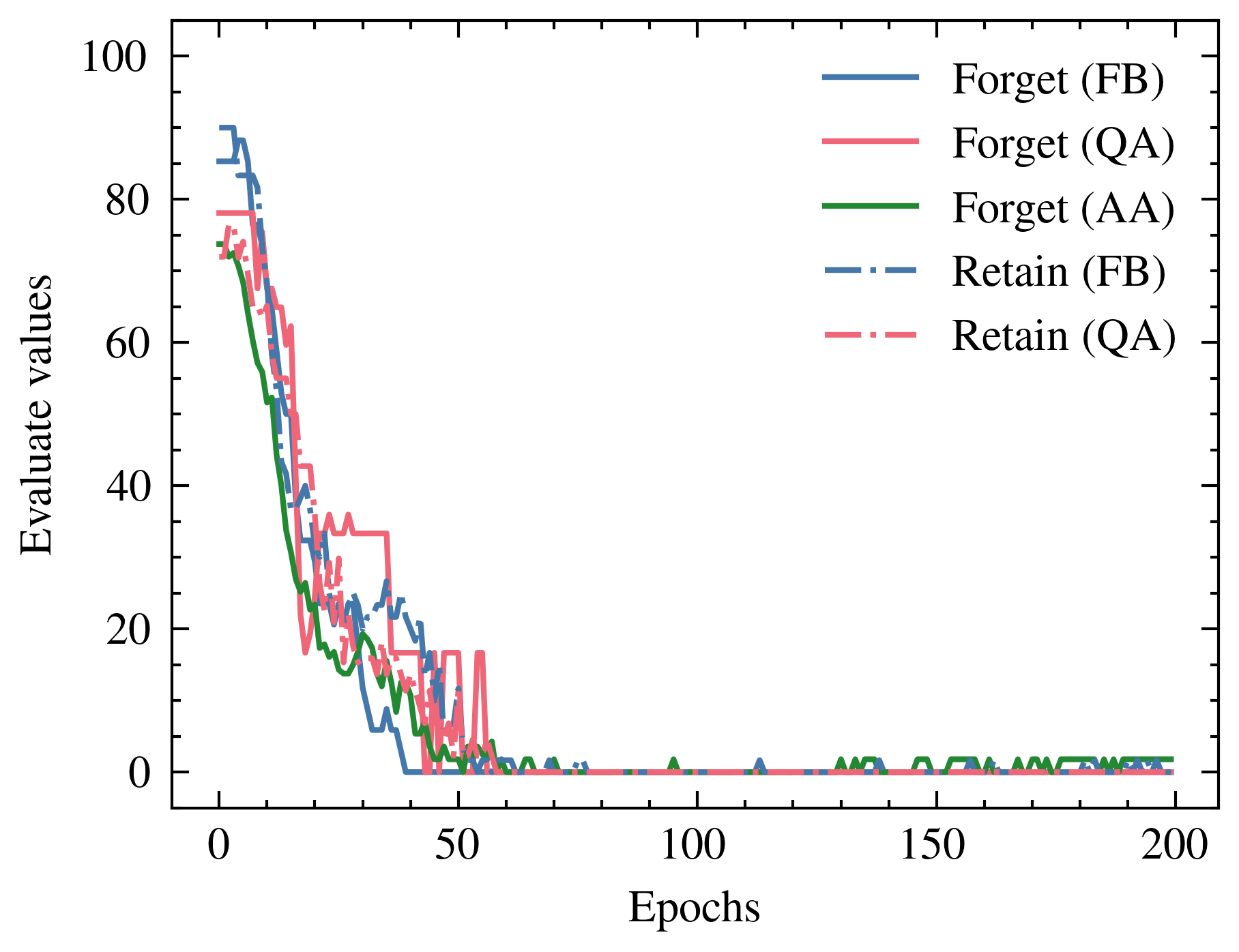}
    \subcaption{RMU (sentence)}
    \end{minipage}
    \begin{minipage}[t]{0.33\textwidth}
    \centering
    \includegraphics[width=\linewidth]{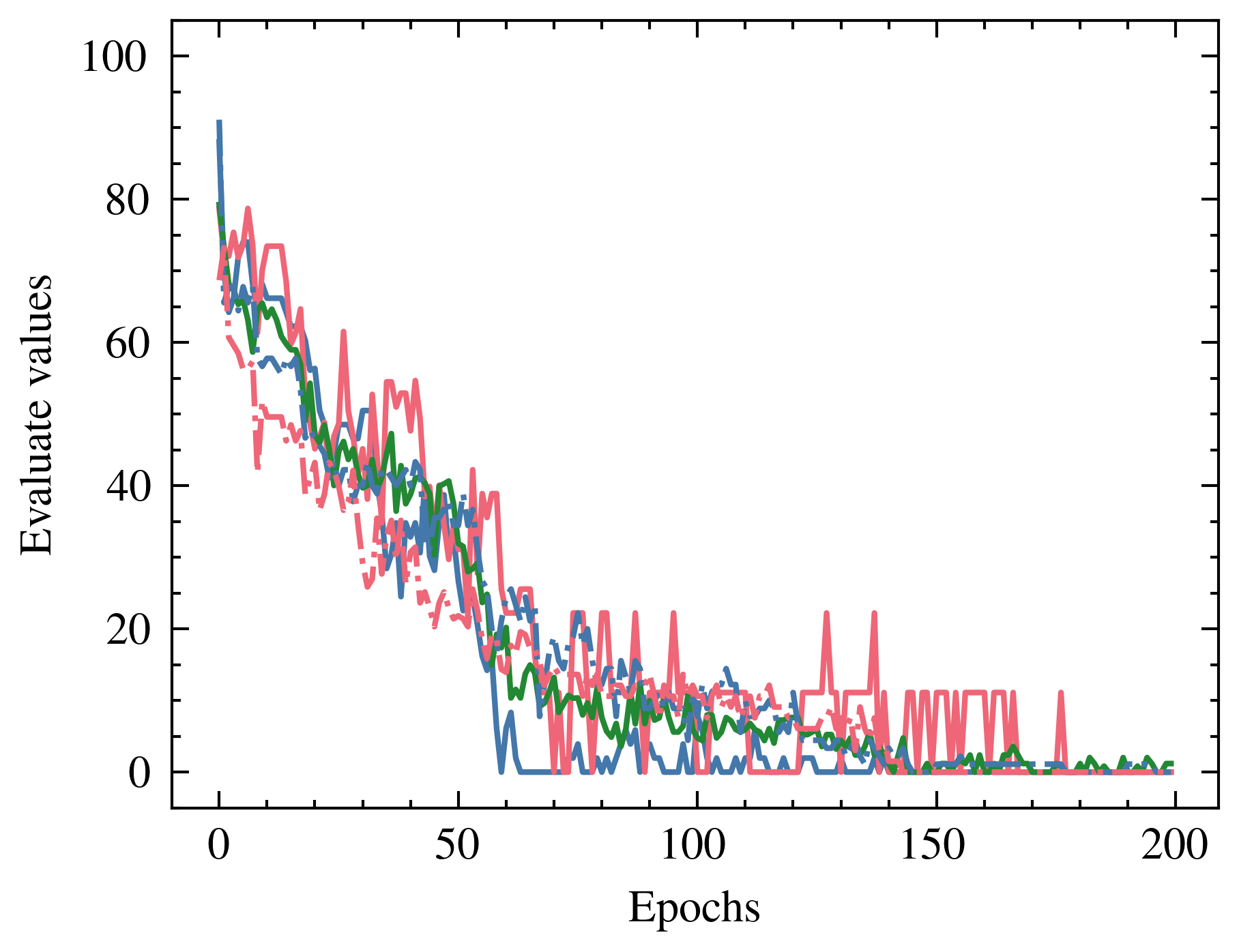}
    \subcaption{RMU (entity)}
    \end{minipage}
    \begin{minipage}[t]{0.33\textwidth}
    \centering
    \includegraphics[width=\linewidth]{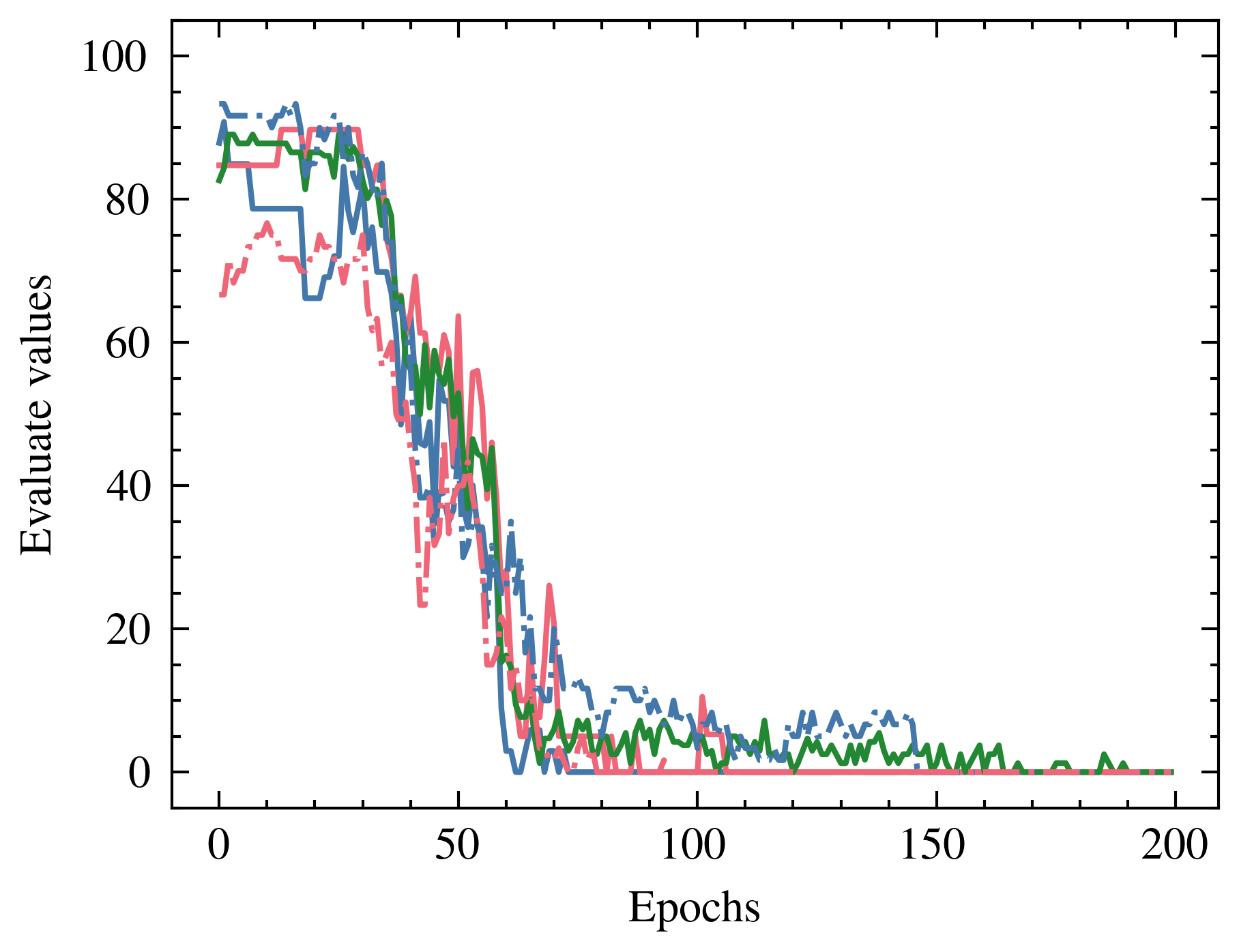}
    \subcaption{RMU (last token)}
    \end{minipage}
    \caption{\textbf{Per-epoch evaluation for baselines on Gemma2-9B Instruct.}}
    \label{perepoch_gemma}
\end{figure*}

In Table~\ref{main_result} in the main paper, we evaluate unlearning methods every 10 epochs due to computational resource constraints.
To address this limitation, we present finer-grained results by reporting evaluation metrics at every epoch.
Experiments are conducted on three target entities in the RWKU dataset.
We measure both forget and retain scores.
Figures.~\ref{proposal_perepoch} and~\ref{proposal_perepoch_gemma} show our method's performance on Llama3.1-8B Instruct and Gemma2-9B Instruct, respectively.
Figures.~\ref{perepoch} and~\ref{perepoch_gemma} show the evaluation results for baseline methods: GA, NPO, and RMU on Llama3.1-8B Instruct and Gemma2-9B Instruct, respectively. \par
These results highlight a key advantage of our method: it substantially reduces the forget score while preserving a high retain score throughout the unlearning process.
In contrast, baseline methods tend to exhibit a correlated degradation in both forget and retain scores, indicating a lack of selective forgetting.
This suggests that our method is more effective at precisely removing target knowledge while minimizing collateral damage to unrelated knowledge.

\subsection{Effects on Attribute Rate}
\begin{figure}[tb]
\begin{center}
\includegraphics[width=0.48\textwidth]{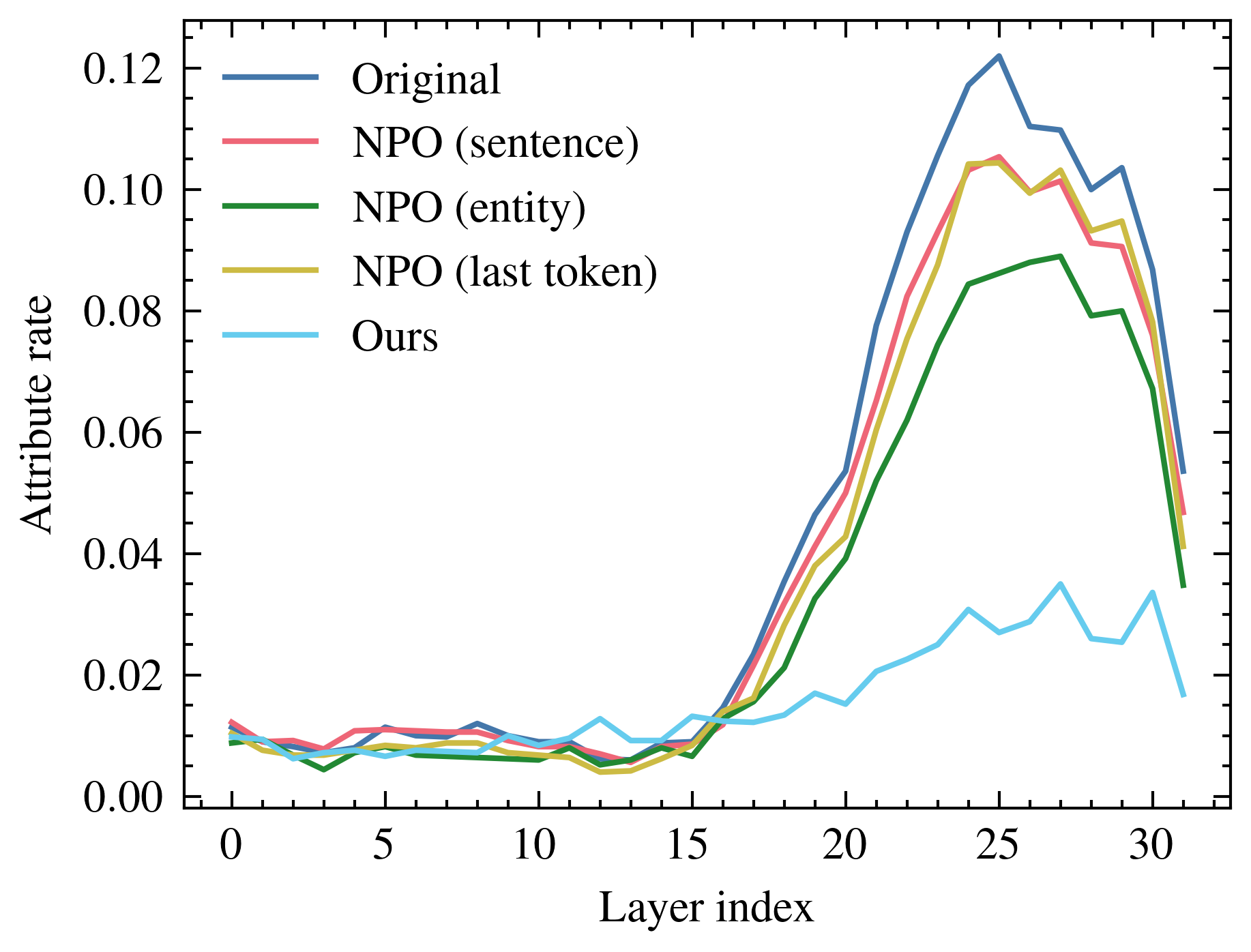}
\end{center}
\caption{\textbf{Attribute rate on Llama3.1-8B Instruct.}
We show the average attribute rate over 20 target entities.
Results are reported for the original model, three \textbf{NPO-based} baselines, and our method.
}
\label{attr_rate_npo}
\end{figure}

\begin{figure}[tb]
\begin{center}
\includegraphics[width=0.48\textwidth]{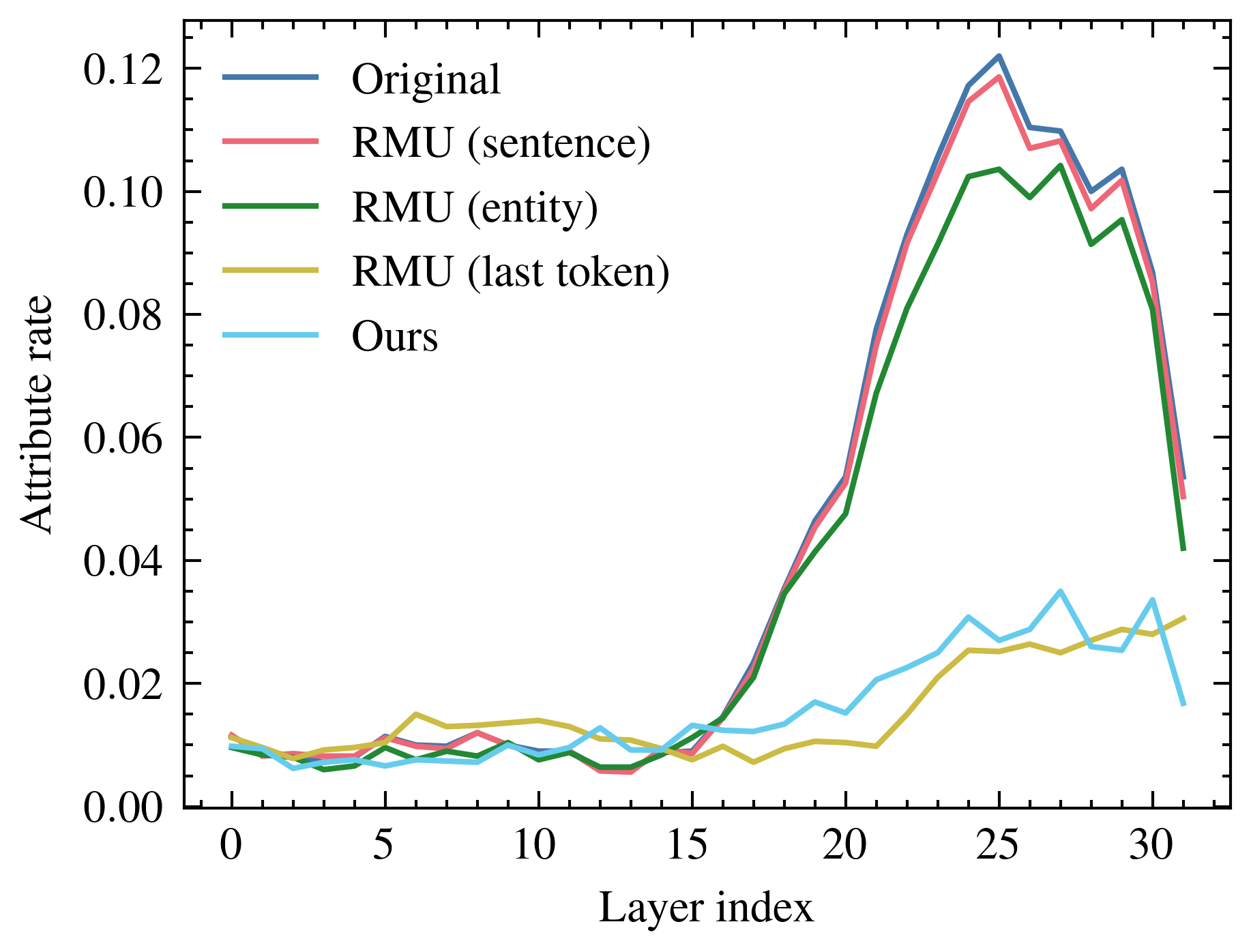}
\end{center}
\caption{\textbf{Attribute rate on Llama3.1-8B Instruct.}
We show the average attribute rate over 20 target entities.
Results are reported for the original model, three \textbf{RMU-based} baselines, and our method.
}
\label{attr_rate_rmu}
\end{figure}

\begin{figure}[tb]
\begin{center}
\includegraphics[width=0.48\textwidth]{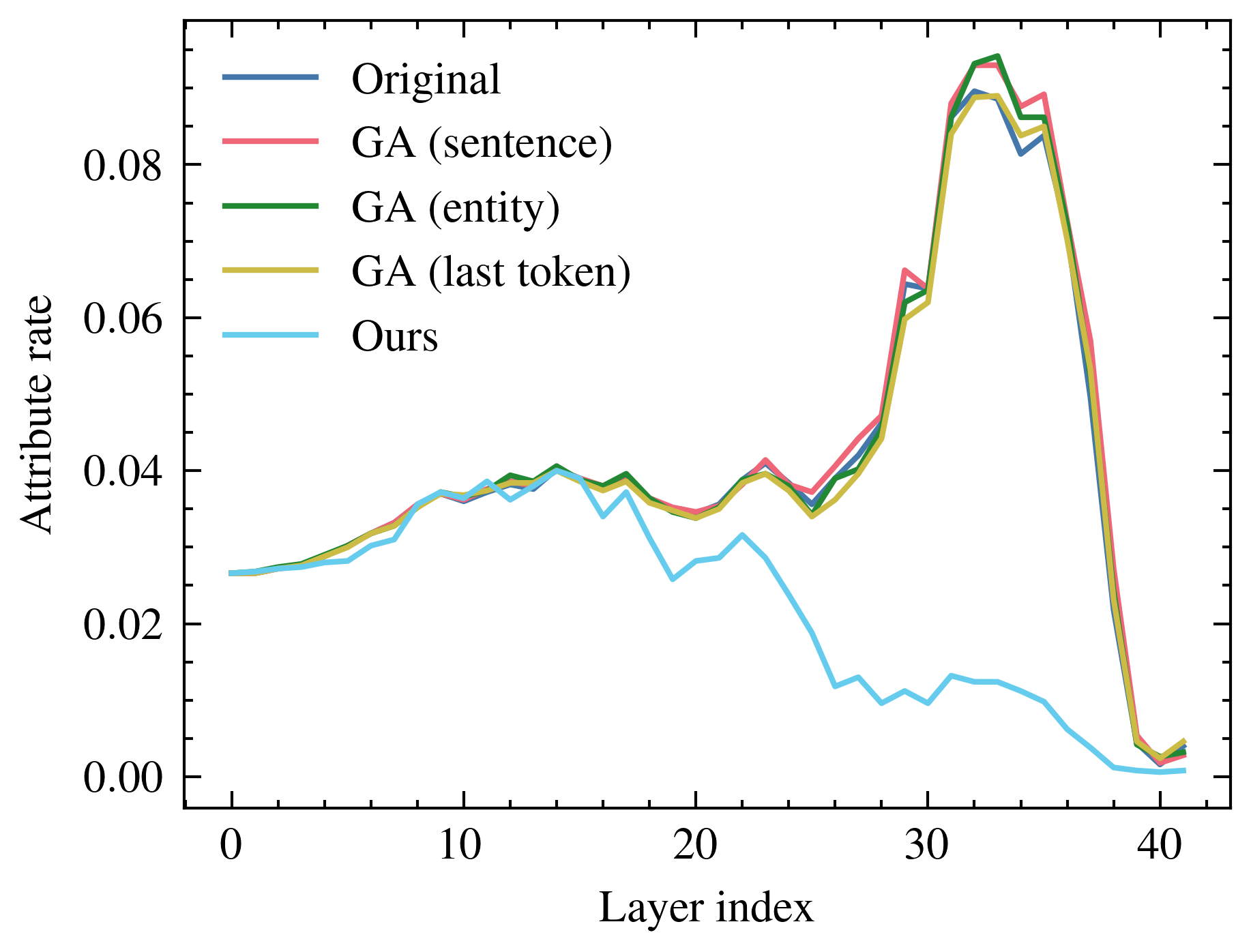}
\end{center}
\caption{\textbf{Attribute rate on Gemma2-9B Instruct.}
We show the average attribute rate over 20 target entities.
Results are reported for the original model, three \textbf{GA-based} baselines, and our method.
}
\label{attr_rate_gemma}
\end{figure}

\begin{figure}[tb]
\begin{center}
\includegraphics[width=0.48\textwidth]{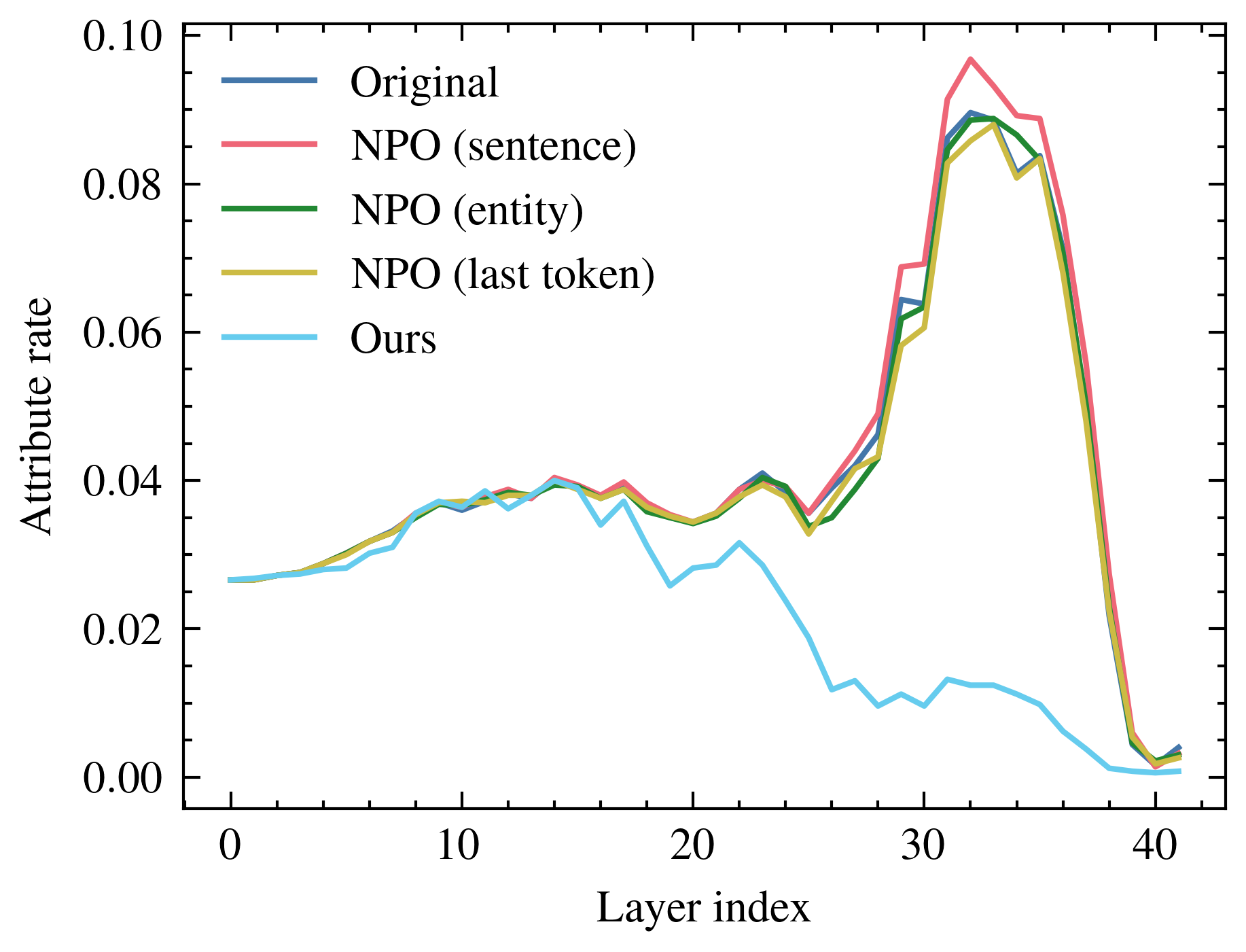}
\end{center}
\caption{\textbf{Attribute rate on Gemma2-9B Instruct.}
We show the average attribute rate over 20 target entities.
Results are reported for the original model, three \textbf{NPO-based} baselines, and our method.
}
\label{attr_rate_npo_gemma}
\end{figure}

\begin{figure}[tb]
\begin{center}
\includegraphics[width=0.48\textwidth]{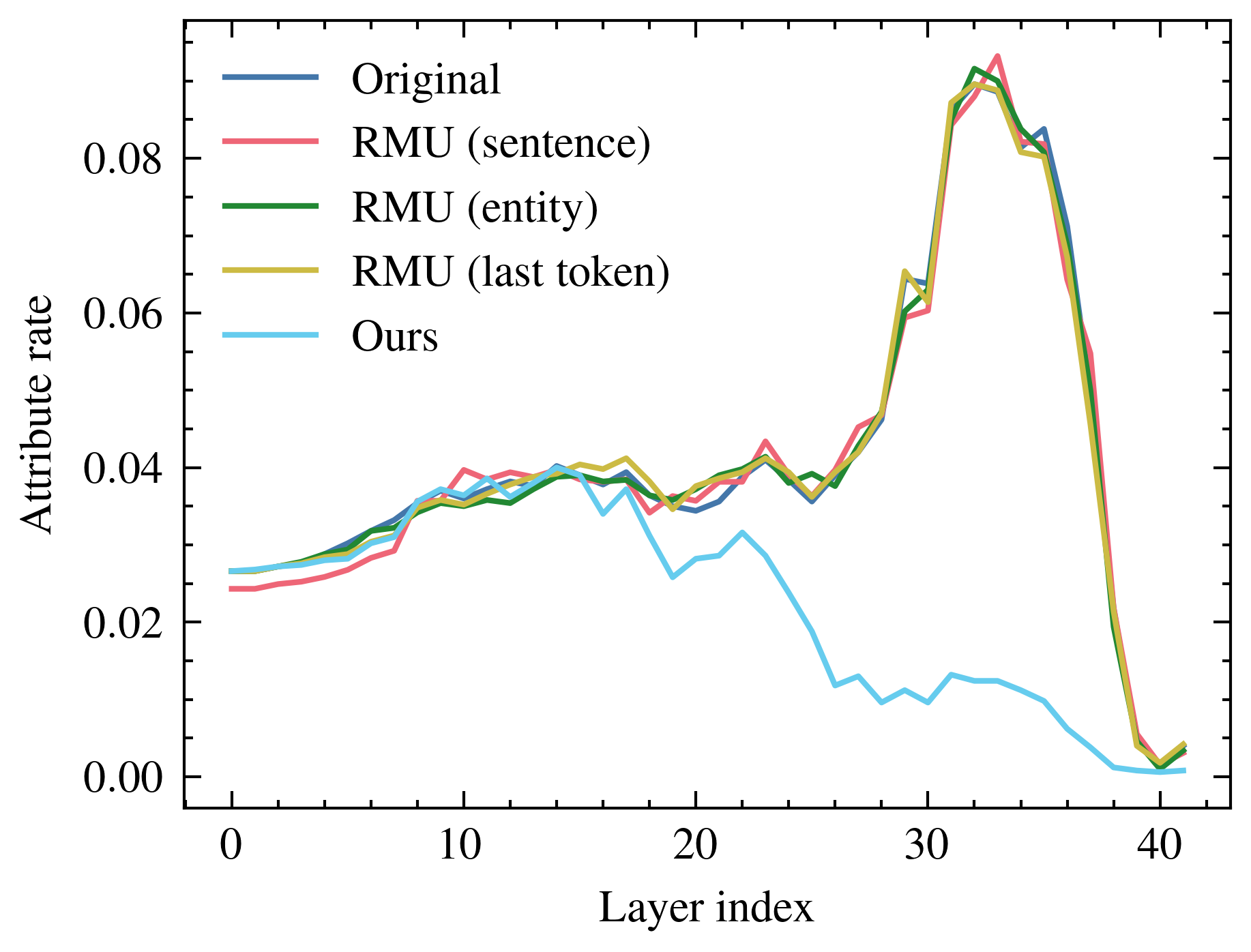}
\end{center}
\caption{\textbf{Attribute rate on Gemma2-9B Instruct.}
We show the average attribute rate over 20 target entities.
Results are reported for the original model, three \textbf{RMU-based} baselines, and our method.
}
\label{attr_rate_rmu_gemma}
\end{figure}

We investigate how each unlearning method affects the attribute rate, which reflects the degree to which the model's activations encode entity-related information.
Here, we include results from LLMs (Llama3.1-8B and Gemma2-9B Instruct models) and baselines (GA, NPO, and RMU).
Figures.~\ref{attr_rate_npo} and~\ref{attr_rate_rmu} show results on Llama3.1-8B Instruct for NPO-based baselines and RMU-based baselines. 
Figures.~\ref{attr_rate_gemma},~\ref{attr_rate_npo_gemma}, and~\ref{attr_rate_rmu_gemma} shows the results on Gemma2-9B Instruct.

\subsection{Effects on Recognition Latents}
\begin{figure*}[tb]
\begin{minipage}{0.48\textwidth}
\begin{center}
\includegraphics[width=\textwidth]{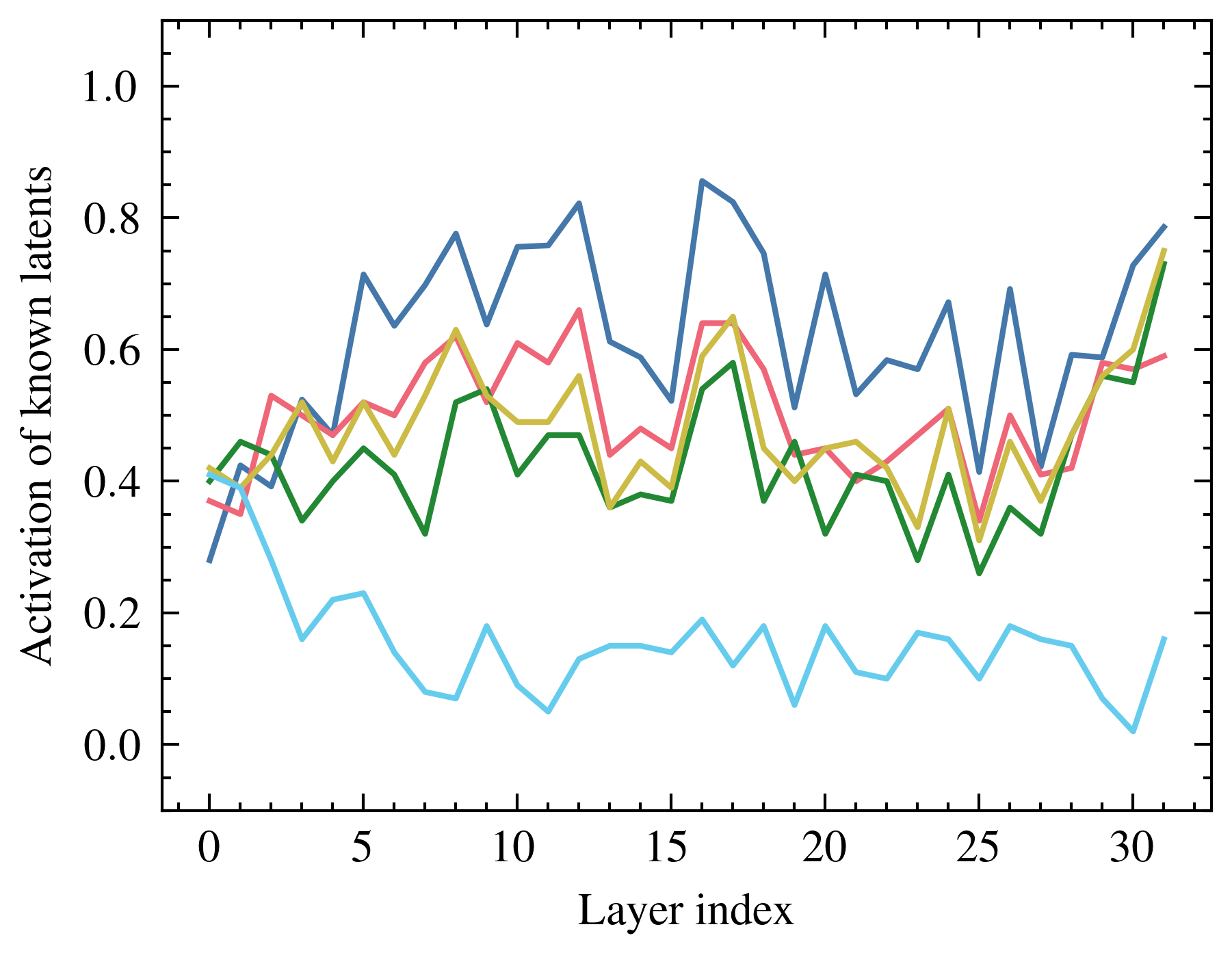}
\subcaption{Results for known latents.}
\end{center}
\end{minipage}
\begin{minipage}{0.48\textwidth}
\begin{center}
\includegraphics[width=\textwidth]{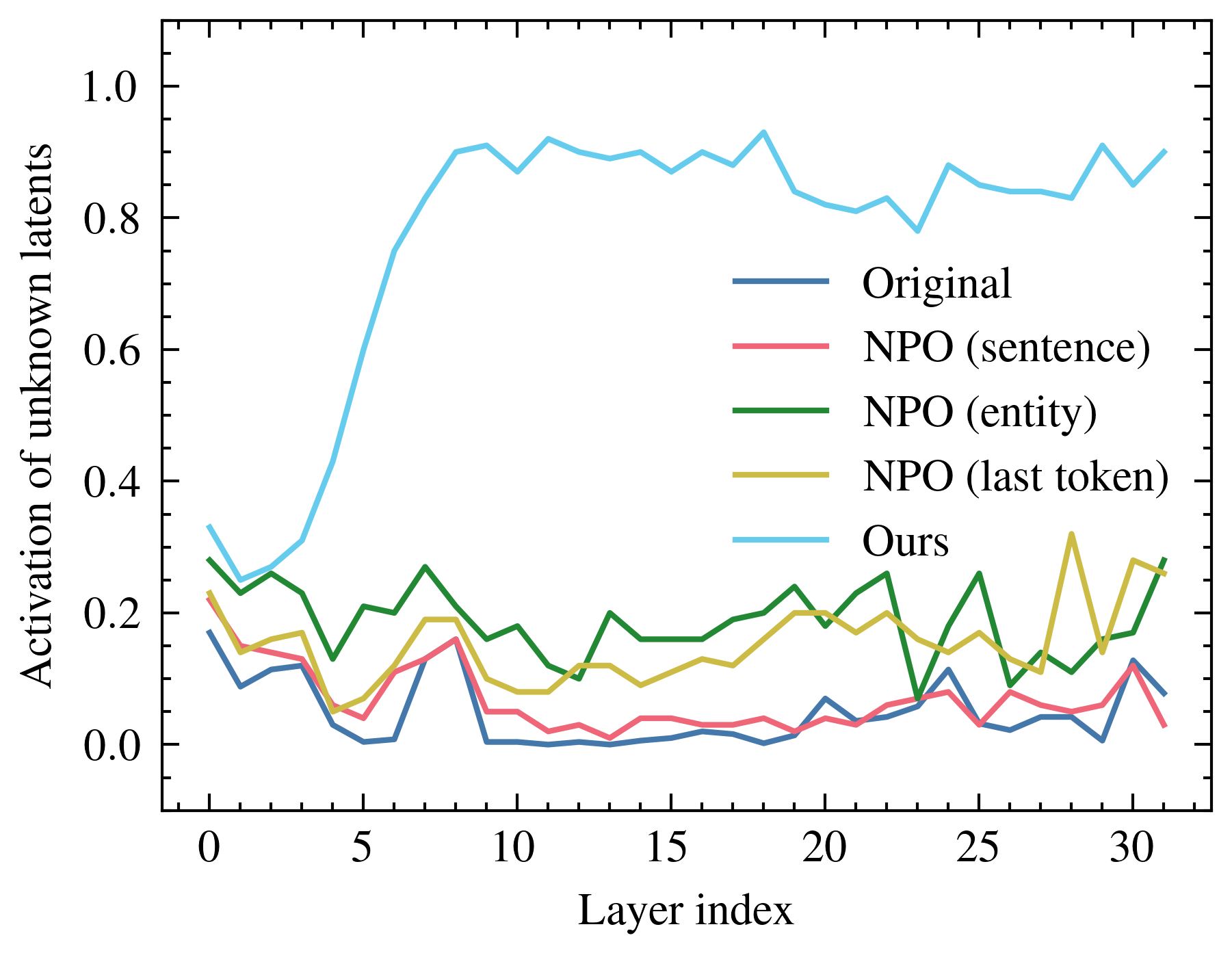}
\subcaption{Results for unknown latents.}
\end{center}
\end{minipage}
\caption{\textbf{Activation frequency of recognition latents.}
We report the average activation frequency of known and unknown latents when the forgotten entity is input, aggregated over 20 targets.
We compare the original model with several NPO-based unlearning configurations and our method.}
\label{rec_latents_fire_npo}
\end{figure*}

\begin{figure*}[tb]
\begin{minipage}{0.48\textwidth}
\begin{center}
\includegraphics[width=\textwidth]{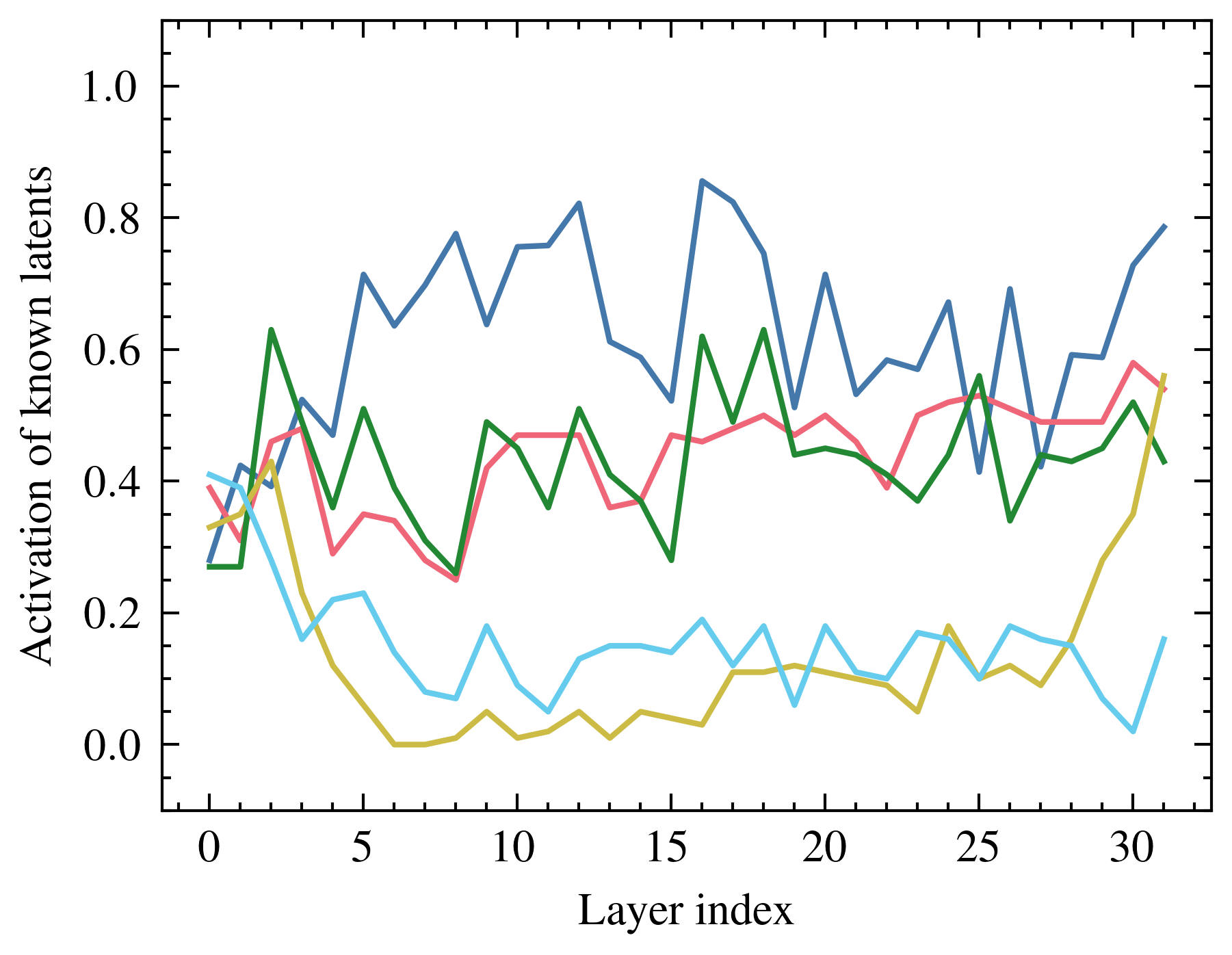}
\subcaption{Results for known latents.}
\end{center}
\end{minipage}
\begin{minipage}{0.48\textwidth}
\begin{center}
\includegraphics[width=\textwidth]{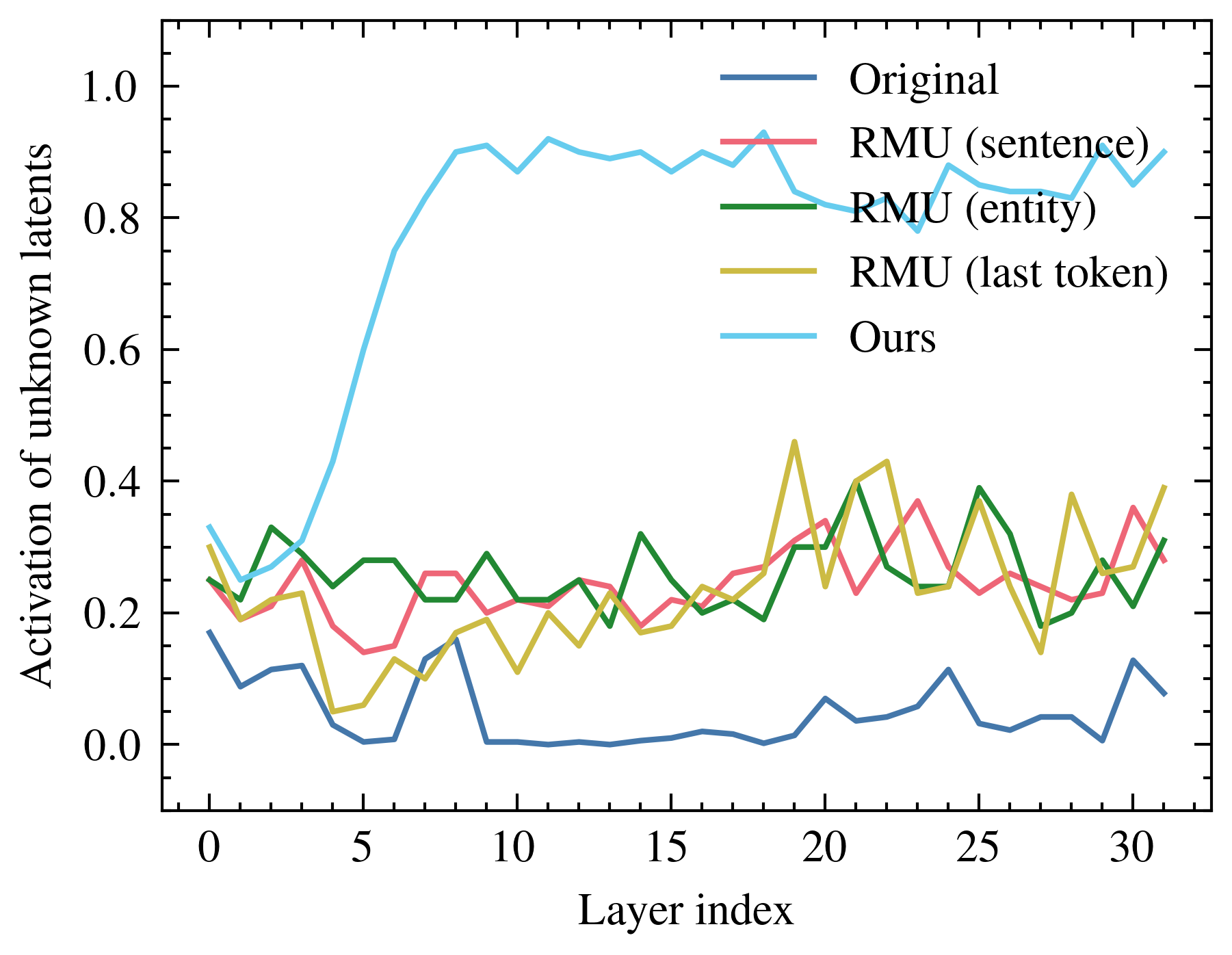}
\subcaption{Results for unknown latents.}
\end{center}
\end{minipage}
\caption{\textbf{Activation frequency of recognition latents.}
We report the average activation frequency of known and unknown latents when the forgotten entity is input, aggregated over 20 targets.
We compare the original model with several RMU-based unlearning configurations and our method.}
\label{rec_latents_fire_rmu}
\end{figure*}

\begin{figure*}[tb]
\begin{minipage}{0.48\textwidth}
\begin{center}
\includegraphics[width=\textwidth]{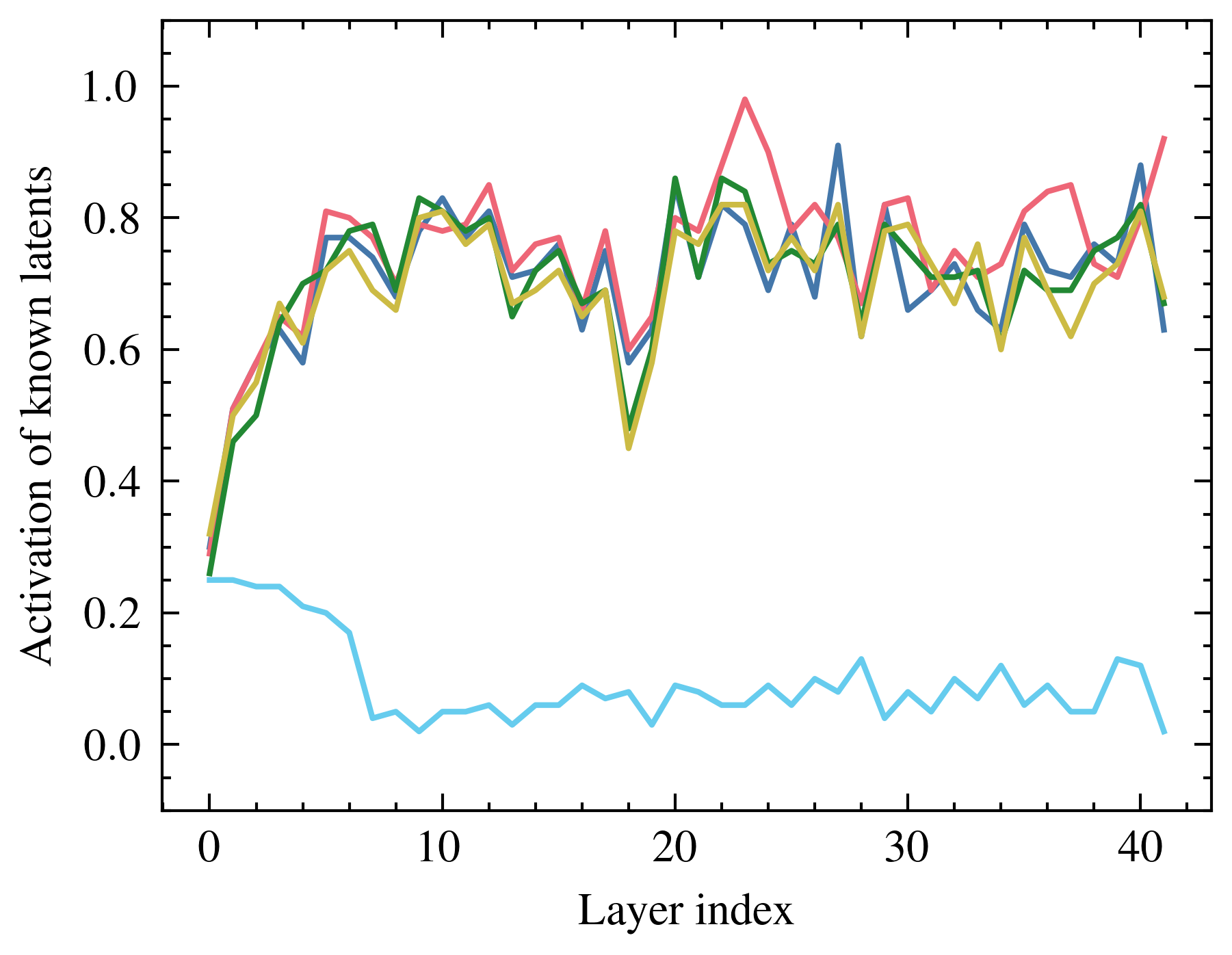}
\subcaption{Results for known latents.}
\end{center}
\end{minipage}
\begin{minipage}{0.48\textwidth}
\begin{center}
\includegraphics[width=\textwidth]{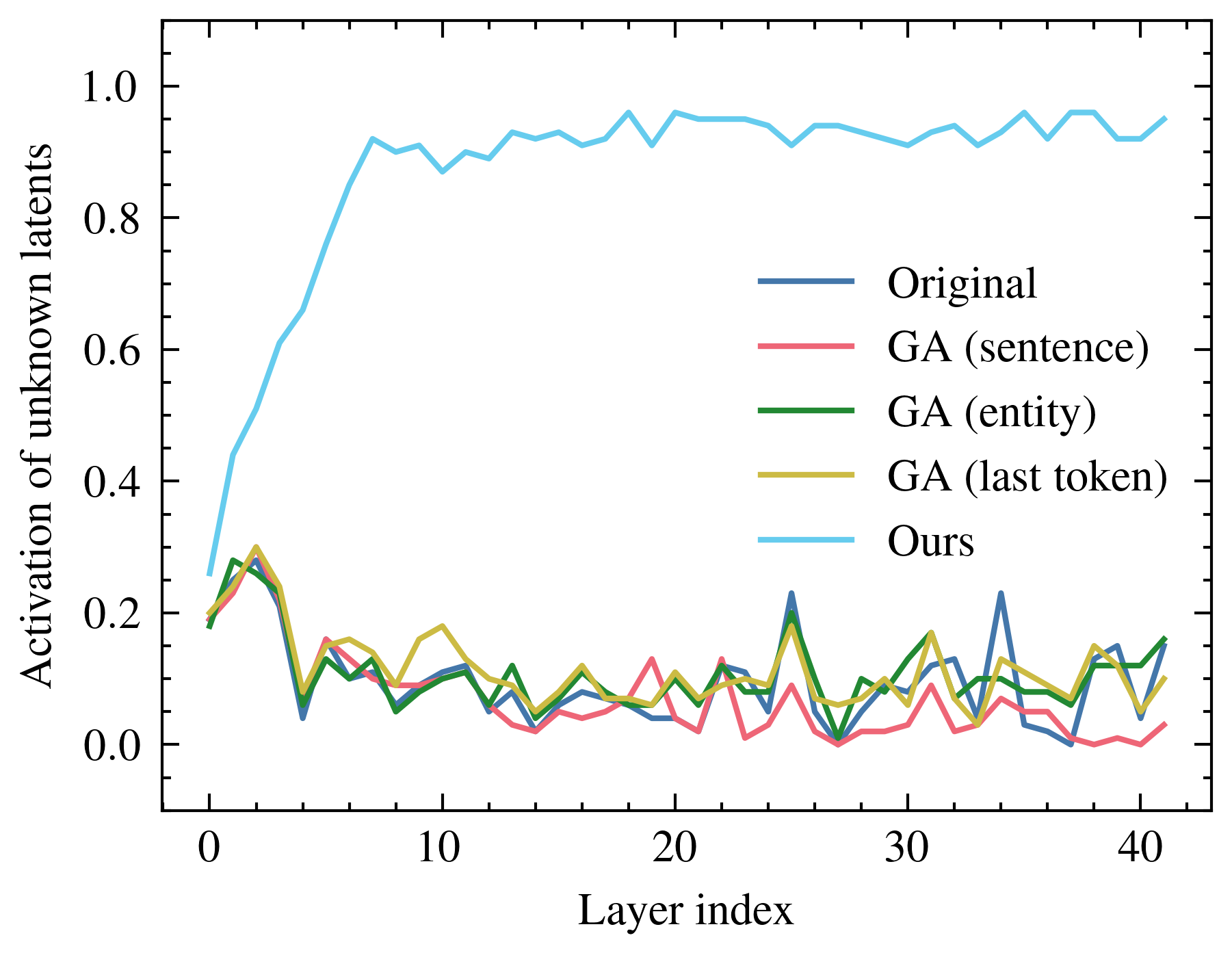}
\subcaption{Results for unknown latents.}
\end{center}
\end{minipage}
\caption{\textbf{Activation frequency of recognition latents.}
We report the average activation frequency of known and unknown latents when the forgotten entity is input, aggregated over 20 targets.
We compare the original model with several GA-based unlearning configurations and our method.}
\label{rec_latents_fire_ga_gemma}
\end{figure*}

\begin{figure*}[tb]
\begin{minipage}{0.48\textwidth}
\begin{center}
\includegraphics[width=\textwidth]{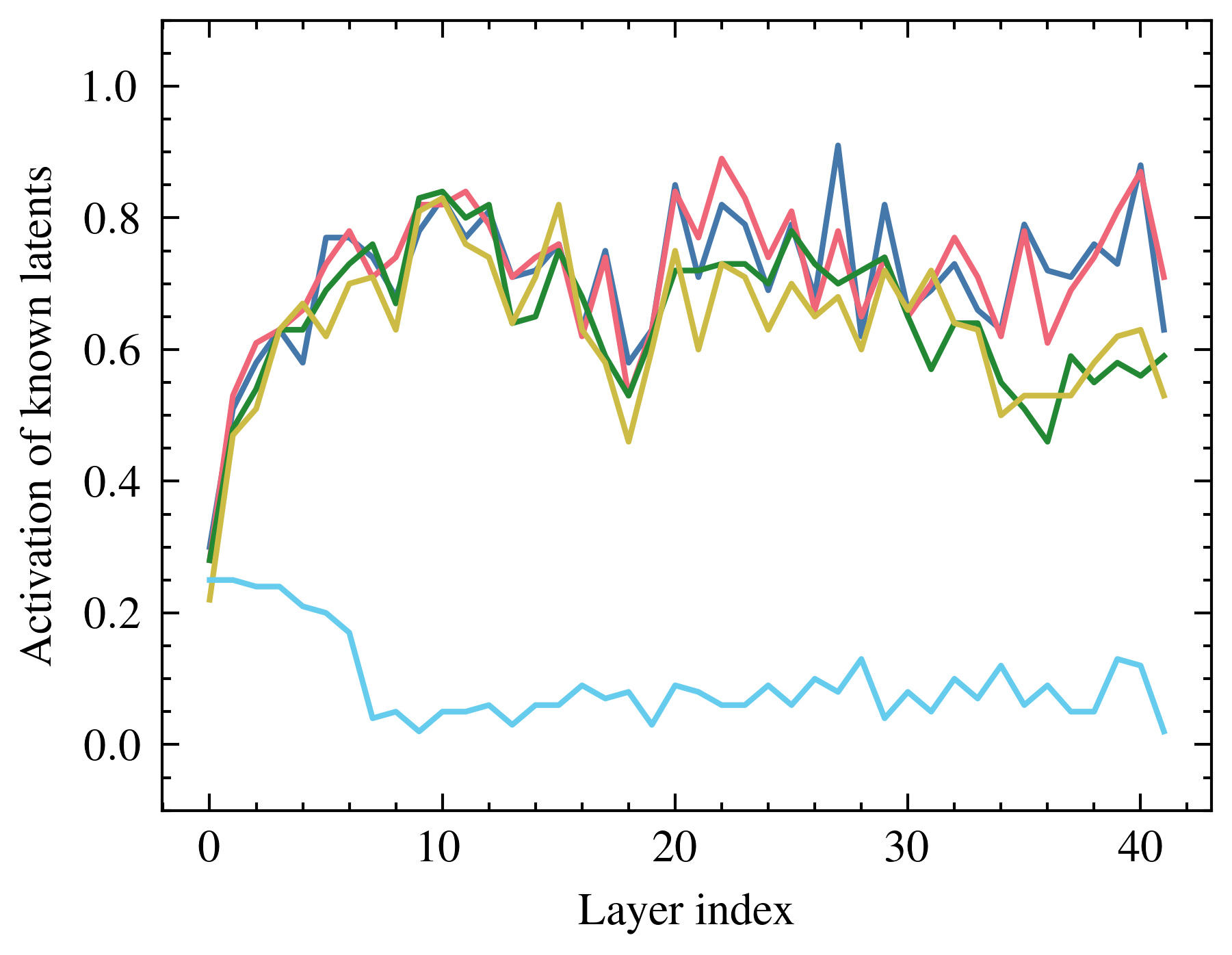}
\subcaption{Results for known latents.}
\end{center}
\end{minipage}
\begin{minipage}{0.48\textwidth}
\begin{center}
\includegraphics[width=\textwidth]{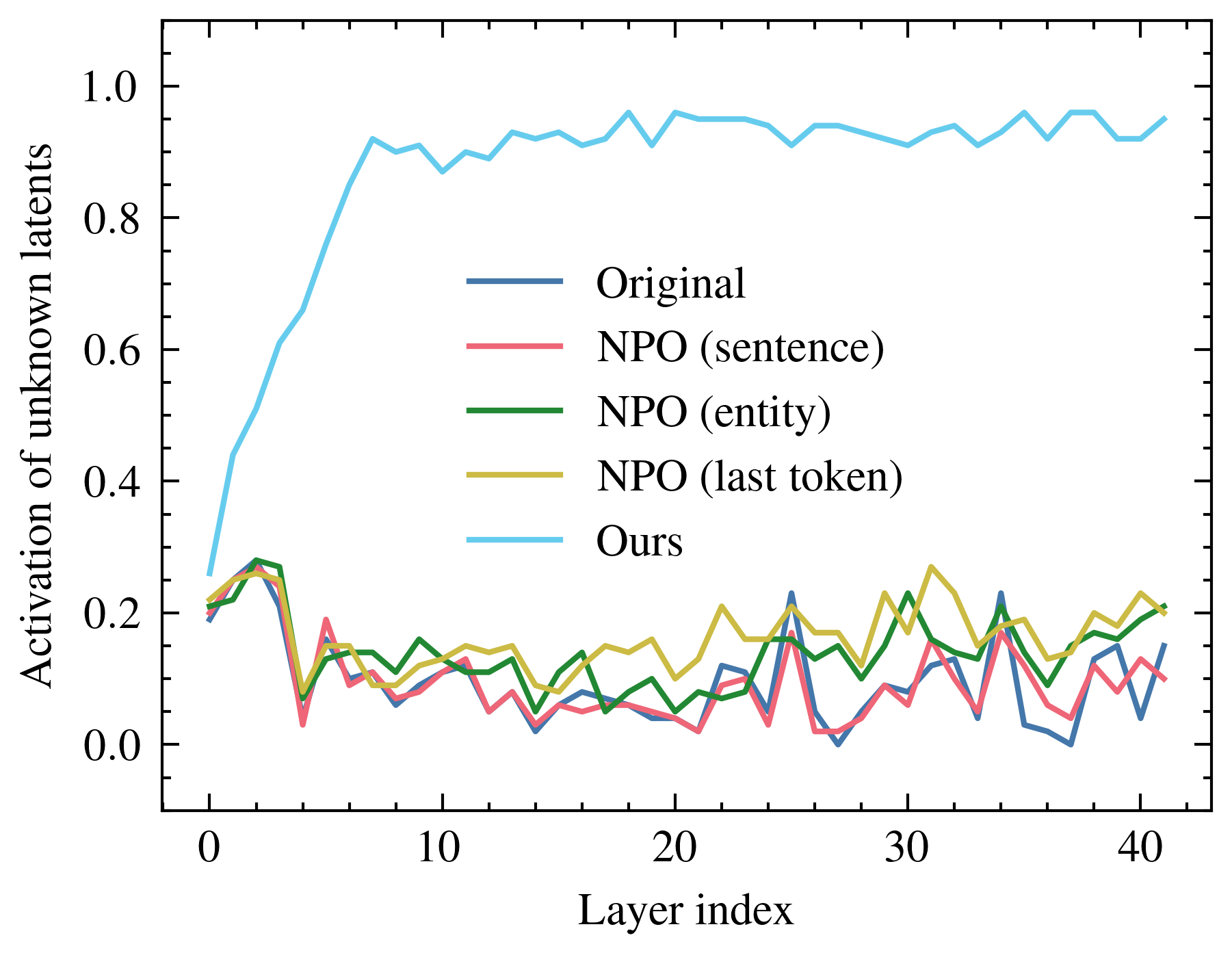}
\subcaption{Results for unknown latents.}
\end{center}
\end{minipage}
\caption{\textbf{Activation frequency of recognition latents.}
We report the average activation frequency of known and unknown latents when the forgotten entity is input, aggregated over 20 targets.
We compare the original model with several NPO-based unlearning configurations and our method.}
\label{rec_latents_fire_npo_gemma}
\end{figure*}

\begin{figure*}[tb]
\begin{minipage}{0.48\textwidth}
\begin{center}
\includegraphics[width=\textwidth]{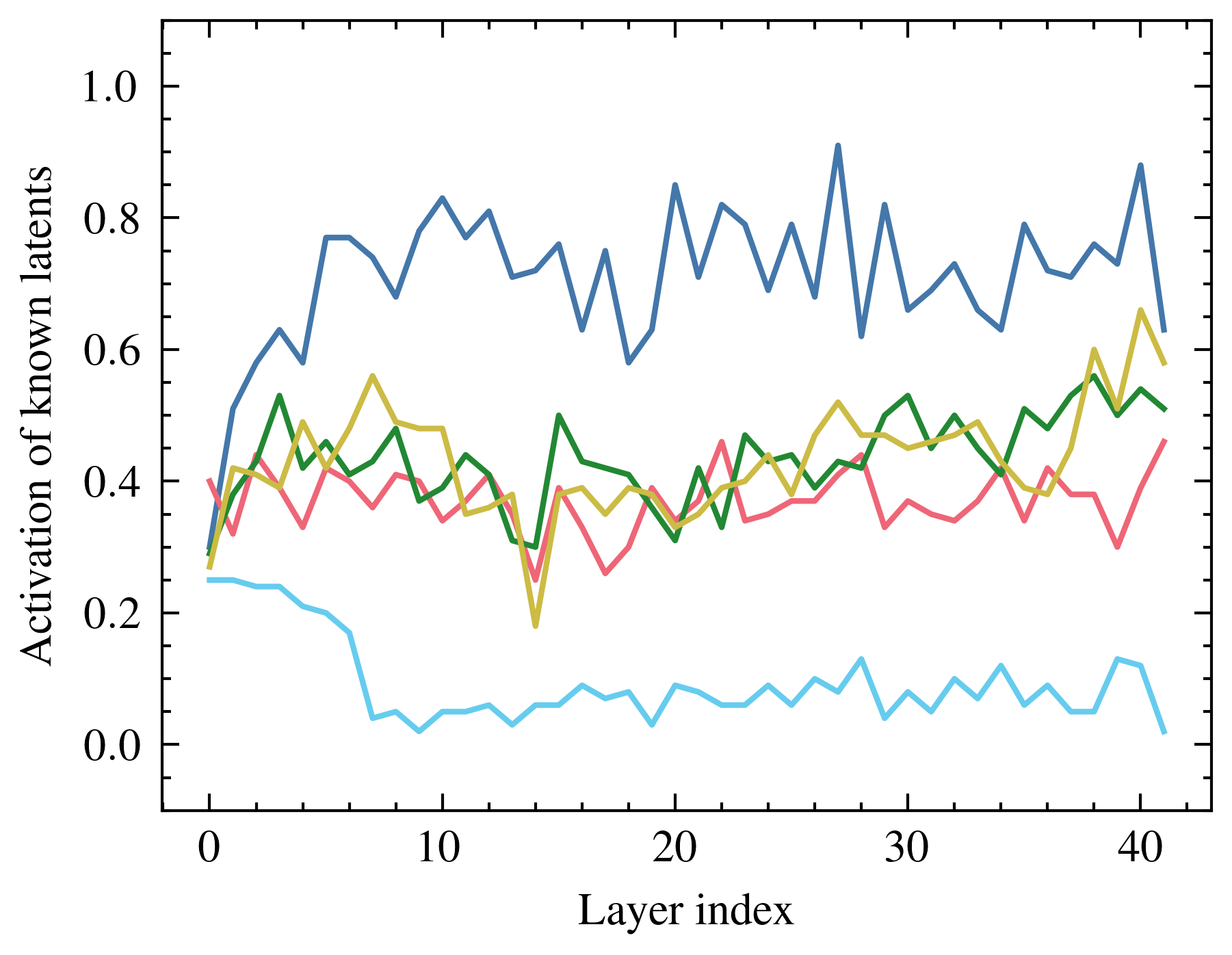}
\subcaption{Results for known latents.}
\end{center}
\end{minipage}
\begin{minipage}{0.48\textwidth}
\begin{center}
\includegraphics[width=\textwidth]{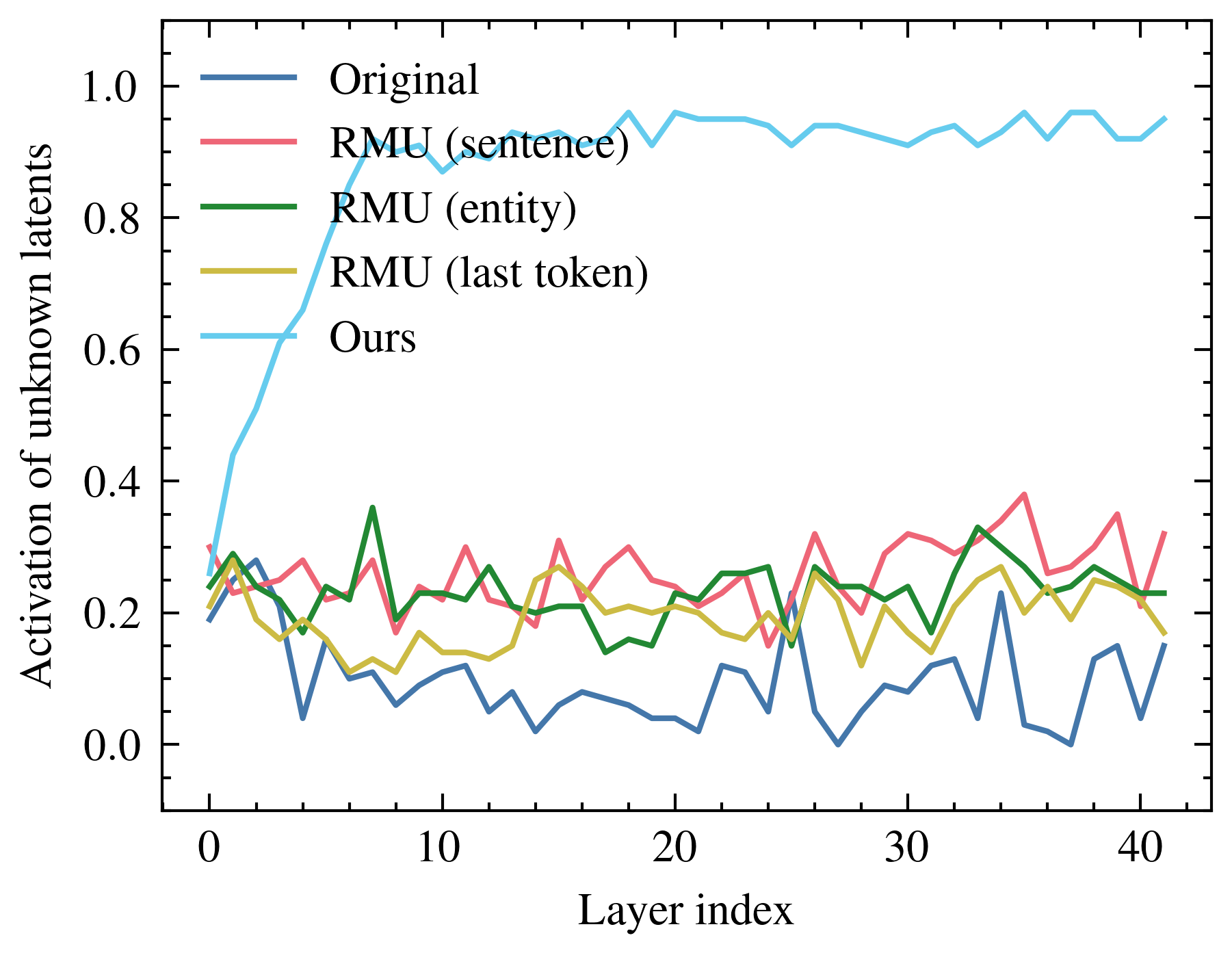}
\subcaption{Results for unknown latents.}
\end{center}
\end{minipage}
\caption{\textbf{Activation frequency of recognition latents.}
We report the average activation frequency of known and unknown latents when the forgotten entity is input, aggregated over 20 targets.
We compare the original model with several RMU-based unlearning configurations and our method.}
\label{rec_latents_fire_rmu_gemma}
\end{figure*}

We analyze how each unlearning method affects the model's recognition latents.
While the main paper reports the result for GA on Llama3.1-8B Instruct, here we extend the analysis to additional baselines (NPO, RMU) and include results on Gemma2-9B Instruct.
Figures.~\ref{rec_latents_fire_npo} and~\ref{rec_latents_fire_rmu} show the average activation frequency of known and unknown latents when the forgotten entity is presented as input on Llama3.1-8B Instruct.
We compare the proposed method with NPO-based baselines and RMU-based baselines, respectively.
In addition, Figures.~\ref{rec_latents_fire_ga_gemma},~\ref{rec_latents_fire_npo_gemma}, and~\ref{rec_latents_fire_rmu_gemma} show the average activation frequncy of known and unknown latents on Gemma2-9B Instruct.
We compare the proposed method with GA-based baselines, NPO-based baselines, and RMU-based baselines, respectively.

\section{Experimental Details}
\label{experimental_setup}
\subsection{Experimental Setup}
In our experiments, we utilize two instruction-tuned LLMs, i.e., Llama3.1-8B Instruct and Gemma2-9B Instruct.
For identifying recognition latents, we utilize SAEs from Llama-Scope~\citep{DBLP:journals/corr/abs-2410-20526} and Gemma-Scope~\citep{DBLP:journals/corr/abs-2408-05147}, respectively.
All experiments were conducted on two NVIDIA H100 GPUs (80GB each). \par
In the unlearning process, we use the Adam optimizer and employ a reverse cosine learning rate schedule during unlearning.
Specifically, we define the schedule as follows:
\begin{eqnarray}
    \lambda_t = \lambda_\mathrm{min} + \frac{1}{2}(\lambda_\mathrm{max}-\lambda_\mathrm{min})\left(1-\cos\left(\frac{t}{T}\pi\right) \right),
\end{eqnarray}
where $\lambda_\mathrm{min}$ and $\lambda_\mathrm{max}$ denote the minimum and maximum learning rates, $T (=200)$ is the maximum number of epochs, and $t$ is the current epoch.
We set the number of optimization steps per epoch to 5.
We used the following learning rate settings for each LLM.
Unless otherwise noted, the same schedule was applied to both our proposal and baselines.
\begin{itemize}
  \item \textbf{Llama3.1-8B Instruct}: $(1\times10^{-6},\ 1\times10^{-5})$.
  \item \textbf{Gemma2-9B Instruct}: $(1\times10^{-7},\ 1\times10^{-5})$.
\end{itemize}
For all unlearning experiments, we set the batch size to 1 when the method operates on entity-specific representations, such as our proposed method and baselines that use the entity name or its last token. 
For baselines on explanatory corpus, including GA (sentence), NPO (sentence), and RMU (sentence), we use a batch size of 4.

\subsection{Evaluation Dataset}
To conduct our main experiments, we use the Real-World Knowledge Unlearning (RWKU) dataset~\citep{DBLP:conf/nips/JinCWHYL00024}, a comprehensive benchmark specifically designed for evaluating knowledge unlearning in large language models. 
RWKU includes 200 real-world individuals as unlearning targets, selected from Wikipedia based on popularity metrics, and provides 13,131 multi-level probes. The dataset consists of three types of evaluation probes: forget, retain, and utility probes. \par
\textbf{Forget probes} are categorized into three formats:
(i) fill-in-the-blank (FB), 
(ii) question-answer (QA), and 
(iii) adversarial attacks (AA).  
AA probes include nine types of jailbreak-style prompts (e.g., prefix injection, role-playing, synonym manipulation, cross-lingual queries) designed to elicit forgotten knowledge, thus serving as a rigorous test of unlearning robustness.
\textbf{Retain probes} adopt the FB and QA formats to assess whether related but non-targeted knowledge remains intact.
We compute forget and retain scores based on exact match accuracy, defined as the fraction of instances in which the predicted answer contains the ground-truth label. 
\textbf{Utility evaluation} in RWKU spans five benchmarks:
\begin{itemize}
  \item \textbf{MMLU}~\citep{huutien2025improvingllmunlearningrobustness}: a collection of multiple-choice questions to assess general abilities. We report 5-shot accuracy.
  \item \textbf{Big-Bench-Hard (BBH)}~\citep{DBLP:conf/acl/SuzgunSSGTCCLCZ23}: includes 27 reasoning tasks. We use the Exact Match (EM) score, defined as the proportion of N-grams in the model’s output that exactly match the ground-truth answer.
  \item \textbf{TruthfulQA}~\citep{lin-etal-2022-truthfulqa}: evaluates truthfulness via the MC1 task; we report 6-shot accuracy.
  \item \textbf{TriviaQA}~\citep{joshi-etal-2017-triviaqa}: evaluates factuality using open-ended questions; we report F1 scores under 6-shot settings.
  \item \textbf{AlpacaEval}~\citep{alpaca_eval}: measures fluency using a weighted average of bi- and tri-gram entropies~\citep{DBLP:conf/nips/MengBAB22,DBLP:conf/nips/ZhangGGGLBD18}.
\end{itemize}
Together, these metrics provide a holistic view of forgetting effectiveness, retention of non-target knowledge, and overall utility preservation.

\subsection{Attribute Rate}
\label{attribute_rate}
To assess how much a model's internal activations encode entity-related information, we introduce the attribute rate~\citep{geva-etal-2023-dissecting}.
Attribute rate is the proportion of entity-related tokens, referred to as ``attributes'', among the top-50 tokens projected from the model's activation.
Let $\bm{h}(\bm{e}^\mathrm{(t)};\bm{\theta}) \in \mathbb{R}^d$ denote the activation of the model $\bm{\theta}$ at layer $l$, and $E\in \mathbb{R}^{|V|\times d}$ denote the model's vocabulary embedding matrix.
We compute a token distribution by projecting the activation onto the vocabulary space:
\begin{eqnarray}
    p_l^\mathrm{(t)} = \mathrm{softmax}(E\bm{h}(\bm{e}^\mathrm{(t)};\bm{\theta}))
\end{eqnarray}
From this distribution, we extract top-50 tokens with the highest probabilities.
The attribute rate is then defined as:
\begin{eqnarray}
    \mathrm{AttributeRate}(\bm{e}^\mathrm{(t)})=\frac{\mathrm{Top}_{50}(p_l^\mathrm{(t)})\cap \mathcal{A}^\mathrm{(t)}}{50},
\end{eqnarray}
where $\mathcal{A}^\mathrm{(t)}$ denotes the set of entity-related attribute tokens for $\bm{e}^\mathrm{(t)}$, obtained from passages in the RWKU dataset.
This metric provides a quantitative approximation of the semantic richness of the subject representation in terms of relevant attributes. \par
To construct $\mathcal{A}^\mathrm{(t)}$, we leverage passages about $\bm{e}^\mathrm{(t)}$ in the RWKU dataset, tokenize the passages, and filter out stopwords and subword fragments.
The resulting set contains tokens that are frequently associated with the target entity in natural text and are therefore likely to represent factual attributes.

\end{document}